\documentclass[letter,12pt]{report}
\usepackage[utf8]{inputenc}
\usepackage{graphicx}
\usepackage{natbib}
\usepackage{amsmath,amsfonts}
\usepackage{hyperref}
\hypersetup{
    colorlinks,
    citecolor=black,
    filecolor=black,
    linkcolor=black,
    urlcolor=black
}
\usepackage{xcolor}
\usepackage{rotating}
\usepackage{todonotes}
\usepackage[toc,page]{appendix}

\title{xVIO:\\
	   A Range-Visual-Inertial Odometry Framework}
\author{Jeff Delaune, David S. Bayard and Roland Brockers \\\\
	   Jet Propulsion Laboratory, California Institute of Technology
    }

\begin{document}

\maketitle

\begin{abstract}
xVIO is a range-visual-inertial odometry algorithm implemented at JPL. It has been
demonstrated with closed-loop controls on-board unmanned rotorcraft equipped
with off-the-shelf embedded computers and sensors. It can operate at daytime
with visible-spectrum cameras, or at night time using thermal infrared cameras.
This report is a complete technical description of xVIO. It includes an overview
of the system architecture, the implementation of the navigation filter, along
with the derivations of the Jacobian matrices which are not already published in
the literature.
\\\\
The research described in this paper was carried out at the
Jet Propulsion Laboratory, California Institute of Technology,
under a contract with the National Aeronautics and Space
Administration (80NM0018D0004).
\\\\
\textcopyright 2020 California Institute of Technology. Government
sponsorship acknowledged.
\end{abstract}

\tableofcontents

\chapter{System Overview}

\emph{Visual-Inertial Odometry} (VIO) is the use of one camera and an
\emph{Inertial Measurements Unit} (IMU) to estimate the position and orientation
of the sensing platform. Range-VIO adds a range sensor measurement to that
sensor fusion problem.

Although xVIO makes no assumption about the platform itself, it is worth
keeping in mind that it was developed for rotorcraft flight applications. The
`x' in xVIO has no particular meaning but one should feel free to see it either
as a pair of propellers, a symbol for accuracy, or sensor fusion. xVIO is
versatile and can run either range-VIO, VIO or simply inertial odometry.

On the date this report was last updated, xVIO had been tested on datasets both
indoors and outdoors, at daytime with visible and at night time with long-wave
infrared cameras, in flight and hand-held, on real and simulated data. It ran at 30 frames per second with modern off-the-shelf sensors and
embedded computers\footnote{Qualcomm Snapdragon Flight Pro}. That proved enough to meet
the real-time contraints and close the control loop to enable autonomous flight.

\begin{figure}[ht]
  \centering
    \includegraphics[width=\textwidth]{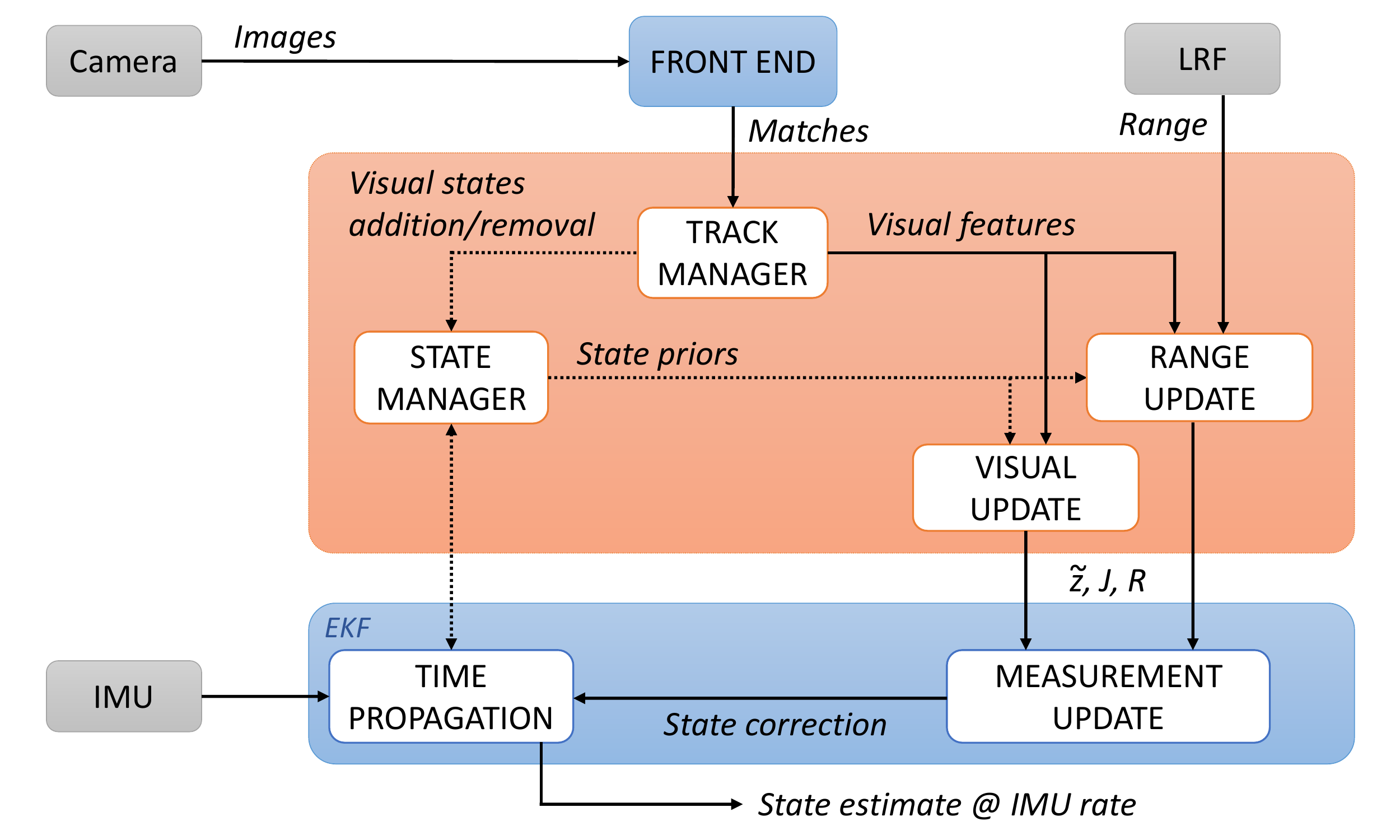}
  \caption{xVIO architecture.}
  \label{fig:system-architecture}
\end{figure}

Figure~\ref{fig:system-architecture} illustrates the software architecture at
system level. It can be used as a guide to read through the next sections, which
will introduce the main software components in the figure: sensor interface,
visual front end, state estimator, and VIO measurement construction.

\section{Sensor Interface}

xVIO code is written in C++11 and compiles into an executable that has been
tested in Linux Ubuntu 16.04, 18.04 and 20.04. It receives timestamped sensor
through two call functions: one that reads an 8-bit grayscale image coming from
the camera driver, another one that reads angular rates $\boldsymbol{\omega}$
and specific force $\boldsymbol{f}$ coming from the IMU driver.

xVIO's callbacks are currently wrapped to interface with the
ROS\footnote{Robot Operating System: \url{http://www.ros.org}} framework, which
provides a multi-computer real-time communication network between the sensor
drivers, xVIO, and the user interface. xVIO was tested with ROS
Kinetic, Melodic and Noetic versions.

\section{Visual Front End}

The visual front end matches feature points between the current image and the
previous one. Its output is a list of $N$ image coordinate pairs.

On the first image, feature points are detected using the FAST algorithm~\citep{
Rosten2006}. A neighborhood parameter can be set in pixels to ensure feature are
not too close to each other. The image can be divided in tiles, and the maximum
number of features per tile enforce to bound the computational cost of the
visual front end if needed.

Subsequently, the features are tracked frame-to-frame with the Lucas-Kanade
algorithm~\citep{Lucas1981}. When the number of feature matches reached below a
user-specified threshold, FAST features are re-detected on the previous image
and matched with the current one. Existing matches are preserved in this. The
neighborhood parameter prevents a new feature to be extracted next to an exising
match. Outlier matches are detected using the RANSAC algorithm with a fundamental
matrix model~\citep{Fischler1981}.

The tracker class uses the OpenCV\footnote{\url{https://opencv.org}}
implementation of these algorithms. More details about the actual Lucas-Kanade
implementation can be found in~\citet{Bouguet2001}. xVIO, the tracker, feature, image and camera class are part of the X library.

\section{State Estimator}

xVIO uses an Extended Kalman Filter (EKF) for state estimation. The specific EKF
implementation is called xEKF and is also part of the X library. It propagates the position, velocity, attitude and inertial
biases states using IMU measurements in the dynamics model discussed in
Subsection~\ref{sub:continous-time-model}. It also implements a ring-buffer to
accommodate sensor and processing delays in real-time operations. xVIO provides xEKF with an update through
image and range measurement residuals $\tilde{\boldsymbol{z}}$, Jacobian matrix
$\boldsymbol{J}$ and covariance $\boldsymbol{R}$.

\section{VIO Measurement Construction}

The core capability of xVIO is the ability to construct visual measurements
from image matches, in order to constrain the inertially-propagated EKF. This
process is illustrated in the orange block in
Figure~\ref{fig:system-architecture}.

Visual measurements can use either the
\emph{Simultaneous Localization And Mapping} (SLAM) or
\emph{Multi-State Constraint Kalman Filter} (MSCKF) paradigms. Both have pros
and cons but \citet{Li2012} showed that they are complimentary and can be
hybridized to get the most accuracy out of the computational resources
available. xVIO follows their approach closely, although not identically. Our
implementation of SLAM and MSCKF, along with a discussion of their pros and
cons, is detailed in Section~\ref{sec:vision-update}. \citeauthor{Li2014}'s
PhD thesis (\citeyear{Li2014}) is a good
reference about this approach in general, along with the references cited
locally in Chapter~\ref{ch:filter-implementation} and
\ref{ch:jacobians-derivations}. It is worth noting that even though EKF-based
VIO is only optimal up to linearization error, \citet{li2013ijrr} showed that they
could compete with approaches based on non-linear optimization in terms of
accuracy and computational cost.

We defined a track as the list image observations of a specific feature. When
feature matches come in from the visual front end, they are being appended
to an existing track and a new one is created by the track manager. Tracks can
be either of the SLAM or MSCKF type. An MSCKF track can be become a SLAM after
an MSCKF measurement through a process described in
Subsection~\ref{sub:feat-state-init}, or to distribute SLAM features among the
image tiles. At the start, the best feature in each style are used for SLAM,
since only this paradigm can provide constraints immediately. MSCKF measurements
are converted to SLAM feature when there are slots available.

SLAM and MSCKF require the creation, destruction and reparametrization of states
and their error covariance. A state manager handles the state queries, actions
and responses between the track manager, SLAM and MSCKF modules and the EKF.

\chapter{Filter Implementation}
\label{ch:filter-implementation}

\section{Notations}

A scalar is denoted by $x$, a vector by $\boldsymbol{x}$, and a matrix by
$\boldsymbol{X}$. The estimate of a random variable is written with the hat operator,
e.g. $\hat{\boldsymbol{x}}$ for a vector.

The coordinates of vector $\boldsymbol{x}$ in reference frame
$\left\{\mathit{a}\right\}$ are represented by $^a\boldsymbol{x}$. The time
derivative of $^a\boldsymbol{x}$ is noted $^a\dot{\boldsymbol{x}}$, such that
\begin{equation}
^a\dot{\boldsymbol{x}} = \frac{d{^a\boldsymbol{x}}}{dt}\;.
\label{eq:vector-time-derivative}
\end{equation}

The position, velocity and acceleration of frame $\left\{\mathit{b}\right\}$
with respect to frame $\left\{\mathit{a}\right\}$ are denoted by
$\boldsymbol{p}_a^b$, $\boldsymbol{v}_a^b$ and $\boldsymbol{a}_a^b$,
respectively.
When their coordinates are expressed in the axes of $\left\{\mathit{a}\right\}$,
velocity and acceleration are defined as time derivatives of position by
\begin{equation}
^a\boldsymbol{v}_a^b = {^a\dot{\boldsymbol{p}}_a^b}\;,
\label{eq:vel-def}
\end{equation}
\begin{equation}
^a\boldsymbol{a}_a^b = {^a\dot{\boldsymbol{v}}_a^b}\;.
\label{eq:acc-def}
\end{equation}
The angular velocity vector of $\left\{\mathit{b}\right\}$ with respect to
$\left\{\mathit{a}\right\}$ is represented by $\boldsymbol{\omega}_{a}^{b}$.

Unless noted otherwise, in the rest of this report these vectors are represented
by default by their coordinates in the origin frame
$\left\{\mathit{a}\right\}$ to simplify the presentation, e.g. for position
\begin{equation}
\boldsymbol{p}_a^b = {^a\boldsymbol{p}_a^b}\;.
\label{eq:default-vector-coordinate-frame}
\end{equation}

The quaternion describing the rotation from $\left\{ \mathit{a}\right\}$ to
$\left\{ \mathit{b}\right\}$ is denoted by $\boldsymbol{q}_{a}^{b}$.
$\boldsymbol{C}(\boldsymbol{q}_{a}^{b})$ is the coordinate change matrix
associated to that quaternion, such that
\begin{equation}
^b\boldsymbol{x} = \boldsymbol{C}(\boldsymbol{q}_{a}^{b}) ^a\boldsymbol{x}\;.
\label{eq:ref-frame-change}
\end{equation}

\section{Reference Frames}

The following frames are used in the filter. They are all right-handed.

\begin{itemize}
	\item World frame $\left\{ \mathit{w} \right\}$. It is fixed with respect to
		  the terrain, and assumed to be an inertial frame\footnote{This is
		  equivalent to neglecting the rotation rate of the target celestial
		  body. This assumption  is usual
		  for the typical mission duration of quadrotors}. The $z$-axis points
		  upwards along the local vertical, which is assumed to be constant
		  within the area of operations\footnote{This is only valid over short
		  distances.}.
		  
	\item IMU frame $\left\{ \mathit{i} \right\}$. It is the reference body
		  frame. It shares the same axes as the accelerometers and gyroscopes
		  measurements.
		  
	\item Camera frame $\left\{ \mathit{c} \right\}$. Its origin is at the optical
	center of the lens. The $z$-axis is pointing outwards along the optical axis.
	The $x$-axis is in the sensor plane, pointing right in the image. The $y$-axis
	completes the right-handed frames.
\end{itemize}

\section{State Space}

\subsection{States}
\label{sub:states}

The state vector $\boldsymbol{x}$ can be divided between the states related to
the IMU $\boldsymbol{x}_{I}$, and those related to vision $\boldsymbol{x}_{V}$.
\begin{equation}
\boldsymbol{x} = \begin{bmatrix}{\boldsymbol{x}_I}^T &
{\boldsymbol{x}_V}^{T}\end{bmatrix}^{T}
\label{eq:state-vec-iv}
\end{equation}

The inertial states $\boldsymbol{x}_{I} \in \mathbb{R}^{16}$ include the
position, velocity and orientation of the IMU frame $\left\{\mathit{i}\right\}$
with respect to the world frame $\left\{\mathit{w}\right\}$, the gyroscope
biases $\boldsymbol{b}_{g}$ and the accelerometer biases $\boldsymbol{b}_{a}$.
\begin{equation}
\boldsymbol{x}_{I} =
\begin{bmatrix}
{\boldsymbol{p}_w^i}^T &
{\boldsymbol{v}_w^i}^T &
{\boldsymbol{q}_{w}^{i}}^T &
{\boldsymbol{b}_{g}}^{T} &
{\boldsymbol{b}_{a}}^{T}
\end{bmatrix}^{T}
\label{eq:state-vec-i}
\end{equation}

The vision states $\boldsymbol{x}_{V} \in \mathbb{R}^{7M+3N}$ can be divided
into two subsets
\begin{equation}
\boldsymbol{x}_V = \begin{bmatrix}{\boldsymbol{x}_S}^T &
{\boldsymbol{x}_F}^{T}\end{bmatrix}^{T}
\label{eq:state-vec-sf}
\end{equation}
where $\boldsymbol{x}_S \in \mathbb{R}^{7M}$ are the \emph{sliding window}
states given by
\begin{equation}
\boldsymbol{x}_S =
{\begin{bmatrix}
{\boldsymbol{p}_w^{c_1}}^T & ... & {\boldsymbol{p}_w^{c_M}}^T &
{\boldsymbol{q}_{w}^{c_1}}^T & ... & {\boldsymbol{q}_{w}^{c_M}}^T
\end{bmatrix}}^T
\label{eq:state-vec-s}
\end{equation}
and $\boldsymbol{x}_F \in \mathbb{R}^{3N}$ are the \emph{feature} states given
by
\begin{equation}
\boldsymbol{x}_F =
{\begin{bmatrix}
{\boldsymbol{f}_{1}}^T & ... & {\boldsymbol{f}_{N}}^T
\end{bmatrix}}^T\;.
\label{eq:state-vec-f}
\end{equation}
The sliding window states $\boldsymbol{x}_S$ include the
positions $\left\{\boldsymbol{p}_w^{c_i}\right\}_{i}$ and the orientations
$\left\{\boldsymbol{q}_{w}^{c_i}\right\}_{i}$ of the camera frame
at the last $M$ image time instances. They are involved in SLAM as well as in
MSCKF updates.
The feature states $\boldsymbol{x}_F$ includes the 3D
coordinates of $N$ features $\left\{\boldsymbol{f}_{j}\right\}_{j}$. They are only
involved in SLAM updates.

\subsection{Error States}
\label{sub:error-states}

Similarly to the state vector $\boldsymbol{x}$, the error state vector
$\delta\boldsymbol{x}$ can be broken into inertial and vision error states as
\begin{equation}
\delta\boldsymbol{x} = \begin{bmatrix}
{\delta\boldsymbol{x}_{I}}^T & {\delta\boldsymbol{x}_{V}}^{T}
\end{bmatrix}^{T}.
\label{eq:state-vec-iv}
\end{equation}

The inertial error states are
\begin{equation}
\delta\boldsymbol{x}_I = {\begin{bmatrix}
{\delta\boldsymbol{p}_w^i}^T &
{\delta\boldsymbol{v}_w^i}^T &
{\delta\boldsymbol{\theta}_w^i}^T &
{\delta\boldsymbol{b}_g}^{T} &
{\delta\boldsymbol{b}_a}^{T}
\end{bmatrix}}^T.
\label{eq:error-state-inertial}
\end{equation}
The position error is defined as the difference between the true and estimated
vector $\delta\boldsymbol{p}_w^i = \boldsymbol{p}_w^i
- \hat{\boldsymbol{p}}_w^i \in \mathbb{R}^3$, and similarly for the velocity and
inertial bias errors. The error quaternion $\delta\boldsymbol{q}$ is defined by
$\boldsymbol{q}=\hat{\boldsymbol{q}}\otimes\delta\boldsymbol{q}$, where
$\otimes$ is the quaternion product. Using the small angle approximation, one can
write
$\delta\boldsymbol{q} \simeq \begin{bmatrix}
1 &
\frac{1}{2}{\delta\boldsymbol{\theta}}^{T}\end{bmatrix}^{T}$,
thus $\delta\boldsymbol{\theta} \in \mathbb{R}^3$ is a correct minimal representation of the
error quaternion. Hence, $\delta\boldsymbol{x}_I \in \mathbb{R}^{15}$ while
$\boldsymbol{x}_I \in \mathbb{R}^{16}$.

Likewise, the vision error states are expressed as
\begin{equation}
\delta\boldsymbol{x}_V =
\begin{bmatrix}
{\delta\boldsymbol{x}_S}^T &
{\delta\boldsymbol{x}_F}^{T}
\end{bmatrix}^{T},
\end{equation}
where
\begin{align}
\delta\boldsymbol{x}_S &=
{\begin{bmatrix}
{\delta\boldsymbol{p}_w^{c_1}}^T & ... & {\delta\boldsymbol{p}_w^{c_M}}^T &
{\delta\boldsymbol{\theta}_{w}^{c_1}}^T & ... & {\delta\boldsymbol{\theta}_{w}^{c_M}}^T
\end{bmatrix}}^T
\label{eq:error-state-vec-s}\\
\delta\boldsymbol{x}_F &=
{\begin{bmatrix}
{\delta\boldsymbol{f}_{1}}^{T} & ... & {\delta\boldsymbol{f}_{N}}^{T}
\end{bmatrix}}^T.
\label{eq:error-state-vec-f}
\end{align}
Because of the size reduction in rotation error states,
$\delta\boldsymbol{x}_{V} \in \mathbb{R}^{6M+3N}$ while $\boldsymbol{x}_{V} \in
\mathbb{R}^{7M+3N}$.

\section{Inertial Propagation}

This section describes the propagation model, which is identical to
\cite{Weiss2011} for the inertial part.

\subsection{Continuous-Time Model}
\label{sub:continous-time-model}

Position, velocity, and attitude vary according to
\begin{equation}
\left \lbrace
\begin{array}{lcl}
\dot{\boldsymbol{p}}_w^i &=& \boldsymbol{v}_w^i \\
\dot{\boldsymbol{v}}_w^i &=& \boldsymbol{a}_w^i\\
\dot{\boldsymbol{q}}_w^i &=& \frac{1}{2}\boldsymbol{\Omega}
	(^i\boldsymbol{\omega}_w^i)\boldsymbol{q}_w^i
\end{array}\right..
\label{eq:state-model}
\end{equation}

The operator $\boldsymbol{\Omega}$ is defined by
\begin{align}
\boldsymbol{\Omega}(\boldsymbol{\omega})& = \begin{bmatrix}
0 & -\boldsymbol{\omega}^{T}\\
\boldsymbol{\omega} & -\lfloor\boldsymbol{\omega}\times\rfloor\end{bmatrix},
\label{eq:angular-velocity-operator}\\
\text{where}\;\lfloor\boldsymbol{\omega}\times\rfloor&=\begin{bmatrix}
0 & -\omega_{z} & \omega_{y}\\
\omega_{z} & 0 & -\omega_{x}\\
-\omega_{y} & \omega_{x} & 0\end{bmatrix}.
\label{eq:antisymmetric-matrix}
\end{align}

The definition of $\boldsymbol{\Omega}$ in
Equation~(\ref{eq:angular-velocity-operator}) depends on the convention used for
the quaternions: the first component is the scalar part, and the three others are
the imaginary parts.

The IMU measurements are modeled as
\begin{align}
\boldsymbol{\omega}_{IMU}& = {^i\boldsymbol{\omega}_w^i} +\boldsymbol{b}_g +
	\boldsymbol{n}_g
\label{eq:imu-gyro-measurement}\\
\boldsymbol{a}_{IMU}&=\boldsymbol{C}(\boldsymbol{q}{}_w^i)
	(\boldsymbol{a}_w^i - ^w\boldsymbol{g}) + \boldsymbol{b}_a
	+ \boldsymbol{n}_a
\label{eq:imu-accelero-measurement}
\end{align}
where $\boldsymbol{g}$ is the local gravity vector, $\boldsymbol{n}_{g}$ and
$\boldsymbol{n}_{a}$ are zero-mean Gaussian white noises. The dynamics of the
inertial biases $\boldsymbol{b}_g$ and $\boldsymbol{b}_a$ are modeled by random
walks, driven by zero-mean Gaussian white noises $\boldsymbol{n}_{b_g}$ and
$\boldsymbol{n}_{b_a}$, respectively.
\begin{equation}
\left \lbrace
\begin{array}{lcl}
\dot{\boldsymbol{b}}_{g} &=& \boldsymbol{n}_{b_g}\\
\dot{\boldsymbol{b}}_{a} &=& \boldsymbol{n}_{b_a}
\end{array}\right.
\label{eq:bias-dynamics}
\end{equation} 

By injecting Equations~(\ref{eq:imu-gyro-measurement}) and
(\ref{eq:imu-accelero-measurement}) in (\ref{eq:state-model}), and including
Equations~(\ref{eq:bias-dynamics}), we can now formulate the full inertial state
propagation model as
\begin{equation}
\left \lbrace
\begin{array}{lcl}
\dot{\boldsymbol{p}}_w^i &=& \boldsymbol{v}_w^i \\
\dot{\boldsymbol{v}}_w^i &=& \boldsymbol{C}(\boldsymbol{q}{}_w^i)^T
	(\boldsymbol{a}_{IMU} - \boldsymbol{b}_{a} - \boldsymbol{n}_{a})
	+ {^w\boldsymbol{g}} \\
\dot{\boldsymbol{q}}_w^i &=& \frac{1}{2}\boldsymbol{\Omega}
	(\boldsymbol{\omega}_{IMU} - \boldsymbol{b}_{g} - \boldsymbol{n}_{g})
	\boldsymbol{q}_w^i \\
\dot{\boldsymbol{b}}_{g} &=& \boldsymbol{n}_{b_g} \\
\dot{\boldsymbol{b}}_{a} &=& \boldsymbol{n}_{b_a}
\end{array}\right..
\label{eq:state-model2}
\end{equation}

Because the vision states refer either camera poses at given time instants, or
3D coordinates of terrain features which are assumed to be static, they have
zero dynamics, i.e.
~
\begin{equation}
\dot{\boldsymbol{x}}_V = \boldsymbol{0}\;.
\label{eq:vision-state-dynamics}
\end{equation}

\subsection{State Propagation}

The dynamics of the inertial state estimate
$\hat{\boldsymbol{x}}_{I} = E[\boldsymbol{x}_I]$ can be obtained by applying the
expectation operator $E$ to each of the terms of System~(\ref{eq:state-model})
\begin{equation}
\left \lbrace
\begin{array}{lcl}
\dot{\hat{\boldsymbol{p}}}_w^i &=& \hat{\boldsymbol{v}}_w^i \\
\dot{\hat{\boldsymbol{v}}}_w^i &=&
	{\boldsymbol{C}(\hat{\boldsymbol{q}}_w^i)}^{T}\hat{\boldsymbol{a}}
	+ ^w\boldsymbol{g} \\
\dot{\hat{\boldsymbol{q}}}_w^i &=&
	\frac{1}{2}\boldsymbol{\Omega}(\hat{\boldsymbol{\omega}})
	\hat{\boldsymbol{q}}_w^i \\
\dot{\hat{\boldsymbol{b}}}_g &=& \boldsymbol{0}_{3\times1} \\
\dot{\hat{\boldsymbol{b}}}_a &=& \boldsymbol{0}_{3\times1}
\end{array}\right.,
\label{eq:state-propagation}
\end{equation}
where $\hat{\boldsymbol{a}}=\boldsymbol{a}_{IMU}-\hat{\boldsymbol{b}}_a$ and
$\hat{\boldsymbol{\omega}}=\boldsymbol{\omega}_{IMU} - \hat{\boldsymbol{b}}_g$. 
These equations are integrated at IMU rate at first order to propagate the
inertial state estimation through time~\citep{Trawny2005}. This process can be
viewed as dead reckoning.

The vision state estimates do not change during inertial propagation
\begin{equation}
\dot{\hat{\boldsymbol{x}}}_V = \boldsymbol{0}\;.
\label{eq:vision-state-propagation}
\end{equation}

\subsection{Covariance Propagation}

System~(\ref{eq:state-model}) can be reformulated with a nonlinear function
$\boldsymbol{f}$ such that
\begin{equation}
\begin{array}{lcl}
\dot{\boldsymbol{x}}_I=\boldsymbol{f}(\boldsymbol{x}_I, \boldsymbol{n}_{IMU})
\end{array}
\label{eq:state-equation}
\end{equation}
with $\boldsymbol{n}_{IMU}=\begin{bmatrix}
		{\boldsymbol{n}_a}^{T} &
	 	{\boldsymbol{n}_{b_a}}^{T} &
	 	{\boldsymbol{n}_g}^{T} &
	 	{\boldsymbol{n}_{b_g}}^{T}	
	 	\end{bmatrix}^{T} \in \mathbb{R}^{12}$ the process noise coming
from the IMU measurements. The function $\boldsymbol{f}$ is time-dependent
since depends on inertial measurements that change with time.

$\boldsymbol{f}$ can be linearized with respect to the estimated inertial
state vector $\hat{\boldsymbol{x}}_{I}$ so that the dynamics of the error state
$\delta\boldsymbol{x}_I$ becomes
\begin{equation}
\dot{\delta\boldsymbol{x}}_I =
	  \boldsymbol{F}_c\delta\boldsymbol{x}_I
	+ \boldsymbol{G}_c\boldsymbol{n}_{IMU}\;.
\label{eq:linearized-state-equation-continuous}
\end{equation}
In discrete time, Equation~(\ref{eq:linearized-state-equation-continuous}) can
be written as
\begin{equation}
\delta\boldsymbol{x}_{I_{k+1}} =
	  \boldsymbol{F}_d\delta\boldsymbol{x}_{I_k}
	+ \boldsymbol{G}_d\boldsymbol{n}_{_k}\;,
\label{eq:linearized-state-equation-discrete}
\end{equation}
where $\boldsymbol{n}_{_k}$ is a zero-mean white Gaussian noise vector with
covariance matrix
$\boldsymbol{Q}_d  = diag(
\sigma_{n_a}^2\boldsymbol{I}_3, 
\sigma_{n_{b_a}}^2\boldsymbol{I}_3, 
\sigma_{n_b}^2\boldsymbol{I}_3,
\sigma_{n_{b_g}}^2\boldsymbol{I}_3)$.
We refer the reader to \cite{Weiss2011} for the expressions of
$\boldsymbol{F}_d$ and $\boldsymbol{G}_d$ implemented in xEKF.

Likewise, from Equation~(\ref{eq:vision-state-dynamics}), the vision error
state dynamics is
\begin{equation}
\dot{\delta\boldsymbol{x}}_V = \boldsymbol{0}\;,
\label{eq:error-vision-state-dynamics}
\end{equation}
which can be written in discrete time as
\begin{equation}
\delta\boldsymbol{x}_{V_{k+1}} = \delta\boldsymbol{x}_{V_k}\;.
\label{eq:error-vision-state-dynamics-discrete}
\end{equation}
 
The error covariance matrix can be divided in four blocks
\begin{equation}
\boldsymbol{P}_{k|k} = \begin{bmatrix}
\boldsymbol{P}_{II_{k|k}} & \boldsymbol{P}_{IV_{k|k}} \\
{\boldsymbol{P}_{IV_{k|k}}}^T & \boldsymbol{P}_{VV_{k|k}}
\end{bmatrix},
\label{eq:error-covariance-matrix}
\end{equation}
where $\boldsymbol{P}_{II_{k|k}}\in\mathbb{R}^{15\times15}$ is the covariance of
the inertial error state vector, and
$\boldsymbol{P}_{VV_{k|k}}\in\mathbb{R}^{(6M+3N)\times(6M+3N)}$ is that of the
vision error states. $\boldsymbol{P}_{II}$ is propagated between $t_k$ and $t_{k+1}$ according to
\begin{equation}
\boldsymbol{P}_{II_{k+1|k}} =
\boldsymbol{F}_d\boldsymbol{P}_{II_{k|k}}{\boldsymbol{F}_d}^T +
\boldsymbol{G}_d\boldsymbol{Q}_d\boldsymbol{G}_d^T\;.
\label{eq:covariance-prop1}
\end{equation}
Covariance block $\boldsymbol{P}_{IV}$ propagates as
\begin{align}
\boldsymbol{P}_{IV_{k+1|k}}
&= E[\delta\boldsymbol{x}_{I_{k+1}}\delta\boldsymbol{x}_{V_{k+1}}^T] \\
&= E[(\boldsymbol{F}_d\delta\boldsymbol{x}_{I_k} + \boldsymbol{G}_d\boldsymbol{n}_{_k})\delta\boldsymbol{x}_{V_k}^T] \\
&= E[\boldsymbol{F}_d\delta\boldsymbol{x}_{I_k}\delta\boldsymbol{x}_{V_k}^T] \\
&= \boldsymbol{F}_d E[\delta\boldsymbol{x}_{I_k}\delta\boldsymbol{x}_{V_k}^T] \\
&= \boldsymbol{F}_d \boldsymbol{P}_{IV_{k|k}}\;.
\label{eq:covariance-prop2}
\end{align}
Likewise for the vision error covariance block
\begin{align}
\boldsymbol{P}_{VV_{k+1|k}}
&= E[\delta\boldsymbol{x}_{V_{k+1}}\delta\boldsymbol{x}_{V_{k+1}}^T] \\
&= E[\delta\boldsymbol{x}_{V_k}\delta\boldsymbol{x}_{V_k}^T] \\
&= \boldsymbol{P}_{VV_{k|k}}\;.
\label{eq:covariance-prop3}
\end{align}

\section{Visual Update}
\label{sec:vision-update}

This section describes the two paradigms which can be used to construct a filter
update\footnote{Before either of these updates is applied, we should note that
the innovation is checked for outliers through a $\chi^2$ test with
$95\%$~confidence.} from an image feature with unknown 3D coordinates: SLAM and
MSCKF. In both, the model for the image
measurement~$^i\boldsymbol{z}_j \in \mathbb{R}^2$ of feature~$\boldsymbol{p}_j$
observed in camera~$\left\{c_i\right\}$ is the normalized projection in the
plane~$^{c_i}z=1$
\begin{equation}
^i\boldsymbol{z}_j =
\frac{1}{^{c_i}z_j}
\begin{bmatrix}^{c_i}x_j\\^{c_i}y_j\end{bmatrix}
+ {^i\boldsymbol{n}_j}\;,
\label{eq:vision-measurement}
\end{equation}
where
\begin{align}
^{c_i}\boldsymbol{p}_j &= \begin{bmatrix}
^{c_i}x_j & ^{c_i}y_j & ^{c_i}z_j \end{bmatrix}^T
\label{eq:feature-cart-coord}\\
&= \boldsymbol{C}(\boldsymbol{q}_w^{c_i})({^w\boldsymbol{p}_j}
- \boldsymbol{p}_w^{c_i})
\label{eq:feature-frame-tf1}
\end{align}
and $^i\boldsymbol{n}_j$ is a zero-mean white Gaussian measurement noise with
covariance matrix $^i\boldsymbol{R}_j = \sigma_V^2\boldsymbol{I}_2$. The standard
deviation $\sigma_V$ of the normalized feature noise $^i\boldsymbol{n}_j$ is
function of the visual front-end performance. We assume it is uniform throughout
the image, as well as identical in both image dimensions. This defines a
nonlinear visual measurement function $\boldsymbol{h}$ such that
\begin{equation}
^i\boldsymbol{z}_j =
\boldsymbol{h}(\boldsymbol{p}_w^{c_i},
\boldsymbol{q}_w^{c_i},
{^w\boldsymbol{p}_j})
+ {^i\boldsymbol{n}_j}
\;.
\label{eq:vision-measurement-h}
\end{equation}
In practice, $^i\boldsymbol{z}_j$ can obtained for any image measurement coming
from a camera calibrated for its pinhole model and distortions.

\subsection{SLAM Update}
\label{sub:slam-update}

The SLAM paradigm can provide a filter update from each observation of a visual
feature $\boldsymbol{p}_j$ corresponding to one of the feature states
$\boldsymbol{f}_j$. 

For a given reference frame $\left\{ r \right\}$, the
cartesian coordinates of $\boldsymbol{p}_j$ in world frame $\left\{ w \right\}$
can be expressed as
\begin{align}
^w\boldsymbol{p}_j &=
\begin{bmatrix} ^wx_j & ^wy_j & ^wz_j \end{bmatrix}^T
\label{eq:inverse-depth-param1}\\
 &= \boldsymbol{p}_w^r +
\frac{1}{\rho_j} \boldsymbol{C}(\boldsymbol{q}_w^r)^T
\begin{bmatrix} \alpha_j \\ \beta_j \\ 1\end{bmatrix}
\label{eq:inverse-depth-param2}
\end{align}
where
$\boldsymbol{f}_j = \begin{bmatrix}
\alpha_j &
\beta_j &
\rho_j
\end{bmatrix}^{T}$
represents the \emph{inverse-depth} parameters of $\boldsymbol{p}_j$ with
respect to the \emph{anchor} pose $\left\{ r \right\}$. The inverse-depth
parametrization has been used to represent feature coordinates in SLAM due to
its improved depth convergence properties~\citep{Civera2008}.

\begin{figure}
  \centering
  \includegraphics[width=0.8\textwidth]{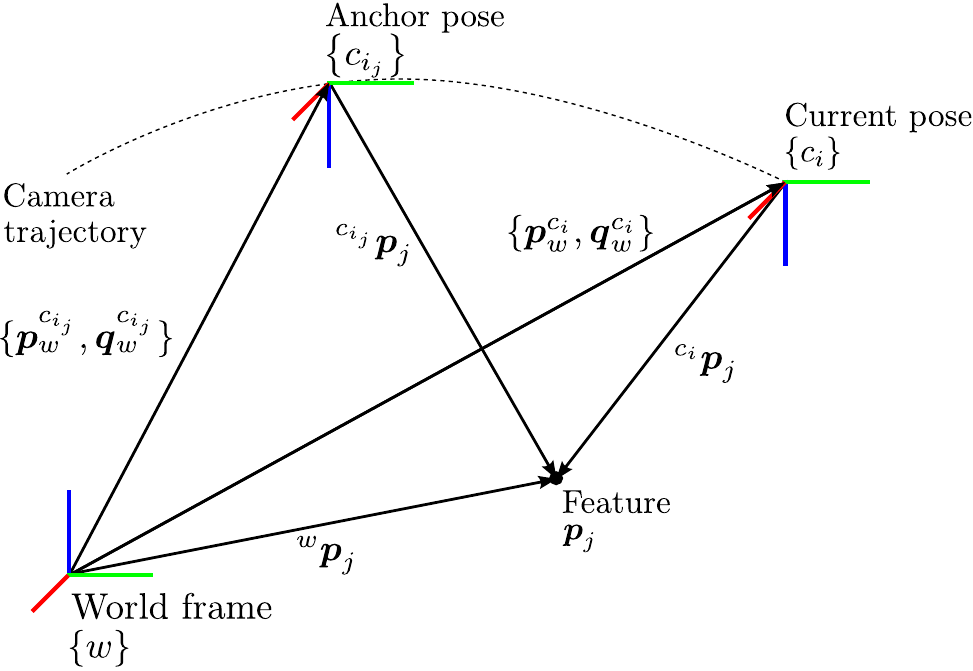}
  \caption{Geometry of SLAM visual measurement of feature $\boldsymbol{p}_j$
  	observed in camera $\left\{ c_i \right\}$, using inverse-depth
  	parametrization with respect to anchor camera $\left\{ c_{i_j} \right\}$ in
  	the sliding window.}
  \label{fig:slam-measurement}
\end{figure}

In xVIO, the anchor is one of the sliding-window states
$\left\{ c_{i_j} \right\}$, i.e. the frame defined by the position
$\boldsymbol{p}_w^{c_{i_j}}$ and orientation $\boldsymbol{q}_{w}^{c_{i_j}}$ at
index $i_j$ in the sliding window. This defines the geometry of the SLAM
measurement of feature $\boldsymbol{p}_j$ in camera $\left\{ c_i \right\}$
illustrated in Figure~\ref{fig:slam-measurement}.

By letting $\left\{ r \right\} = \left\{ c_{i_j} \right\}$ in
Equation~(\ref{eq:inverse-depth-param2}) and injecting it into
Equation~(\ref{eq:feature-frame-tf1}), we can write 
\begin{equation}
^{c_i}\boldsymbol{p}_j = \boldsymbol{C}(\boldsymbol{q}_w^{c_i})
\Bigg(\boldsymbol{p}_w^{c_{i_j}} +
\frac{1}{\rho_j} \boldsymbol{C}(\boldsymbol{q}_w^{c_{i_j}})^T
\begin{bmatrix} \alpha_j \\ \beta_j \\ 1\end{bmatrix}
- \boldsymbol{p}_w^{c_i}\Bigg)
\label{eq:feature-frame-tf2}
\end{equation}
such that the visual measurement model of Equation~(\ref{eq:vision-measurement-h})
can be linearized as
\begin{equation}
^i\delta\boldsymbol{z}_{j} \simeq
\boldsymbol{H}_{\boldsymbol{p}_{i_j}} \delta\boldsymbol{p}_w^{c_{i_j}} +
\boldsymbol{H}_{\boldsymbol{p}_i} \delta\boldsymbol{p}_w^{c_i} +
\boldsymbol{H}_{\boldsymbol{\theta}_{i_j}} \delta\boldsymbol{\theta}_w^{c_{i_j}} +
\boldsymbol{H}_{\boldsymbol{\theta}_i} \delta\boldsymbol{\theta}_w^{c_i} +
\boldsymbol{H}_{\boldsymbol{f}_j} \delta\boldsymbol{f}_j +
{^i\boldsymbol{n}_j}
\label{eq:slam-linearization}
\end{equation}
with
\begin{align}
\boldsymbol{H}_{\boldsymbol{p}_{i_j}} & =
{^i\boldsymbol{J}_j} \boldsymbol{C}(\hat{\boldsymbol{q}}_w^{c_i})
\label{eq:slam-jacobian-p-i0}\\
\boldsymbol{H}_{\boldsymbol{p}_i} & =
- {^i\boldsymbol{J}_j}\boldsymbol{C}(\hat{\boldsymbol{q}}_w^{c_i})
\label{eq:slam-jacobian-p-i}\\
\boldsymbol{H}_{\boldsymbol{\theta}_{i_j}} & = - \frac{1}{\hat{\rho}_j}
{^i\boldsymbol{J}_j}
\boldsymbol{C}(\hat{\boldsymbol{q}}_w^{c_i})
\boldsymbol{C}(\hat{\boldsymbol{q}}_w^{c_{i_j}})^T
\left\lfloor\begin{bmatrix}
\hat{\alpha}_j \\
\hat{\beta}_j \\
1\end{bmatrix} \times\right\rfloor
\label{eq:slam-jacobian-theta-i0}\\
\boldsymbol{H}_{\boldsymbol{\theta}_i} & =
{^i\boldsymbol{J}_j} \left\lfloor \boldsymbol{C}(\hat{\boldsymbol{q}}_w^{c_i}) \Bigg(
\hat{\boldsymbol{p}}_w^{c_{i_j}}
- \hat{\boldsymbol{p}}_w^{c_i}
+ \frac{1}{\hat{\rho}_j} \boldsymbol{C}(\hat{\boldsymbol{q}}_w^{c_{i_j}})^T
\begin{bmatrix} \hat{\alpha}_j \\ \hat{\beta}_j \\ 1\end{bmatrix}
\Bigg) \times\right\rfloor
\label{eq:slam-jacobian-theta-i}\\
\boldsymbol{H}_{\boldsymbol{f}_j} & = 
\frac{1}{\hat{\rho}_j} {^i\boldsymbol{J}_j}
\boldsymbol{C}(\hat{\boldsymbol{q}}_w^{c_i})
\boldsymbol{C}(\hat{\boldsymbol{q}}_w^{c_{i_j}})^T
\begin{bmatrix}
1 & 0 & -\frac{\hat{\alpha}_j}{\hat{\rho}_j} \\
0 & 1 & -\frac{\hat{\beta}_j}{\hat{\rho}_j} \\
0 & 0 & -\frac{1}{\hat{\rho}_j}
\end{bmatrix}
\label{eq:slam-jacobian-feature}\\
\text{and}\;
^i\boldsymbol{J}_j & = \frac{1}{^{c_i}\hat{z}_j}\begin{bmatrix}\boldsymbol{I}_{2} & - \hat{\boldsymbol{z}}_{j}\end{bmatrix}.
\label{eq:msckf-jacobian-prefix}
\end{align}
These Jacobians are derived in Section~\ref{sec:slam-meas-jac-der}.

Note that when $i = i_j$, $^i\delta\boldsymbol{z}_{j} =
{^{i_j}\delta\boldsymbol{z}_{j}} =
\begin{bmatrix} \delta\alpha_j \\ \delta\beta_j \end{bmatrix}$.
Since SLAM measurements are applied at every frame, they correspond to the
most recent pose in the sliding window, so $i = 1$.

\subsection{MSCKF Update}
\label{sub:msckf-update}

\begin{figure}
  \centering
  \includegraphics[width=0.8\textwidth]{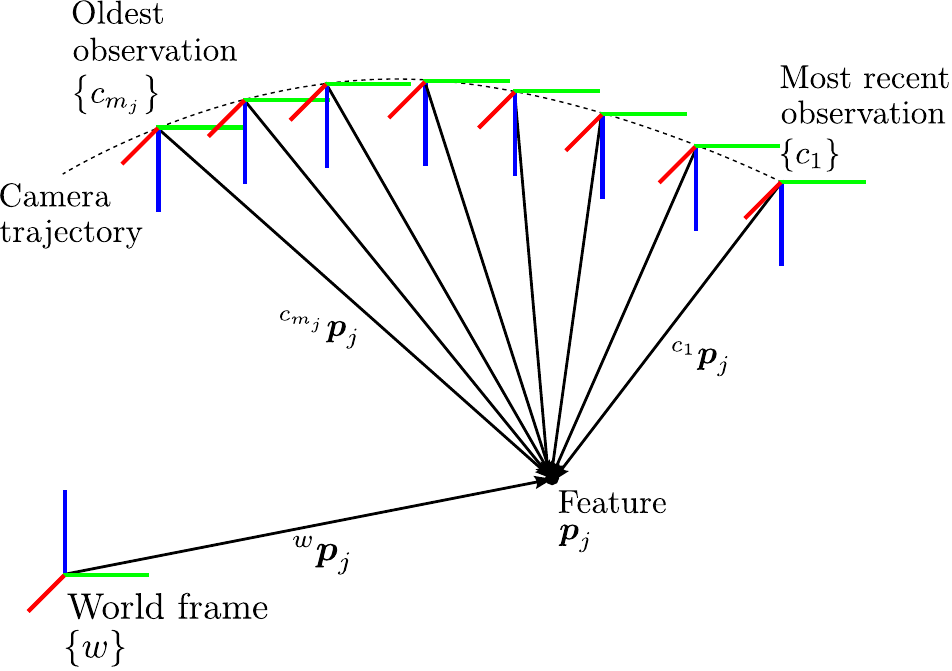}
  \caption{Geometry of MSCKF visual measurements of feature
  	$\boldsymbol{p}_j$, tracked continuously between the most recent camera pose
  	$\left\{ c_1 \right\}$ and camera pose $\left\{ c_{m_j} \right\}$ in the
  	sliding window.}
  \label{fig:msckf-measurement}
\end{figure}

MSCKF's multi-state constraints can provide an update from a feature that has been
observed over the last $m$ images, as illustrated in
Figure~\ref{fig:msckf-measurement}. The $m$ measurements are processed as a
batch when the track is lost, or its length exceeds the size of the sliding
window. MSCKF was first proposed by \citet{Mourikis2007} to reduce the
computational cost per feature with respect to SLAM, since the features do not
need to be included in the state any more. MSCKF requires the following
conditions though:
\begin{enumerate}
	\item $m \leq M$: each image measurement must have a corresponding pose
		  state in the sliding window.
    \item $m \geq 2$: each feature must be triangulated from the prior poses, so
    	      it must be visible in at least two images.
    	\item $\Delta p^c \geq \Delta p^c_{min} > 0$: there must be a minimum
    	camera translation between the images for feature triangulation\footnote{In
    	presence of noise, $\Delta p^c_{min}$ is tuned for numerical stability.}.
\end{enumerate}

Overall, MSCKF vs. SLAM can be seen as a trade-off since the
computational cost is still cubic in $M$, the size of the sliding window, in
MSCKF; while it is cubic in $N$, the number of feature states in SLAM.
Additionally, while MSCKF can process more features for a given computational
cost, SLAM can provide a update at image rate on each of these features and
does not have to wait until the corresponding tracks ends. Finally, SLAM does
not have a minimum translation requirement, which means that it can constrain
the state even if the sensor platform is not moving. For these reasons, xVIO
uses both in a hybrid fashion, as proposed by~\citet{Li2012}.
  
Let us start by considering one visual measurement where prior estimates are
available for the camera position, orientation, and the 3D coordinates of the
feature. Using Equation~(\ref{eq:vision-measurement-h}), we can form the measurement
prediction
${^i\hat{\boldsymbol{z}}_{j}} =
\boldsymbol{h}(\hat{\boldsymbol{p}}_w^{c_i},
\hat{\boldsymbol{q}}_w^{c_i},
{^w\hat{\boldsymbol{p}}_j})$
. The measurement innovation
$^i\delta\boldsymbol{z}_{j} =
{^i\boldsymbol{z}_{j}} - {^i\hat{\boldsymbol{z}}_{j}}$ can then be linearized as

\begin{equation}
^i\delta\boldsymbol{z}_{j} \simeq
{^{(i,j)}\boldsymbol{H}}_{\boldsymbol{p}} \delta\boldsymbol{p}_w^{c_i} +
{^{(i,j)}\boldsymbol{H}}_{\boldsymbol{\theta}} \delta\boldsymbol{\theta}_w^{c_i} +
{^i\boldsymbol{H}}_{\boldsymbol{p}_j} {^w\delta\boldsymbol{p}}_j +
{^i\boldsymbol{n}_j}\;,
\label{eq:msckf-linearization1}
\end{equation}
\begin{align}
\text{with}\;
{^{(i,j)}\boldsymbol{H}}_{\boldsymbol{p}} & =
- {^i\boldsymbol{J}_j}\boldsymbol{C}(\hat{\boldsymbol{q}}_w^{c_i})\;,
\label{eq:msckf-jacobian-p}\\
{^{(i,j)}\boldsymbol{H}}_{\boldsymbol{\theta}} & =
{^i\boldsymbol{J}_j}\lfloor\boldsymbol{C}(\hat{\boldsymbol{q}}_w^{c_i})({^w\hat{\boldsymbol{p}}_j}
- \hat{\boldsymbol{p}}_w^{c_i})\times\rfloor,
\label{eq:msckf-jacobian-theta}\\
{^i\boldsymbol{H}}_{\boldsymbol{p}_j} & = {^i\boldsymbol{J}_j}\boldsymbol{C}(\hat{\boldsymbol{q}}_w^{c_i})\;,
\label{eq:msckf-jacobian-feature}\\
\text{and}\;
^i\boldsymbol{J}_j & = \frac{1}{^{c_i}\hat{z}_j}\begin{bmatrix}\boldsymbol{I}_{2} & - \hat{\boldsymbol{z}}_{j}\end{bmatrix}.
\label{eq:msckf-jacobian-prefix}
\end{align}

If ${^i\boldsymbol{z}_{j}}$ is a MSCKF measurement, by definition
$\delta\boldsymbol{p}_w^{c_i}$ and $\delta\boldsymbol{\theta}_w^{c_i}$ are part
of the error state vector $\delta\boldsymbol{x}$ defined in
Equation~(\ref{eq:state-vec-iv}), but not ${^w\delta\boldsymbol{p}}_j$.
Thus Equation~(\ref{eq:msckf-linearization1}) can be written
\begin{equation}
^i\delta\boldsymbol{z}_{j} \simeq
{^i\boldsymbol{H}}_{j} \delta\boldsymbol{x} +
{^i\boldsymbol{H}}_{\boldsymbol{p}_j} {^w\delta\boldsymbol{p}}_j +
{^i\boldsymbol{n}_j}\;,
\label{eq:ekf-linearization2}
\end{equation}
\begin{equation}
\text{with}\;{^i\boldsymbol{H}}_{j} = \begin{bmatrix}
\boldsymbol{0}_{2\times15} & ... &
{^{(i,j)}\boldsymbol{H}}_{\boldsymbol{p}} & ... &
{^{(i,j)}\boldsymbol{H}}_{\boldsymbol{\theta}} & ... &
\boldsymbol{0}_{2\times3N}
\end{bmatrix}.
\label{eq:meas-jacobian-total}
\end{equation}

By stacking up the residuals $^i\delta\boldsymbol{z}_{j}$ for each of the $m_j$
observations of feature $\boldsymbol{p}_j$ $(m_j \leq M)$, one can form the
overall residual $\delta\boldsymbol{z}_{j} \in \mathbb{R}^{2m_j}$
\begin{equation}
\delta\boldsymbol{z}_{j} \simeq
{\boldsymbol{H}}_{j} \delta\boldsymbol{x} +
{\boldsymbol{H}}_{\boldsymbol{p}_j} {^w\delta\boldsymbol{p}}_j +
{\boldsymbol{n}_j}\;,
\label{eq:residual-per-feature}
\end{equation}
where $\boldsymbol{n}_j \in \mathbb{R}^{2m_j\times2m_j}$ has covariance
$\boldsymbol{R}_j = \sigma_V^2\boldsymbol{I}_{2m_j}$.

Equation~(\ref{eq:residual-per-feature}) cannot be applied as an EKF update
because the ${^w\delta\boldsymbol{p}}_j$ term can neither be formulated as a
linear combination of the states, nor integrated in the noise. To correct for
that, we can multiply on each side by the left nullspace of
${\boldsymbol{H}}_{\boldsymbol{p}_j}$, denoted as $\boldsymbol{A}_j$, so that
\begin{equation}
\delta\boldsymbol{z}_{0_j} = \boldsymbol{A}_j^T \delta\boldsymbol{z}_{j} \simeq
\boldsymbol{A}_j^T {\boldsymbol{H}}_{j} \delta\boldsymbol{x} +
\boldsymbol{A}_j^T {\boldsymbol{n}_j} = 
{\boldsymbol{H}}_{0_j} \delta\boldsymbol{x} +
\boldsymbol{n}_{0_j}\;.
\label{eq:projected-residual-per-feature}
\end{equation}
From Equation~(\ref{eq:msckf-jacobian-feature}), one can verify that
${\boldsymbol{H}}_{\boldsymbol{p}_j} \in \mathbb{R}^{2m_j\times3}$ has full
column rank. Per the rank-nullity theorem, its left nullspace has dimension
$2m_j - 3$, thus $\boldsymbol{A}_j \in \mathbb{R}^{(2m_j-3)\times2m_j}$ and
$\delta\boldsymbol{z}_{0_j} \in \mathbb{R}^{2m_j-3}$.

Note that to perform this update, an estimate of $^w\boldsymbol{p}_j$ must be known
beforehands and is triangulated\footnote{This triangulation is the origin of
MSCKF requirements 2 and 3, presented at the beginning of this subsection.}
using the sliding window state priors as described in \citet{Mourikis2007}.
We also apply the QR decomposition of matrix $\boldsymbol{H}_0$, resulting from
stacking up all the MSCKF measurements per feature of
Equation~(\ref{eq:projected-residual-per-feature}) together as described in
\citet{Mourikis2007}.

\section{Range-Visual Update}
\label{sec:range-update}

\begin{figure}
  \centering
  \includegraphics[width=0.6\textwidth]{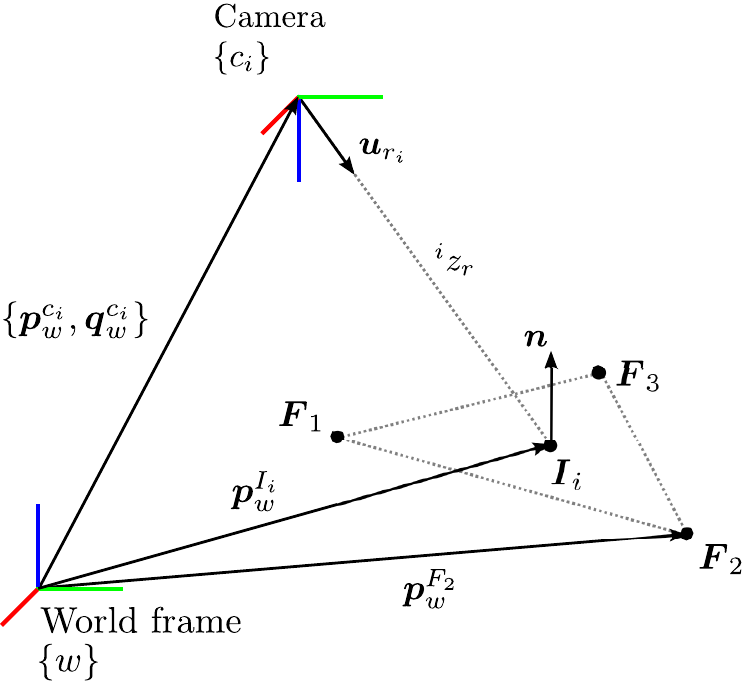}
  \caption{Geometry of the range measurement ${^i{z}}_r$ at time $i$.
  	$\boldsymbol{F}_1$, $\boldsymbol{F}_2$ and $\boldsymbol{F}_3$ are three
	SLAM features in the scene. $\boldsymbol{u}_{r_i}$ is the unit vector oriented
	along the measurement axis of the LRF, which is assumed to have the same origin
	as the camera frame $\left\{c_i\right\}$. $\boldsymbol{I}_i$ is the
	intersection point of this axis with the scene.}
  \label{fig:rv-meas-3d}
\end{figure}

Terrain range measurement models which are used alongside visual measurements
typically assumed the ground is \emph{globally-flat}~\citep{Bayard2019}. Instead,
we assume that the terrain is only \emph{locally-flat} between three SLAM
features surrounding the terrain point with respect to which the range is
measured. We also assume zero translation between the optical center of the
camera and the origin of the \emph{Laser Range Finder} (LRF)\footnote{This
assumption can be removed by measuring the position of the LRF in camera
frame.}. Figure~\ref{fig:rv-meas-3d} illustrates these assumptions.

$\boldsymbol{u}_r$ is the unit vector oriented along the optical axis of the
LRF. $\boldsymbol{p}_I$ is the intersection of this axis with the terrain.
$\boldsymbol{p}_{j_1}$, $\boldsymbol{p}_{j_2}$ and $\boldsymbol{p}_{j_3}$ are
SLAM features forming a triangle around $\boldsymbol{p}_I$ in image space.
$\boldsymbol{n}$ is a normal vector to the plane containing
$\boldsymbol{p}_{j_1}$, $\boldsymbol{p}_{j_2}$, $\boldsymbol{p}_{j_3}$ and
$\boldsymbol{p}_I$. If ${{^w\boldsymbol{u}_{r_i}}^T}{^w\boldsymbol{n}} \neq 0$, we can express the range measurement
at time $i$ as
\begin{align}
^i{z}_r &=
{^i{z}_r} \frac{{{^w\boldsymbol{u}_{r_i}}^T}{^w\boldsymbol{n}}}
	{{{^w\boldsymbol{u}_{r_i}}^T}{^w\boldsymbol{n}}}
\label{eq:rv-meas-1}\\
&=
\frac{(\boldsymbol{p}_w^I - \boldsymbol{p}_w^{c_i})^T{^w\boldsymbol{n}}}
	{{{^w\boldsymbol{u}_{r_i}}^T}{^w\boldsymbol{n}}}
\label{eq:rv-meas-2}\\
&=
\frac{(\boldsymbol{p}_w^I - \boldsymbol{p}_w^{f_{j_2}} + \boldsymbol{p}_w^{f_{j_2}} - \boldsymbol{p}_w^{c_i})^T{^w\boldsymbol{n}}}
	{{{^w\boldsymbol{u}_{r_i}}^T}{^w\boldsymbol{n}}}
\label{eq:rv-meas-3}
\end{align}
where
\begin{equation}
^w\boldsymbol{n} = (\boldsymbol{p}_w^{f_{j_1}} - \boldsymbol{p}_w^{f_{j_2}})
	\times(\boldsymbol{p}_w^{f_{j_3}} - \boldsymbol{p}_w^{f_{j_2}})\;.
\label{eq:facet-normal}
\end{equation}

Since $(\boldsymbol{p}_w^I - \boldsymbol{p}_w^{f_{j_2}})^T{^w\boldsymbol{n}} = 0$, that means
\begin{equation}
^i{z}_r =
\frac{(\boldsymbol{p}_w^{f_{j_2}} - \boldsymbol{p}_w^{c_i})^T{^w\boldsymbol{n}}}
	{{{^w\boldsymbol{u}_{r_i}}^T}{^w\boldsymbol{n}}}
\label{eq:rv-meas-4}
\end{equation}

Note that $^w\boldsymbol{n}$ is not a unit
vector in general, and a Mahalanobis distance test is performed at the filter
level to detect range outliers.

Equation~(\ref{eq:rv-meas-3}) also shows the range is a nonlinear function of the state
\begin{equation}
^iz_r =
\boldsymbol{h}_r(\boldsymbol{x}) + ^in_r
\label{eq:rv-nonlin}
\end{equation}
which can be linearized to update an EKF as
\begin{equation}
^i\delta z_r \simeq
\begin{aligned}
&
\boldsymbol{H}_{\boldsymbol{p}_{i_1}} \delta\boldsymbol{p}_w^{c_{i_1}}
+ \boldsymbol{H}_{\boldsymbol{p}_{i_2}} \delta\boldsymbol{p}_w^{c_{i_2}}
+ \boldsymbol{H}_{\boldsymbol{p}_{i_3}} \delta\boldsymbol{p}_w^{c_{i_3}}
+ \boldsymbol{H}_{\boldsymbol{p}_i} \delta\boldsymbol{p}_w^{c_i}
\\
&
+ \boldsymbol{H}_{\boldsymbol{\theta}_{i_1}} \delta\boldsymbol{\theta}_w^{c_{i_1}}
+ \boldsymbol{H}_{\boldsymbol{\theta}_{i_2}} \delta\boldsymbol{\theta}_w^{c_{i_2}}
+ \boldsymbol{H}_{\boldsymbol{\theta}_{i_3}} \delta\boldsymbol{\theta}_w^{c_{i_3}}
+ \boldsymbol{H}_{\boldsymbol{\theta}_i} \delta\boldsymbol{\theta}_w^{c_i}
\\
&
+ \boldsymbol{H}_{\boldsymbol{f}_{j_1}} \delta\boldsymbol{f}_{j_1}
+ \boldsymbol{H}_{\boldsymbol{f}_{j_2}} \delta\boldsymbol{f}_{j_2}
+ \boldsymbol{H}_{\boldsymbol{f}_{j_3}} \delta\boldsymbol{f}_{j_3}
\\
&
+ ^in_r
\end{aligned}
\label{eq:rv-linearization-2}
\end{equation}
where
\begin{align}
\boldsymbol{H}_{\boldsymbol{p}_{i_1}} &= \boldsymbol{H}_{\boldsymbol{p}_{j_1}}
\label{eq:rv-hpi1}\\
\boldsymbol{H}_{\boldsymbol{p}_{i_2}} &= \boldsymbol{H}_{\boldsymbol{p}_{j_2}}
\label{eq:rv-hpi2}\\
\boldsymbol{H}_{\boldsymbol{p}_{i_3}} &= \boldsymbol{H}_{\boldsymbol{p}_{j_3}}
\label{eq:rv-hpi3}\\
\boldsymbol{H}_{\boldsymbol{p}_i} &= - \frac{1}{\hat{b}} ^w\hat{\boldsymbol{n}}^T
\label{eq:rv-hpi-2}\\
\boldsymbol{H}_{\boldsymbol{\theta}_{i_1}} &= - \frac{1}{\hat{\rho_{j_1}}} \boldsymbol{H}_{\boldsymbol{p}_{j_1}} 
\boldsymbol{C}(\hat{\boldsymbol{q}}_w^{c_{i_{j_1}}})^T
\left\lfloor \begin{bmatrix} \hat{\alpha_{j_1}} \\ \hat{\beta_{j_1}} \\ 1\end{bmatrix} \times \right\rfloor
\label{eq:rv-hthetai1}\\
\boldsymbol{H}_{\boldsymbol{\theta}_{i_2}} &= - \frac{1}{\hat{\rho_{j_2}}} \boldsymbol{H}_{\boldsymbol{p}_{j_2}} 
\boldsymbol{C}(\hat{\boldsymbol{q}}_w^{c_{i_{j_2}}})^T
\left\lfloor \begin{bmatrix} \hat{\alpha_{j_2}} \\ \hat{\beta_{j_2}} \\ 1\end{bmatrix} \times \right\rfloor
\label{eq:rv-hthetai2}\\
\boldsymbol{H}_{\boldsymbol{\theta}_{i_3}} &= - \frac{1}{\hat{\rho_{j_3}}} \boldsymbol{H}_{\boldsymbol{p}_{j_3}} 
\boldsymbol{C}(\hat{\boldsymbol{q}}_w^{c_{i_{j_3}}})^T
\left\lfloor \begin{bmatrix} \hat{\alpha_{j_3}} \\ \hat{\beta_{j_3}} \\ 1\end{bmatrix} \times \right\rfloor
\label{eq:rv-hthetai3}\\
\boldsymbol{H}_{\boldsymbol{\theta}_i} &= - \frac{\hat{a}}{\hat{b}^2} \left( \lfloor {^c\boldsymbol{u}_r} \times\rfloor \boldsymbol{C}\left(\hat{\boldsymbol{q}}_w^{c_i} \right) {^w\hat{\boldsymbol{n}}} \right)^T
\label{eq:rv-thetai-2}\\
\boldsymbol{H}_{\boldsymbol{f}_{j_1}} &= \frac{1}{\hat{\rho_{j_1}}} \boldsymbol{H}_{\boldsymbol{p}_{j_1}} \boldsymbol{C}(\hat{\boldsymbol{q}}_w^{c_{i_{j_1}}})^T
\begin{bmatrix} 1 & 0 & - \frac{\hat{\alpha_{j_1}}}{\hat{\rho_{j_1}}} \\ 0 & 1 & - \frac{\hat{\beta_{j_1}}}{\hat{\rho_{j_1}}} \\ 0 & 0& - \frac{1}{\hat{\rho_{j_1}}}\end{bmatrix}
\label{eq:rv-hfj1}\\
\boldsymbol{H}_{\boldsymbol{f}_{j_2}} &= \frac{1}{\hat{\rho_{j_2}}} \boldsymbol{H}_{\boldsymbol{p}_{j_2}} \boldsymbol{C}(\hat{\boldsymbol{q}}_w^{c_{i_{j_2}}})^T
\begin{bmatrix} 1 & 0 & - \frac{\hat{\alpha_{j_2}}}{\hat{\rho_{j_2}}} \\ 0 & 1 & - \frac{\hat{\beta_{j_2}}}{\hat{\rho_{j_2}}} \\ 0 & 0& - \frac{1}{\hat{\rho_{j_2}}}\end{bmatrix}
\label{eq:rv-hfj2}\\
\boldsymbol{H}_{\boldsymbol{f}_{j_3}} &= \frac{1}{\hat{\rho_{j_3}}} \boldsymbol{H}_{\boldsymbol{p}_{j_3}} \boldsymbol{C}(\hat{\boldsymbol{q}}_w^{c_{i_{j_3}}})^T
\begin{bmatrix} 1 & 0 & - \frac{\hat{\alpha_{j_3}}}{\hat{\rho_{j_3}}} \\ 0 & 1 & - \frac{\hat{\beta_{j_3}}}{\hat{\rho_{j_3}}} \\ 0 & 0& - \frac{1}{\hat{\rho_{j_3}}}\end{bmatrix}
\label{eq:rv-hfj3}
\end{align}
and
\begin{align}
\boldsymbol{H}_{\boldsymbol{p}_{j_1}} &= \frac{1}{\hat{b}}
\left( \left\lfloor(\hat{\boldsymbol{p}}_w^{f_{j_3}} - \hat{\boldsymbol{p}}_w^{f_{j_2}})\times\right\rfloor
\left(\hat{\boldsymbol{p}}_w^{f_{j_2}} - \hat{\boldsymbol{p}}_w^{I_i} \right) \right)^T
\label{eq:rv-hpj1-bis}\\
\boldsymbol{H}_{\boldsymbol{p}_{j_2}} &= \frac{1}{\hat{b}} \left( ^w\hat{\boldsymbol{n}} + 
\left\lfloor(\hat{\boldsymbol{p}}_w^{f_{j_1}} - \hat{\boldsymbol{p}}_w^{f_{j_3}})\times\right\rfloor
\left(\hat{\boldsymbol{p}}_w^{f_{j_2}} -\hat{\boldsymbol{p}}_w^{I_i} \right) \right)^T 
\label{eq:rv-hpj2-bis}\\
\boldsymbol{H}_{\boldsymbol{p}_{j_3}} &= \frac{1}{\hat{b}} \left( \left\lfloor(\hat{\boldsymbol{p}}_w^{f_{j_2}} - \hat{\boldsymbol{p}}_w^{f_{j_1}})\times\right\rfloor \left(\hat{\boldsymbol{p}}_w^{f_{j_2}} - \hat{\boldsymbol{p}}_w^{I_i} \right) \right)^T
\label{eq:rv-hpj3-bis}\\
a &= (\boldsymbol{p}_w^{f_{j_2}} - \boldsymbol{p}_w^{c_i})^T{^w\boldsymbol{n}}
\label{eq:rv-aterm}\\
b &= {{^w\boldsymbol{u}_{r_i}}^T}{^w\boldsymbol{n}}
\label{eq:rv-bterm}
\end{align}

To construct the range update in practice, we perform a Delaunay triangulation in image space
over the SLAM features, and select the triangle in which the intersection of the LRF beam
with the scene is located. We opted for the Delaunay triangulation since it maximizes the smallest angle of all possible
triangulations~\citep{gartner2013ETH}. This property avoids ``long and skinny" triangles that
do not provide strong local planar constraints.

\begin{figure}[h]
\centering
\includegraphics[width=0.92\columnwidth]{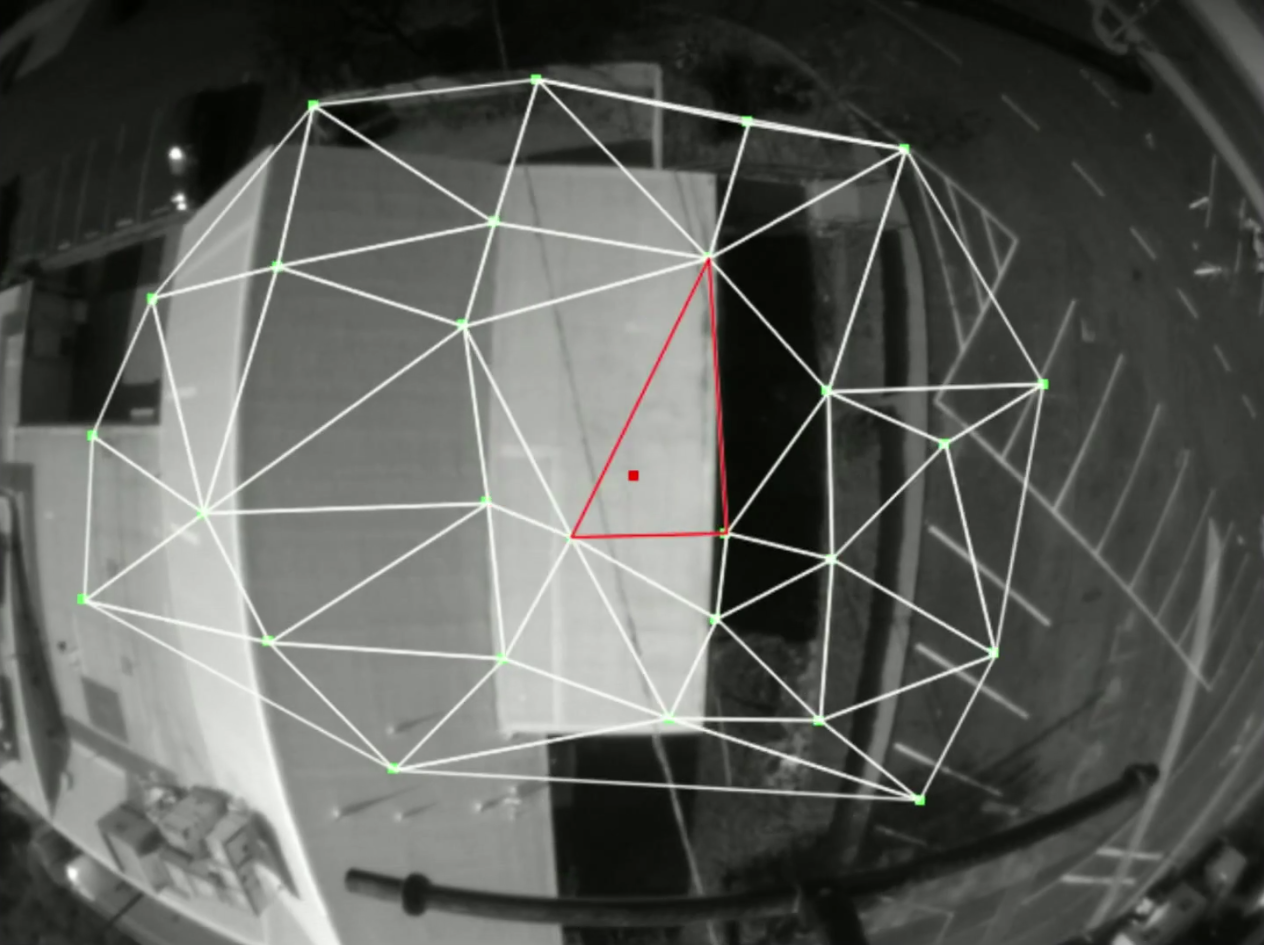}\\
\caption{Delaunay triangulation between image features tracked in the outdoor flight dataset.
The features used as filter states for SLAM form the corners of the triangles. The red dot
represents the intersection point of the LRF beam with the scene. The surrounding red triangle
is the ranged facet.}
\label{fig:delaunay}
\end{figure}

Figure~\ref{fig:delaunay} shows the Delaunay triangulation, and
the triangle selected as a ranged facet, over a sample image from our outdoor test sequence.
It also illustrates the partitioning of the scene into triangular facets, with SLAM features at their corners. Note
that that if the state estimator uses only 3 SLAM features in a lightweight fashion, then
this is equivalent to a globally-flat world assumption. Conversely, when the density of SLAM features
increases in the image, the limit of the area of the facets tends to zero and the scene assumption virtually
disappears.

\section{State Management}

The inertial state vector $\boldsymbol{x}_I$  and its error
covariance matrix $\boldsymbol{P}_{II}$ are initialized from values provided
by the user. After that, they both follow the propagation-update cycle
common to all Kalman filters. The management of the vision state vector
$\boldsymbol{x}_V$ and the rest of the error covariance matrix is less
straightforward though, and is the focus of this section.

\subsection{Sliding Window States}

The $M$ poses in the sliding window states $\boldsymbol{x}_S$ are managed in a
\emph{first-in first-out} fashion at image rate. Each time an image is acquired
with the camera, the following sliding window cycle happens in
$\boldsymbol{x}_S$:
\begin{enumerate}
	\item the oldest pose is removed;
	\item the remaining poses are shifted one slot, making the first slot empty;
	\item the current camera pose is inserted in the first slot.
\end{enumerate}
Equations~(\ref{eq:state-vec-s-tl}) and (\ref{eq:state-vec-s-tl+1}) illustrate
this sliding window process between two successives images $l$ and $l+1$
\begin{align}
\boldsymbol{x}_{S_l} &=
{\begin{bmatrix}
{\boldsymbol{p}_w^{c_l}}^T & ... & {\boldsymbol{p}_w^{c_{l-M+1}}}^T &
{\boldsymbol{q}_{w}^{c_l}}^T & ... & {\boldsymbol{q}_{w}^{c_{l-M+1}}}^T
\end{bmatrix}}^T
\label{eq:state-vec-s-tl}\\
\boldsymbol{x}_{S_{l+1}} &=
{\begin{bmatrix}
{\boldsymbol{p}_w^{c_{l+1}}}^T & ... & {\boldsymbol{p}_w^{c_{l-M+2}}}^T &
{\boldsymbol{q}_{w}^{c_{l+1}}}^T & ... & {\boldsymbol{q}_{w}^{c_{l-M+2}}}^T
\end{bmatrix}}^T
\label{eq:state-vec-s-tl+1}
\end{align}

The estimates for the newest camera position $\boldsymbol{q}_{w}^{c_{l+1}}$ and
orientation $\boldsymbol{q}_{w}^{c_{l+1}}$ in the sliding window is computed
using the pose of the camera with respect to the IMU\footnote{The pose of the
camera with respect to the IMU is assumed to be known through calibration.}
$\left\{{^i\boldsymbol{p}}_i^c, \boldsymbol{q}_i^c\right\}$ and the current pose
estimate of the IMU at time $t_{l+1}$
$\left\{\hat{\boldsymbol{p}}_w^i(t_{l+1}),
\hat{\boldsymbol{q}}_w^i(t_{l+1})\right\}$.
\begin{align}
\hat{\boldsymbol{p}}_w^{c_{l+1}} & =
	\hat{\boldsymbol{p}}_w^i(t_{l+1}) +
	\boldsymbol{C}(\hat{\boldsymbol{q}}_w^i(t_{l+1}))^{T}{^i\boldsymbol{p}}_i^c
\\
\hat{\boldsymbol{q}}_w^{c_{l+1}} & =
	\hat{\boldsymbol{q}}_w^i(t_{l+1}) \otimes \boldsymbol{q}_i^c
\end{align}

The error covariance matrix also needs to be modified to reflect the sliding
window changes. To achieve that, we can form the new error states in the sliding
window~\citep{Delaune2013},
\begin{align}
\delta\boldsymbol{p}_w^{c_{l+1}} & =
	\delta\boldsymbol{p}_w^i(t_{l+1})
	- \boldsymbol{C}(\hat{\boldsymbol{q}}_w^i(t_{l+1}))^{T}
		\lfloor\boldsymbol{p}_i^c\times\rfloor \delta\boldsymbol{\theta}_w^i(t_{l+1})
\\
\delta\boldsymbol{\theta}_w^{c_{l+1}} & =
	\boldsymbol{C}(\boldsymbol{q}_i^c) \delta\boldsymbol{\theta}_w^i(t_{l+1})
\end{align}
which can be written as
\begin{align}
\delta\boldsymbol{p}_w^{c_{l+1}} & =
	\begin{bmatrix}
		\boldsymbol{I}_3 &
		\boldsymbol{0}_3 &
		-{\boldsymbol{C}(\hat{\boldsymbol{q}}_w^i)}^{T}\lfloor\boldsymbol{p}_i^c\times\rfloor &
		\boldsymbol{0}_{3 \times \left(6 \left(M+1\right) + 3N \right)}
	\end{bmatrix}
	\delta\boldsymbol{x}(t_{l+1})
\\
\delta\boldsymbol{\theta}_w^{c_{l+1}} & =
	\begin{bmatrix}
		\boldsymbol{0}_{3\times6} &
		\boldsymbol{C}(\boldsymbol{q}_i^c) &
		\boldsymbol{0}_{3 \times \left(6 \left(M+1\right) + 3N \right)}
	\end{bmatrix}
	\delta\boldsymbol{x}(t_{l+1}).
\end{align}
Thus we can write the structure change of the error state at time $t_{l+1}$ as
\begin{equation}
\delta\boldsymbol{x}(t_{l+1}) \leftarrow \boldsymbol{J} \delta\boldsymbol{x}(t_{l+1})
\label{eq:state-management}
\end{equation}
with
\begin{equation}
\boldsymbol{J}=\begin{bmatrix}

\boldsymbol{I}_3 &
\boldsymbol{0} &
\boldsymbol{0}  &
\boldsymbol{0}  &
\boldsymbol{0} &
\boldsymbol{0} &
\boldsymbol{0} &
\boldsymbol{0} &
\boldsymbol{0} \\

\boldsymbol{0} &
\boldsymbol{I}_3 &
\boldsymbol{0}  &
\boldsymbol{0}  &
\boldsymbol{0} &
\boldsymbol{0} &
\boldsymbol{0} &
\boldsymbol{0} &
\boldsymbol{0} \\

\boldsymbol{0} &
\boldsymbol{0} &
\boldsymbol{I}_3 &
\boldsymbol{0} &
\boldsymbol{0} &
\boldsymbol{0} &
\boldsymbol{0} &
\boldsymbol{0} &
\boldsymbol{0} \\

\boldsymbol{0}  &
\boldsymbol{0} &
\boldsymbol{0}  &
\boldsymbol{I}_6 &
\boldsymbol{0} &
\boldsymbol{0} &
\boldsymbol{0} &
\boldsymbol{0} &
\boldsymbol{0} \\

\boldsymbol{I}_3 &
\boldsymbol{0} &
-{\boldsymbol{C}(\hat{\boldsymbol{q}}_w^i)}^{T}\lfloor\boldsymbol{p}_i^c\times\rfloor &
\boldsymbol{0} &
\boldsymbol{0} &
\boldsymbol{0}_3 &
\boldsymbol{0} &
\boldsymbol{0} &
\boldsymbol{0} \\

\boldsymbol{0} &
\boldsymbol{0} &
\boldsymbol{0} &
\boldsymbol{0} &
\boldsymbol{I}_{3(M-1)} &
\boldsymbol{0} &
\boldsymbol{0} &
\boldsymbol{0} &
\boldsymbol{0} \\

\boldsymbol{0} &
\boldsymbol{0} &
\boldsymbol{C}(\boldsymbol{q}_i^c) &
\boldsymbol{0} &
\boldsymbol{0} &
\boldsymbol{0} &
\boldsymbol{0} &
\boldsymbol{0}_3 &
\boldsymbol{0} \\

\boldsymbol{0} &
\boldsymbol{0} &
\boldsymbol{0} &
\boldsymbol{0} &
\boldsymbol{0} &
\boldsymbol{0} &
\boldsymbol{I}_{3(M-1)} &
\boldsymbol{0} &
\boldsymbol{0} \\

\boldsymbol{0} &
\boldsymbol{0} &
\boldsymbol{0} &
\boldsymbol{0} &
\boldsymbol{0} &
\boldsymbol{0} &
\boldsymbol{0} &
\boldsymbol{0} &
\boldsymbol{I}_{3N}

\end{bmatrix}.
\label{eq:error-state-management-jacobian}
\end{equation}
Since
$\boldsymbol{P}(t_{l+1}) = E[\delta\boldsymbol{x}(t_{l+1})
\delta\boldsymbol{x}(t_{l+1})^T]$,
we modify the error covariance matrix according to
\begin{equation}
\boldsymbol{P}(t_{l+1}) \leftarrow \boldsymbol{J} \boldsymbol{P}(t_{l+1}) \boldsymbol{J}^{T}
\label{eq:covariance-management}
\end{equation}

\subsection{Feature States}
\label{sub:feat-state-init}

A feature state can be initialized in two ways:
\begin{itemize}
	\item from a MSCKF measurement, which means it has already been observed $m$
	times and a depth prior is available;
	\item from its first observation, which means its depth is unknown.
\end{itemize}
\citet{Li2012} show that it is most efficient to initialize a feature after a
MSCKF update. However, it may not be always possible to do so because of the
constraints listed at the beginning of Section~\ref{sub:msckf-update}, for
instance if the sensor platform is not moving in translation.

\subsubsection{Initialization from MSCKF feature}

\citet{Li2012b} proposed a method to initialize the state and error covariance of
a SLAM feature $\boldsymbol{p}_j$ from its MSCKF measurements. We detail the
equations below for completeness but the reader should refer to their paper for
the full demonstration.

Before a MSCKF update happens, the state vector can be augmented with the
feature coordinates $\boldsymbol{f}_j$ triangulated from the pose priors in MSCKF,
and the covariance matrix with infinite uncertainty on these coordinates.
\begin{equation}
\hat{\boldsymbol{x}}_{aug_{k+1|k}} \leftarrow
\begin{bmatrix}
\hat{\boldsymbol{x}}_{k+1|k} \\
\hat{\boldsymbol{f}}_j
\end{bmatrix}
\label{eq:slam-init-msckf-state-augmentation}
\end{equation}
\begin{equation}
\boldsymbol{P}_{aug_{k+1|k}} \leftarrow
\begin{bmatrix}
\boldsymbol{P}_{k+1|k} & \boldsymbol{0} \\
\boldsymbol{0} & \mu \boldsymbol{I}
\end{bmatrix}
\label{eq:slam-init-msckf-cov-augmentation}
\end{equation}
with $\mu \rightarrow \infty$.

Equation~(\ref{eq:residual-per-feature}) can be rewritten as
\begin{equation}
\delta\boldsymbol{z}_{j} \simeq
\begin{bmatrix}
\boldsymbol{H}_j &
\boldsymbol{H}_{\boldsymbol{f}_j}
\end{bmatrix}
\delta \boldsymbol{x}_{aug_{k+1|k}} +
{\boldsymbol{n}_j}\;.
\label{eq:residual-slam-init-msckf-0}
\end{equation}
and left-multiplied by the full-rank matrix
$\boldsymbol{W} = \begin{bmatrix}
\boldsymbol{A} & \boldsymbol{B}
\end{bmatrix}$
where $\boldsymbol{A}$ is the left nullspace of
$\boldsymbol{H}_{\boldsymbol{f}_j}$ and $\boldsymbol{B}$ its  column space, such
that
\begin{align}
\delta\boldsymbol{z}_{c_j} & = \boldsymbol{W}^T \delta\boldsymbol{z}_{j} 
\label{eq:residual-slam-init-msckf-1}\\
\delta\boldsymbol{z}_{c_j} & =
\begin{bmatrix}
\boldsymbol{A}^T \delta\boldsymbol{z}_{j} \\
\boldsymbol{B}^T \delta\boldsymbol{z}_{j}
\end{bmatrix} 
\label{eq:residual-slam-init-msckf-2}\\
& \simeq
\begin{bmatrix}
\boldsymbol{A}^T \boldsymbol{H}_j & \boldsymbol{A}^T \boldsymbol{H}_{\boldsymbol{f}_j} \\
\boldsymbol{B}^T \boldsymbol{H}_j & \boldsymbol{B}^T \boldsymbol{H}_{\boldsymbol{f}_j}
\end{bmatrix}
\delta \boldsymbol{x}_{aug_{k+1|k}} +
\boldsymbol{W}^T \boldsymbol{n}_j
\label{eq:residual-slam-init-msckf-3}\\
& \simeq
\begin{bmatrix}
\boldsymbol{H}_{0_j} & \boldsymbol{0}\\
\boldsymbol{H}_{1_j} & \boldsymbol{H}_{2_j}
\end{bmatrix}
\delta \boldsymbol{x}_{aug_{k+1|k}} +
\boldsymbol{W}^T \boldsymbol{n}_j
\label{eq:residual-slam-init-msckf-4}
\end{align}

\citet{Li2012b} show that the state vector and error covariance matrix resulting
from this update are
\begin{equation}
\hat{\boldsymbol{x}}_{aug_{k+1|k+1}} =
\begin{bmatrix}
\hat{\boldsymbol{x}}_{k+1|k} + \Delta\boldsymbol{x}_{k+1} \\
\hat{\boldsymbol{f}}_j -
\boldsymbol{H}_{2_j}^{-1}\boldsymbol{H}_{1_j}\Delta\boldsymbol{x}_{k+1} +
\boldsymbol{H}_{2_j}^{-1}\boldsymbol{B}^T \delta\boldsymbol{z}_{j}
\end{bmatrix}
\label{eq:slam-init-msckf-new-feature}
\end{equation}
\begin{equation}
\boldsymbol{P}_{aug_{k+1|k+1}} =
\begin{bmatrix}
\boldsymbol{P}_{k+1|k+1} & -(\boldsymbol{H}_{2_j}^{-1}\boldsymbol{H}_{1_j}\boldsymbol{P}_{k+1|k+1})^T \\
- \boldsymbol{H}_{2_j}^{-1}\boldsymbol{H}_{1_j}\boldsymbol{P}_{k+1|k+1} & \boldsymbol{P}_{22_{k+1|k+1}}
\end{bmatrix}
\label{eq:slam-init-msckf-cov-new-feature}
\end{equation}
where $\Delta\boldsymbol{x}_{k+1}$ and $\boldsymbol{P}_{k+1|k+1}$ are
respectively the state correction and error covariance matrix resulting from the
standard MSCKF update and
\begin{equation}
\boldsymbol{P}_{22_{k+1|k+1}} =
\boldsymbol{H}_{2_j}^{-1}\boldsymbol{H}_{1_j}
\boldsymbol{P}_{k+1|k+1}
(\boldsymbol{H}_{2_j}^{-1}\boldsymbol{H}_{1_j})^T
+ \sigma_V^2 \boldsymbol{H}_{2_j}^{-1}\boldsymbol{H}_{2_j}^{-T}
\label{eq:slam-init-msckf-cov-new-feature-1}
\end{equation}

The above process can be applied independently for each feature to initialize, or
as batch like in xVIO by stacking the matrices $\boldsymbol{H}_{*_j}$.

\subsubsection{Unknown-Depth Initialization}

When MSCKF cannot be applied, we use \citet{Montiel2006} to initialize a SLAM
feature $\boldsymbol{p}_j$ from its first observation. The $95\%$ acceptance
region for the depth is assumed to be between $d_{min}$ and $\infty$, which
provides the initial inverse-depth estimate  $\hat{\rho}_0 = \frac{1}{2d_{min}}$
and uncertainty $\sigma_{\rho_0} = \frac{1}{4d_{min}}$. The anchor for the
inverse-depth parametrization of the new feature is the camera frame 
$\left\{c_{i_j}\right\}$ where its first observation took place, such that
$^{i_j}\boldsymbol{z}_j =
\begin{bmatrix} \hat{\alpha}_{0_j} & \hat{\beta}_{0_j} \end{bmatrix}^T$. The
state estimate and error covariance matrix are augmented with this new SLAM
feature as
\begin{equation}
\hat{\boldsymbol{x}}_{aug_{k|k}} \leftarrow
\begin{bmatrix}
\hat{\boldsymbol{x}}_{k|k} \\
\hat{\alpha}_{0_j} \\
\hat{\beta}_{0_j} \\
\hat{\rho}_{0_j}
\end{bmatrix}
\label{eq:slam-init-state-augmentation}
\end{equation}
\begin{equation}
\boldsymbol{P}_{aug_{k|k}} \leftarrow
\begin{bmatrix}
\boldsymbol{P}_{k|k} & \boldsymbol{0} & \boldsymbol{0}\\
\boldsymbol{0} & \sigma_V^2 \boldsymbol{I}_2 & \boldsymbol{0} \\
\boldsymbol{0} & \boldsymbol{0} & \sigma_{\rho_0}^2
\end{bmatrix}.
\label{eq:slam-init-cov-augmentation}
\end{equation}
%

\subsection{Feature States Reparametrization}

The inverse-depth parametrization
$\boldsymbol{f}_{j_1} = \begin{bmatrix}
\alpha_{j_1} &
\beta_{j1} &
\rho_{j_1}
\end{bmatrix}^{T}$
of a SLAM feature $\boldsymbol{p}_j$ depends on its anchor pose
$\left\{c_{i_1}\right\}$. If a different anchor $\left\{c_{i_2}\right\}$ is
considered, the relationship
\begin{equation}
^w\boldsymbol{p}_j = \boldsymbol{p}_w^{c_{i_1}} +
\frac{1}{\rho_{j_1}} \boldsymbol{C}(\boldsymbol{q}_w^{c_{i_1}})^T
\begin{bmatrix} \alpha_{j_1} \\ \beta_{j_1} \\ 1\end{bmatrix}
= \boldsymbol{p}_w^{c_{i_2}} +
\frac{1}{\rho_{j_2}} \boldsymbol{C}(\boldsymbol{q}_w^{c_{i_2}})^T
\begin{bmatrix} \alpha_{j_2} \\ \beta_{j_2} \\ 1\end{bmatrix}
\label{eq:inverse-depth-param-equiv}
\end{equation}
provides the conversion equations
\begin{equation}
\frac{1}{\rho_{j_2}}
\begin{bmatrix} \alpha_{j_2} \\ \beta_{j_2} \\ 1\end{bmatrix} =
\boldsymbol{C}(\boldsymbol{q}_w^{c_{i_2}})
\Bigg(
- \boldsymbol{p}_w^{c_{i_2}} +
\boldsymbol{p}_w^{c_{i_1}} +
\frac{1}{\rho_{j_1}} \boldsymbol{C}(\boldsymbol{q}_w^{c_{i_1}})^T
\begin{bmatrix} \alpha_{j_1} \\ \beta_{j_1} \\ 1\end{bmatrix}
\Bigg)\;.
\label{eq:inverse-depth-param-conv}
\end{equation}

In xVIO, the anchor is one of the sliding window poses. When this anchor is about
to leave the window, that feature state estimate
$\hat{\boldsymbol{f}}_j = \begin{bmatrix}
\hat{\alpha}_j &
\hat{\beta}_j &
\hat{\rho}_j
\end{bmatrix}^{T}$
is reparametrized with the most recent pose in the sliding window as an
anchor. This is done using Equation~(\ref{eq:inverse-depth-param-conv}).
Likewise, the rows and and columns associated to that feature in the error
covariance matrix are modified using
\begin{equation}
\boldsymbol{P}_{k|k} \leftarrow \boldsymbol{J}_j \boldsymbol{P}_{k|k} \boldsymbol{J}_j^{T}
\label{eq:covariance-reparam}
\end{equation}
where
\begin{equation}
\boldsymbol{J}_j = \left[\;\begin{matrix}

\boldsymbol{I}_{15+6M+3(j-1)} &
\boldsymbol{0}
\vspace{2pt}
\\
\hline
 &
\hspace{-70pt}\raisebox{-2pt}{\mbox{$\boldsymbol{J}_{1_j}\boldsymbol{H}_{1_j}$}}
\vspace{2pt}
\\
\hline
\raisebox{-2pt}{\mbox{$\boldsymbol{0}$}} &
\raisebox{-2pt}{\mbox{$\boldsymbol{I}_{3(N-j)}$}}
\end{matrix}\;\right]\;,
\label{eq:covariance-augmentation-jacobian}
\end{equation}
with
\begin{equation}
\boldsymbol{J}_{1_j} = \hat{\rho}_{j_2}
\begin{bmatrix}
1 & 0 & -\hat{\alpha}_{j_2} \\
0 & 1 & -\hat{\beta}_{j_2} \\
0 & 0 & -\hat{\rho}_{j_2}
\end{bmatrix}
\label{eq:reparam-jacobian-prefix}
\end{equation}
and
\begin{equation}
\boldsymbol{H}_{1_j} =
\begin{aligned}
&
\left[\begin{matrix}
  \boldsymbol{0}_{3\times15} & {^j\boldsymbol{H}}_{\boldsymbol{p}_1} &
  \boldsymbol{0}_{3\times(M-6)} & {^j\boldsymbol{H}}_{\boldsymbol{p}_M} & 
  {^j\boldsymbol{H}}_{\boldsymbol{\theta}_1}
\end{matrix}\right.\\
&\qquad\qquad\qquad
\left.\begin{matrix}
  \boldsymbol{0}_{3\times(M-6)} & {^j\boldsymbol{H}}_{\boldsymbol{\theta}_M} &
  \boldsymbol{0}_{3\times(j-1)} & {\boldsymbol{H}}_{\boldsymbol{f}_j} &
  \boldsymbol{0}_{3\times(N-j)}
\end{matrix}\right]\;.
\end{aligned}
\label{eq:reparam-jacobian}
\end{equation}

Section~\ref{sec:feature-reparam} demonstrates the following expressions for the
Jacobian matrices:
\begin{align}
{^j\boldsymbol{H}}_{\boldsymbol{p}_1} & =
\boldsymbol{C}(\hat{\boldsymbol{q}}_w^{c_M})
\label{eq:reparam-jacobian-p-1}\\
{^j\boldsymbol{H}}_{\boldsymbol{p}_M} & =
- \boldsymbol{C}(\hat{\boldsymbol{q}}_w^{c_M})
\label{eq:reparam-jacobian-p-M}\\
{^j\boldsymbol{H}}_{\boldsymbol{\theta}_1} & =  - \frac{1}{\hat{\rho}_{j_1}}
\boldsymbol{C}(\hat{\boldsymbol{q}}_w^{c_M})
\boldsymbol{C}(\hat{\boldsymbol{q}}_w^{c_1})^T
\left\lfloor
\begin{bmatrix} \hat{\alpha}_{j_1} \\ \hat{\beta}_{j_1} \\ 1
\end{bmatrix}
\times\right\rfloor
\label{eq:reparam-jacobian-theta-1}\\
{^j\boldsymbol{H}}_{\boldsymbol{\theta}_M} & =
\left\lfloor
\boldsymbol{C}(\hat{\boldsymbol{q}}_w^{c_M})
\left( - \hat{\boldsymbol{p}}_w^{c_M}
+ \hat{\boldsymbol{p}}_w^{c_1} + \frac{1}{\hat{\rho}_{j_1}}
\boldsymbol{C}(\hat{\boldsymbol{q}}_w^{c_1})^T
\begin{bmatrix} \hat{\alpha}_{j_1} \\ \hat{\beta}_{j_1} \\ 1\end{bmatrix}
\right)
\times\right\rfloor
\label{eq:reparam-jacobian-theta-M}\\
{\boldsymbol{H}}_{\boldsymbol{f}_j} & =
\frac{1}{\hat{\rho}_{j_1}}
\boldsymbol{C}(\hat{\boldsymbol{q}}_w^{c_M})
\boldsymbol{C}(\hat{\boldsymbol{q}}_w^{c_1})^T
\begin{bmatrix}
1 & 0 & -\frac{\hat{\alpha}_{j_1}}{\hat{\rho}_{j_1}} \\
0 & 1 & -\frac{\hat{\beta}_{j_1}}{\hat{\rho}_{j_1}} \\
0 & 0 & -\frac{1}{\hat{\rho}_{j_1}}
\end{bmatrix}
\label{eq:reparam-jacobian-feature}
\end{align}

\chapter{Observability Analysis}
\label{ch:observability}

The observability analysis of the linearized range-VIO system is necessary since our system is based on an EKF. We note that the observability analysis of the nonlinear system would also
be required for completeness but is out of the scope of this paper.

To simplify the equations, our analysis assumes a state vector $\boldsymbol{x}^0=\begin{bmatrix}{\boldsymbol{x}_{I}}^T &
{\boldsymbol{x}_P}^{T}\end{bmatrix}^{T}$, where $\boldsymbol{x}_{I}$ was defined in Equation~(\ref{eq:state-vec-i})
and
$\boldsymbol{x}_P =
{\begin{bmatrix}
{^w\boldsymbol{p}_1}^T & ... & {^w\boldsymbol{p}_N}^T
\end{bmatrix}}^T$
includes the cartesian coordinates of the $N$ SLAM features, $N \geq 3$. \citet{li2013ijrr} proved that observability analysis for the linearized system
based on $\boldsymbol{x}^0$ is equivalent to observability analysis for the linearized system defined with $\boldsymbol{x}$ in the previous chapter.

\section{Observability Matrix}

For $k\geq1$, $\boldsymbol{M}_k = \boldsymbol{H}_k \boldsymbol{\Phi}_{k,1}$ is the k-th block row of observability matrix
$\boldsymbol{M}$. $\boldsymbol{H}_k$ is the Jacobian of the range measurement in Equation~(\ref{eq:rv-meas-4}) at time $k$
with respect to $\boldsymbol{x}^0$, which is derived in Equations~(\ref{eq:rv-linearization-1}-\ref{eq:rv-hpj3}). $\boldsymbol{\Phi}_{k,1}$
is the state transition matrix from time $1$ to time $k$~\citep{hesch2012minn}.

Without loss of generality, we can assume the ranged facet is constructed from the first 3 features in $\boldsymbol{x}_P$. Then we derive the following expression for $\boldsymbol{M}_k$ in Appendix~\ref{sec:obsmat-der}.
\begin{equation}
    \begin{aligned}
        &
        \boldsymbol{M}_k = \frac{1}{b}\big[
        \begin{matrix}
            \begin{array}{c|c|c|c|c|c|c}
                \boldsymbol{M}_{k,p} &
                \boldsymbol{M}_{k,v} &
                \boldsymbol{M}_{k,q} &
                \boldsymbol{M}_{k,b_g} &
                \boldsymbol{M}_{k,b_a}
            \end{array}
        \end{matrix}\\
        &\quad\quad\quad\quad\quad
        \begin{matrix}
            \begin{array}{c|c|c|c}
                \boldsymbol{M}_{k,p_1} &
                \boldsymbol{M}_{k,p_2} &
                \boldsymbol{M}_{k,p_3} &
                \boldsymbol{0}_{1\times3(N-3)}
            \end{array}
        \end{matrix}\big]
    \end{aligned}
\label{eq:obs-matrix}
\end{equation}
where
\begin{align}
    \boldsymbol{M}_{k,p} &= - {^w}\boldsymbol{n}^T
    \label{eq:obs-matrix-p}
    \\
    \boldsymbol{M}_{k,v} &= - (k-1) \delta t {^w}\boldsymbol{n}^T
    \label{eq:obs-matrix-v}
    \\
    &\mkern-50mu\begin{aligned}
        \boldsymbol{M}_{k,\theta} &= {^w}\boldsymbol{n} \Bigg( - \frac{a}{b} \boldsymbol{C}\left(\boldsymbol{q}_w^{c_k}\right)^T \left\lfloor {^c\boldsymbol{u}_r} \times \right\rfloor \boldsymbol{C}\left(\boldsymbol{q}_w^{i_k}\right)
        \\
        & - \Bigl\lfloor \boldsymbol{p}_w^{i_1} - \boldsymbol{v}_w^{i_1} (k-1) \delta t - \frac{1}{2}{^w}\boldsymbol{g}(k-1)^2 \delta t^2
        \\
        & \quad - \boldsymbol{p}_w^{i_k}\times \Bigr\rfloor \Bigg) \boldsymbol{C}\left(\boldsymbol{q}_{i_1}^w\right)
    \end{aligned}
    \label{eq:obs-matrix-theta}
    \\
    \boldsymbol{M}_{k,b_g} &= - \frac{a}{b}  {^w}\boldsymbol{n}^T \boldsymbol{C}\left(\boldsymbol{q}_w^{c_k}\right)^T \lfloor {^c\boldsymbol{u}_r} \times\rfloor \boldsymbol{\phi}_{12}  - {^w}\boldsymbol{n} \boldsymbol{\phi}_{52}
    \label{eq:obs-matrix-bg}
\end{align}
\begin{align}
    \boldsymbol{M}_{k,b_a} &= - {^w}\boldsymbol{n}^T \boldsymbol{\phi}_{54}
    \label{eq:obs-matrix-ba}
    \\
    \boldsymbol{M}_{k,p_1} &= \left( \left\lfloor(\boldsymbol{p}_w^{F_{3}} - \boldsymbol{p}_w^{F_{2}})\times\right\rfloor
\left(\boldsymbol{p}_w^{F_{2}} - \boldsymbol{p}_w^{I_k} \right) \right)^T
    \label{eq:obs-matrix-f1}
    \\
    \boldsymbol{M}_{k,p_2} &= \left( {^w}\boldsymbol{n} + 
    \left\lfloor(\boldsymbol{p}_w^{F_{1}} - \boldsymbol{p}_w^{F_{3}})\times\right\rfloor
    \left(\boldsymbol{p}_w^{F_{2}} -\boldsymbol{p}_w^{I_k} \right) \right)^T 
    \label{eq:obs-matrix-f2}
    \\
    \boldsymbol{M}_{k,p_3} &= \left( \left\lfloor(\boldsymbol{p}_w^{F_{2}} - \boldsymbol{p}_w^{F_{1}})\times\right\rfloor \left(\boldsymbol{p}_w^{F_{2}} - \boldsymbol{p}_w^{I_k} \right) \right)^T
    \label{eq:obs-matrix-f3}
\end{align}
and $a$ and $b$ are defined in Equations~(\ref{eq:rv-aterm}) and (\ref{eq:rv-bterm}), respectively.
$\boldsymbol{\phi}_*$ are integral terms defined in \citet{hesch2012minn}.

\section{Unobservable Directions}
\label{sec:unobs-dir}

One can verify that the vectors spanning a global position or a rotation about the gravity vector still belong
to the right nullspace of $\boldsymbol{M}_k$. Thus, the ranged facet update does not improve the
observabily over VIO under generic motion~\cite{li2013ijrr}, which is intuitive. Likewise, in the absence of
rotation, the global orientation is still not observable~\citep{wu2016minn}.

In the constant acceleration case though, unlike VIO~\citep{wu2016minn}, one can demonstrate that the vector
\begin{equation}
\boldsymbol{N}_s =
\begin{bmatrix}
{\boldsymbol{p}_w^{i_1}}^T &
{\boldsymbol{v}_w^{i_1}}^T &
{\boldsymbol{0}_{6\times1}}^T &
- {^{i}\boldsymbol{a}_w^{i}}^T &
{\boldsymbol{p}_w^{F_1}}^T &
... &
{\boldsymbol{p}_w^{F_N}}^T
\end{bmatrix}^{T}
\label{eq:scale-null-vector}
\end{equation}
which spans the scale dimension, does not belong the right nullspace of $\boldsymbol{M}_k$. The demonstration is provided in Appendix~\ref{sec:obs-aconst}. ${^{i}\boldsymbol{a}_w^{i}}$
is the constant acceleration of the IMU frame in world frame, resolved in the IMU frame. Unlike VIO,
range-VIO thus enables scale convergence even in the absence of acceleration excitations. Note that when the velocity is null,
i.e. in hover, the following unobservable direction appears instead, as discussed in Appendix~\ref{sec:obs-aconst}.
\begin{equation}
\boldsymbol{N}_h =
\begin{bmatrix}
{\boldsymbol{0}_{24\times1}}^T &
{\boldsymbol{p}_w^{F_4}}^T &
... &
{\boldsymbol{p}_w^{F_N}}^T
\end{bmatrix}^{T}
\label{eq:partial-scale-hover-null-vector}
\end{equation}
It corresponds to the depth of the SLAM features not included in the facet. This result means that in
the absence of translation, i.e. when feature depths are unknown and uncorrelated from each other, the
ranged facet provides no constrain on the features outside the facet. As soon as the platform starts moving,
visual measurements begin to correlated all feature depths, and  the depths of all features become observable
from a single ranged facet.


\begin{appendices}

\chapter{Jacobians Derivations}
\label{ch:jacobians-derivations}

In this chapter, some expressions from Chapter~\ref{ch:filter-implementation} are
demonstrated. We focus on the expressions which are not already demonstrated in
the literature. For the others, a reference was provided next to the expression.

In this section, since all derivatives are evaluated using the available state
estimates we simplify the notation
$\frac{\partial{()}}{\partial{()}} \equiv
\frac{\partial{()}}{\partial{()}}
\Bigr|_{\boldsymbol{x} = \hat{\boldsymbol{x}}}$.

\section{SLAM Measurement}
\label{sec:slam-meas-jac-der}

Equations~(\ref{eq:slam-jacobian-p-i0})--(\ref{eq:slam-jacobian-feature})
provide the expressions of the measurement Jacobians
$\boldsymbol{H}_{\boldsymbol{p}_{i_j}}$, $\boldsymbol{H}_{\boldsymbol{p}_i}$,
$\boldsymbol{H}_{\boldsymbol{\theta}_{i_j}}$,
$\boldsymbol{H}_{\boldsymbol{\theta}_i}$, $\boldsymbol{H}_{\boldsymbol{f}_j}$
for SLAM feature $\boldsymbol{p}_j$ observed in camera $\left\{c_i\right\}$
with respect to the anchor camera position $\boldsymbol{p}_w^{c_{i_j}}$, the
current camera position $\boldsymbol{p}_w^{c_i}$, the anchor camera
orientation $\boldsymbol{q}_w^{c_{i_j}}$, the current camera orientation
$\boldsymbol{q}_w^{c_i}$ and the feature coordinates $\boldsymbol{f}_j$,
respectively. Before proceeding to their derivation, we first recall the
normalized vision measurement model from Equation~(\ref{eq:vision-measurement})
\begin{equation}
^i\boldsymbol{z}_j =
\frac{1}{^{c_i}z_j}
\begin{bmatrix}^{c_i}x_j\\^{c_i}y_j\end{bmatrix}\;.
\label{eq:slam-meas-1}\\
\end{equation}

At first order, using small angle approximation about
$\hat{\boldsymbol{q}}_w^{c_i}$, one can write
\begin{align}
\boldsymbol{C}(\boldsymbol{q}_w^{c_i}) &=
\boldsymbol{C}(\hat{\boldsymbol{q}}_w^{c_i} \otimes
\delta\boldsymbol{q}_w^{c_i})
\label{eq:curr-quat-lin-1}\\
&=
\boldsymbol{C}(\delta\boldsymbol{q}_w^{c_i})
\boldsymbol{C}(\hat{\boldsymbol{q}}_w^{c_i})
\label{eq:curr-quat-lin-2}\\
&\simeq
(\boldsymbol{I}_3 - \lfloor\delta\boldsymbol{\theta}_w^{c_i}\times\rfloor)
\boldsymbol{C}(\hat{\boldsymbol{q}}_w^{c_i})\;.
\label{eq:curr-quat-lin-3}
\end{align}
Similarly about $\hat{\boldsymbol{q}}_w^{c_{i_j}}$,
\begin{equation}
\boldsymbol{C}(\boldsymbol{q}_w^{c_{i_j}}) \simeq
(\boldsymbol{I}_3 - \lfloor\delta\boldsymbol{\theta}_w^{c_{i_j}}\times\rfloor)
\boldsymbol{C}(\hat{\boldsymbol{q}}_w^{c_{i_j}})\;.
\label{eq:quat-lin-2}
\end{equation}
Hence at first order, we can rewrite Equation~(\ref{eq:feature-frame-tf2}) about
$\hat{\boldsymbol{x}}$ as
\begin{align}
^{c_i}\boldsymbol{p}_j &\simeq
\begin{aligned}
&
\left.(\boldsymbol{I}_3 - \lfloor\delta\boldsymbol{\theta}_w^{c_i}\times\rfloor)
\boldsymbol{C}(\hat{\boldsymbol{q}}_w^{c_i})
\Bigg(\boldsymbol{p}_w^{c_{i_j}}\right.\\
&
\left.+ \frac{1}{\rho_j}
\Big((\boldsymbol{I}_3 - \lfloor\delta\boldsymbol{\theta}_w^{c_{i_j}}\times\rfloor)
\boldsymbol{C}(\hat{\boldsymbol{q}}_w^{c_{i_j}})\Big)^T
\begin{bmatrix} \alpha_j \\ \beta_j \\ 1\end{bmatrix}
- \boldsymbol{p}_w^{c_i}\Bigg)\right.
\end{aligned}
\label{eq:feature-frame-tf2-lin-1}\\
&\simeq
\begin{aligned}
&
\left.(\boldsymbol{I}_3 - \lfloor\delta\boldsymbol{\theta}_w^{c_i}\times\rfloor)
\boldsymbol{C}(\hat{\boldsymbol{q}}_w^{c_i})
\Bigg(\boldsymbol{p}_w^{c_{i_j}}\right.\\
&
\left.+ \frac{1}{\rho_j}
\boldsymbol{C}(\hat{\boldsymbol{q}}_w^{c_{i_j}})^T
(\boldsymbol{I}_3 + \lfloor\delta\boldsymbol{\theta}_w^{c_{i_j}}\times\rfloor)
\begin{bmatrix} \alpha_j \\ \beta_j \\ 1\end{bmatrix}
- \boldsymbol{p}_w^{c_i}\Bigg)\right.
\end{aligned}\;,
\label{eq:feature-frame-tf2-lin-2}
\end{align}
which will be useful in the rest of this section.

\subsection{Jacobian Prefix}

The derivation of the Jacobian matrices in this section is made easier if we use
the derivative chain rule to express each Jacobian with respect to
$^{c_i}\boldsymbol{p}_j = \begin{bmatrix}
^{c_i}x_j & ^{c_i}y_j & ^{c_i}z_j \end{bmatrix}^T$ first. For instance,
\begin{equation}
\frac{\partial{^i\boldsymbol{z}_j}}{\partial{\boldsymbol{p}_w^{c_{i_j}}}} =
\frac{\partial{^i\boldsymbol{z}_j}}{\partial{^{c_i}\boldsymbol{p}_j}}.
\frac{\partial{^{c_i}\boldsymbol{p}_j}}{\partial{\boldsymbol{p}_w^{c_{i_j}}}}
\label{eq:chain-rule}
\end{equation}
where
\begin{align}
\frac{\partial{^i\boldsymbol{z}_j}}{\partial{^{c_i}\boldsymbol{p}_j}} &=
\begin{bmatrix}
\frac{\partial{\big(\frac{^{c_i}x_j}{^{c_i}z_j}}\big)}{\partial{^{c_i}x_j}} &
\frac{\partial{\big(\frac{^{c_i}x_j}{^{c_i}z_j}}\big)}{\partial{^{c_i}y_j}} &
\frac{\partial{\big(\frac{^{c_i}x_j}{^{c_i}z_j}}\big)}{\partial{^{c_i}z_j}} \\
\frac{\partial{\big(\frac{^{c_i}y_j}{^{c_i}z_j}}\big)}{\partial{^{c_i}x_j}} &
\frac{\partial{\big(\frac{^{c_i}y_j}{^{c_i}z_j}}\big)}{\partial{^{c_i}y_j}} &
\frac{\partial{\big(\frac{^{c_i}y_j}{^{c_i}z_j}}\big)}{\partial{^{c_i}z_j}}
\end{bmatrix}
\label{eq:jac-prefix-1}\\
 &=
\begin{bmatrix}
\frac{1}{^{c_i}\hat{z}_j} &
0 &
-\frac{^{c_i}\hat{x}_j}{^{c_i}\hat{z}_j^2} \\
0 &
\frac{1}{^{c_i}\hat{z}_j} &
-\frac{^{c_i}\hat{y}_j}{^{c_i}\hat{z}_j^2}
\end{bmatrix}
\label{eq:jac-prefix-2}\\
 &= \frac{1}{^{c_i}\hat{z}_j}
\begin{bmatrix}
\boldsymbol{I}_2 & -^i\hat{\boldsymbol{z}}_j
\end{bmatrix}
\label{eq:jac-prefix-3}\\
&= {^i\boldsymbol{J}_j}\;.
\label{eq:jac-prefix-4}
\end{align}

\subsection{Anchor Camera Position}

By injecting Equation~(\ref{eq:jac-prefix-4}) in Equation~(\ref{eq:chain-rule}),
and then using Equation~(\ref{eq:feature-frame-tf2-lin-2}), we get
\begin{align}
\frac{\partial{^i\boldsymbol{z}_j}}{\partial{\boldsymbol{p}_w^{c_{i_j}}}} &=
{^i\boldsymbol{J}_j}
\frac{\partial{^{c_i}\boldsymbol{p}_j}}{\partial{\boldsymbol{p}_w^{c_{i_j}}}}
\label{eq:jac-anchor-pos-1}\\
&=
{^i\boldsymbol{J}_j}
\boldsymbol{C}(\hat{\boldsymbol{q}}_w^{c_i})
\label{eq:jac-anchor-pos-2}\\
&= \boldsymbol{H}_{\boldsymbol{p}_{i_j}}\;.
\label{eq:jac-anchor-pos-3}
\end{align}

\subsection{Current Camera Position}

\begin{align}
\frac{\partial{^i\boldsymbol{z}_j}}{\partial{\boldsymbol{p}_w^{c_i}}} &=
{^i\boldsymbol{J}_j}
\frac{\partial{^{c_i}\boldsymbol{p}_j}}{\partial{\boldsymbol{p}_w^{c_i}}}
\label{eq:jac-curr-pos-1}\\
&=
-{^i\boldsymbol{J}_j}
\boldsymbol{C}(\hat{\boldsymbol{q}}_w^{c_i})
\label{eq:jac-curr-pos-2}\\
&= \boldsymbol{H}_{\boldsymbol{p}_i}\;.
\label{eq:jac-curr-pos-3}
\end{align}

\subsection{Anchor Camera Orientation}

\begin{align}
\frac{\partial{^i\boldsymbol{z}_j}}
{\partial{\delta\boldsymbol{\theta}_w^{c_{i_j}}}} &=
{^i\boldsymbol{J}_j}
\frac{\partial{^{c_i}\boldsymbol{p}_j}}
{\partial{\delta\boldsymbol{\theta}_w^{c_{i_j}}}}
\label{eq:jac-anchor-att-1}\\
&\simeq
\begin{aligned}
&
{^i\boldsymbol{J}_j}
\frac{\partial{}}{\partial{\delta\boldsymbol{\theta}_w^{c_{i_j}}}}
\Bigg(\frac{1}{\rho_j}(\boldsymbol{I}_3
- \lfloor\delta\boldsymbol{\theta}_w^{c_i}\times\rfloor)\\
&
\boldsymbol{C}(\hat{\boldsymbol{q}}_w^{c_i})
\boldsymbol{C}(\hat{\boldsymbol{q}}_w^{c_{i_j}})^T
\lfloor\delta\boldsymbol{\theta}_w^{c_{i_j}}\times\rfloor
\begin{bmatrix} \alpha_j \\ \beta_j \\ 1\end{bmatrix}\Bigg)
\end{aligned}
\label{eq:jac-anchor-att-2}\\
&\simeq
\begin{aligned}
&
{^i\boldsymbol{J}_j}
\frac{\partial{}}{\partial{\delta\boldsymbol{\theta}_w^{c_{i_j}}}}
\Bigg(-\frac{1}{\rho_j}(\boldsymbol{I}_3 -
\lfloor\delta\boldsymbol{\theta}_w^{c_i}\times\rfloor)\\
&
\boldsymbol{C}(\hat{\boldsymbol{q}}_w^{c_i})
\boldsymbol{C}(\hat{\boldsymbol{q}}_w^{c_{i_j}})^T
\left\lfloor\begin{bmatrix} \alpha_j \\ \beta_j \\ 1
\end{bmatrix}\times\right\rfloor \delta\boldsymbol{\theta}_w^{c_{i_j}}\Bigg)
\end{aligned}
\label{eq:jac-anchor-att-3}\\
&\simeq -\frac{1}{\hat{\rho}_j}
{^i\boldsymbol{J}_j}
\boldsymbol{C}(\hat{\boldsymbol{q}}_w^{c_i})
\boldsymbol{C}(\hat{\boldsymbol{q}}_w^{c_{i_j}})^T
\left\lfloor\begin{bmatrix} \hat{\alpha}_j \\ \hat{\beta}_j \\ 1
\end{bmatrix}\times\right\rfloor
\label{eq:jac-anchor-att-4}
\\
&\simeq \boldsymbol{H}_{\boldsymbol{\theta}_{i_j}}
\label{eq:jac-anchor-att-5}
\end{align}

\subsection{Current Camera Orientation}

\begin{align}
\frac{\partial{^i\boldsymbol{z}_j}}
{\partial{\delta\boldsymbol{\theta}_w^{c_i}}} &=
{^i\boldsymbol{J}_j}
\frac{\partial{^{c_i}\boldsymbol{p}_j}}
{\partial{\delta\boldsymbol{\theta}_w^{c_i}}}
\label{eq:jac-curr-att-1}\\
&\simeq
\begin{aligned}
&
{^i\boldsymbol{J}_j}
\frac{\partial{}}{\partial{\delta\boldsymbol{\theta}_w^{c_i}}}
\Bigg(- \lfloor\delta\boldsymbol{\theta}_w^{c_i}\times\rfloor
\boldsymbol{C}(\hat{\boldsymbol{q}}_w^{c_i})
\bigg(\boldsymbol{p}_w^{c_{i_j}} \\
&
+ \frac{1}{\rho_j}
\boldsymbol{C}(\hat{\boldsymbol{q}}_w^{c_{i_j}})^T
(\boldsymbol{I}_3 + \lfloor\delta\boldsymbol{\theta}_w^{c_{i_j}}\times\rfloor)
\begin{bmatrix} \alpha_j \\ \beta_j \\ 1\end{bmatrix}
- \boldsymbol{p}_w^{c_i}\bigg)\Bigg)
\end{aligned}
\label{eq:jac-curr-att-2}\\
&\simeq
\begin{aligned}
&
{^i\boldsymbol{J}_j}
\frac{\partial{}}{\partial{\delta\boldsymbol{\theta}_w^{c_i}}}
\Bigg(\Biggl\lfloor
\boldsymbol{C}(\hat{\boldsymbol{q}}_w^{c_i})
\bigg(\boldsymbol{p}_w^{c_{i_j}} \\
&
+ \frac{1}{\rho_j}
\boldsymbol{C}(\hat{\boldsymbol{q}}_w^{c_{i_j}})^T
(\boldsymbol{I}_3 + \lfloor\delta\boldsymbol{\theta}_w^{c_{i_j}}\times\rfloor)
\begin{bmatrix} \alpha_j \\ \beta_j \\ 1\end{bmatrix}
- \boldsymbol{p}_w^{c_i}\bigg)\times\Biggl\rfloor
\delta\boldsymbol{\theta}_w^{c_i}\Bigg)
\end{aligned}
\label{eq:jac-curr-att-3}\\
 &\simeq
{^i\boldsymbol{J}_j}
\left\lfloor
\boldsymbol{C}(\hat{\boldsymbol{q}}_w^{c_i})
\Bigg(\hat{\boldsymbol{p}}_w^{c_{i_j}} -
\hat{\boldsymbol{p}}_w^{c_i} +
\frac{1}{\hat{\rho}_j}
\boldsymbol{C}(\hat{\boldsymbol{q}}_w^{c_{i_j}})^T
\begin{bmatrix} \hat{\alpha}_j \\ \hat{\beta}_j \\ 1
\end{bmatrix}\Bigg)\times\right\rfloor
\label{eq:jac-curr-att-4}\\
 &\simeq \boldsymbol{H}_{\boldsymbol{\theta}_i}
\label{eq:jac-curr-att-5}
\end{align}

\subsection{Feature Coordinates}

To demonstrate the expression for $\boldsymbol{H}_{\boldsymbol{f}_j}$, we first
compute the Jacobians for each of its components.

\begin{align}
\frac{\partial{^i\boldsymbol{z}_j}}
{\partial{\alpha_j}} &=
{^i\boldsymbol{J}_j}
\frac{\partial{^{c_i}\boldsymbol{p}_j}}{\partial{\alpha_j}}
\label{eq:jac-feat-1}\\
&=
\begin{aligned}
&
{^i\boldsymbol{J}_j}
\frac{\partial{}}{\partial{\alpha_j}}
\Bigg(
\left.
\boldsymbol{C}(\boldsymbol{q}_w^{c_i})
\frac{1}{\rho_j}
\boldsymbol{C}(\boldsymbol{q}_w^{c_{i_j}})^T
\begin{bmatrix} \alpha_j \\ \beta_j \\ 1\end{bmatrix}
\right)
\end{aligned}
\label{eq:jac-feat-2}\\
 &= \frac{1}{\hat{\rho}_j}
{^i\boldsymbol{J}_j}
\boldsymbol{C}(\hat{\boldsymbol{q}}_w^{c_i})
\boldsymbol{C}(\hat{\boldsymbol{q}}_w^{c_{i_j}})^T
\begin{bmatrix} 1 \\ 0 \\ 0\end{bmatrix}
\label{eq:jac-feat-3}\\
\frac{\partial{^i\boldsymbol{z}_j}}
{\partial{\beta_j}} &=
\frac{1}{\hat{\rho}_j}
{^i\boldsymbol{J}_j}
\boldsymbol{C}(\hat{\boldsymbol{q}}_w^{c_i})
\boldsymbol{C}(\hat{\boldsymbol{q}}_w^{c_{i_j}})^T
\begin{bmatrix} 0 \\ 1 \\ 0\end{bmatrix}
\label{eq:jac-feat-4}\\
\frac{\partial{^i\boldsymbol{z}_j}}
{\partial{\rho_j}} &=
-\frac{1}{\hat{\rho}_j^2}
{^i\boldsymbol{J}_j}
\boldsymbol{C}(\hat{\boldsymbol{q}}_w^{c_i})
\boldsymbol{C}(\hat{\boldsymbol{q}}_w^{c_{i_j}})^T
\begin{bmatrix} \hat{\alpha}_j \\ \hat{\beta}_j \\ 1\end{bmatrix}
\label{eq:jac-feat-5}
\end{align}

Stacking the expressions above column-wise leads to

\begin{align}
\frac{\partial{^i\boldsymbol{z}_j}}
{\partial{\boldsymbol{f}_j}} &=
\frac{1}{\hat{\rho}_j}
{^i\boldsymbol{J}_j}
\boldsymbol{C}(\hat{\boldsymbol{q}}_w^{c_i})
\boldsymbol{C}(\hat{\boldsymbol{q}}_w^{c_{i_j}})^T
\begin{bmatrix}
1 & 0 & -\frac{\hat{\alpha}_j}{\hat{\rho}_j} \\
0 & 1 & -\frac{\hat{\beta}_j}{\hat{\rho}_j} \\
0 & 0 & -\frac{1}{\hat{\rho}_j}
\end{bmatrix}
\label{eq:jac-feat-6}\\
&= \boldsymbol{H}_{\boldsymbol{f}_j}
\label{eq:jac-feat-7}
\end{align}

\section{Range-Visual Update}
\label{sec:range-visual-update-jac}

In this section, we will show that the measurement Jacobian of the range-visual
update can be decomposed into simpler Jacobians, derive preliminary expressions
for these with respect to a simpler alternative state vector $\boldsymbol{x}'$,
derive the expression of the full Jacobian with respect to $\boldsymbol{x}'$, and
finally the Jacobian with respect to the actual filter state $\boldsymbol{x}$.

\subsection{Decomposition of the Jacobian}

We can rewrite the range-visual update model of
Equation~(\ref{eq:rv-meas-3}) as
\begin{equation}
^i{z}_r =
\frac{a}{b}
\label{eq:rv-ab}
\end{equation}
with
\begin{align}
a &= (\boldsymbol{p}_w^{f_{j_2}} - \boldsymbol{p}_w^{c_i})^T{^w\boldsymbol{n}}
	= \boldsymbol{a}^T_0{^w\boldsymbol{n}}
\label{eq:rv-aterm}\\
b &= {{^w\boldsymbol{u}_{r_i}}^T}{^w\boldsymbol{n}}
	= \boldsymbol{b}^T_0{^w\boldsymbol{n}}\;.
\label{eq:rv-bterm}
\end{align}

The measurement Jacobian can be expressed as
\begin{align}
\boldsymbol{H} &= \frac{\partial{^i{z}_r}}{\partial{\boldsymbol{x}}}
\label{eq:rv-jac}\\
&= \frac{1}{\hat{b}^2}\left(\frac{\partial{a}}{\partial{\boldsymbol{x}}}\hat{b}
   				 - \hat{a}\frac{\partial{b}}{\partial{\boldsymbol{x}}}\right)
\label{eq:rv-jac-ab}\\
&= \frac{1}{\hat{b}^2}\left(\frac{\partial{\left(\boldsymbol{a}^T_0{^w\boldsymbol{n}}\right)}}{\partial{\boldsymbol{x}}}\hat{b}
   				 - \hat{a}\frac{\partial{\left(\boldsymbol{b}^T_0{^w\boldsymbol{n}}\right)}}{\partial{\boldsymbol{x}}}\right)
\end{align}

The partial derivative of the cross product of two vectors of identical size
$\boldsymbol{u}$ and $\boldsymbol{v}$, with respect to a third vector $\boldsymbol{w}$
can be demonstrated to be
\begin{equation}
\frac{\partial{\left(\boldsymbol{u}^T\boldsymbol{v}\right)}}{\partial{\boldsymbol{w}}}
= \boldsymbol{v}^T\frac{\partial{\boldsymbol{u}}}{\partial{\boldsymbol{w}}}
+ \boldsymbol{u}^T\frac{\partial{\boldsymbol{v}}}{\partial{\boldsymbol{w}}}\;.
\label{eq:dot-product-derivative}
\end{equation}

Then we can write $\boldsymbol{H}$ as
\begin{align}
\boldsymbol{H} &=
\frac{1}{\hat{b}^2}
\left(\left({^w\hat{\boldsymbol{n}}}^T \frac{\partial{\boldsymbol{a}_0}}{\partial{\boldsymbol{x}}}
			 + \hat{\boldsymbol{a}}^T_0 \frac{\partial{^w\boldsymbol{n}}}{\partial{\boldsymbol{x}}} \right) \hat{b}
- \hat{a} \left({^w\hat{\boldsymbol{n}}}^T \frac{\partial{\boldsymbol{b}_0}}{\partial{\boldsymbol{x}}}
			 + \hat{\boldsymbol{b}}^T_0 \frac{\partial{^w\boldsymbol{n}}}{\partial{\boldsymbol{x}}} \right)\right)
\label{eq:rv-jac-a0b0-1}\\
&=
\frac{1}{\hat{b}} \left({^w\hat{\boldsymbol{n}}}^T \frac{\partial{\boldsymbol{a}_0}}{\partial{\boldsymbol{x}}}
			+ \hat{\boldsymbol{a}}^T_0 \frac{\partial{^w\boldsymbol{n}}}{\partial{\boldsymbol{x}}} \right)
- \frac{\hat{a}}{\hat{b}^2} \left({^w\hat{\boldsymbol{n}}}^T \frac{\partial{\boldsymbol{b}_0}}{\partial{\boldsymbol{x}}}
			 + \hat{\boldsymbol{b}}^T_0 \frac{\partial{^w\boldsymbol{n}}}{\partial{\boldsymbol{x}}} \right)\;.
\label{eq:rv-jac-a0b0-2}
\end{align}

\subsection{Preliminary Derivations}

Recalling Equations~(\ref{eq:state-vec-iv})-(\ref{eq:state-vec-f}), we break
down the state vector $\boldsymbol{x}$ into its inertial states
$\boldsymbol{x}_{I} \in \mathbb{R}^{16}$, sliding window states
$\boldsymbol{x}_S \in \mathbb{R}^{7M}$, and feature states
$\boldsymbol{x}_F \in \mathbb{R}^{3N}$.
\begin{equation}
\boldsymbol{x} = \begin{bmatrix}
{\boldsymbol{x}_I}^T &
{\boldsymbol{x}_S}^{T} &
{\boldsymbol{x}_F}^{T}
\end{bmatrix}^{T}
\label{eq:state-vec-isf}
\end{equation}

From Equation~(\ref{eq:inverse-depth-param2}), we also know the feature states are
expressed in inverse-depth coordinates
$\boldsymbol{f}_j = \begin{bmatrix}
\alpha_j &
\beta_j &
\rho_j
\end{bmatrix}^{T}$
using anchor poses selected among the sliding window states.

At this point, we introduce an alternative state vector $\boldsymbol{x}'$ that
retains the same inertial and sliding window states as $\boldsymbol{x}$, but
the feature states are expressed by their cartesian coordinates in world frame
$^w\boldsymbol{p}_j = \begin{bmatrix}
^wx_j & ^wy_j & ^wz_j \end{bmatrix}^T$, i.e.
\begin{equation}
\boldsymbol{x}' = \begin{bmatrix}
{\boldsymbol{x}_I}^T &
{\boldsymbol{x}_S}^{T} &
{\boldsymbol{x}'_F}^{T}
\end{bmatrix}^{T}
\label{eq:state-vec-isf'}
\end{equation}
with
\begin{equation}
\boldsymbol{x}'_F =
{\begin{bmatrix}
{^w\boldsymbol{p}_1}^T & ... & {^w\boldsymbol{p}_N}^T
\end{bmatrix}}^T\;.
\label{eq:state-vec-f}
\end{equation}
It is worth noting that $\boldsymbol{x}' \in \mathbb{R}^{16+7M+3N}$, i.e. maintains
the same size as $\boldsymbol{x}$.

Equation~(\ref{eq:rv-jac-a0b0-2}) gives measurement Jacobian from the partial
derivatives $^w\boldsymbol{n}$, $\boldsymbol{a}_0$ and $\boldsymbol{b}_0$ with
respect to the state vector $\boldsymbol{x}$. To make the derivation easier, we
will consider the partials with respect to $\boldsymbol{x}'$ first, before going
back to $\boldsymbol{x}$ in the last subsection.

Before the first derivation, we remind the reader that for two vectors
$\boldsymbol{u}, \boldsymbol{v} \in \mathbb{R}^3$, and a vector $\boldsymbol{w}$
of any dimension, we can prove that
\begin{equation}
\frac{\partial{\left(\boldsymbol{u} \times \boldsymbol{v}\right)}}{\partial{\boldsymbol{w}}}
= \frac{\partial{\boldsymbol{u}}}{\partial{\boldsymbol{w}}} \times \boldsymbol{v}
+ \boldsymbol{u} \times \frac{\partial{\boldsymbol{v}}}{\partial{\boldsymbol{w}}} 
\end{equation}

Now let us derive the partials of interest.
\begin{align}
\frac{\partial{^w\boldsymbol{n}}}{\partial{\boldsymbol{x}'}} &=
\frac{\partial{}}{\partial{\boldsymbol{x}'}}
\left(
(\boldsymbol{p}_w^{f_{j_1}} - \boldsymbol{p}_w^{f_{j_2}})
	\times(\boldsymbol{p}_w^{f_{j_3}} - \boldsymbol{p}_w^{f_{j_2}})
\right)
\label{eq:rv-jac-N-x'-1}\\
&=
\begin{matrix}
\frac{\partial{(\boldsymbol{p}_w^{f_{j_1}} - \boldsymbol{p}_w^{f_{j_2}})}}{\partial{\boldsymbol{x}'}}
\times (\hat{\boldsymbol{p}}_w^{f_{j_3}} - \hat{\boldsymbol{p}}_w^{f_{j_2}})
+ (\hat{\boldsymbol{p}}_w^{f_{j_1}} - \hat{\boldsymbol{p}}_w^{f_{j_2}})
\times \frac{\partial{(\boldsymbol{p}_w^{f_{j_3}} - \boldsymbol{p}_w^{f_{j_2}})}}{\partial{\boldsymbol{x}'}}
\end{matrix}
\label{eq:rv-jac-N-x'-2}\\
&=
\begin{matrix}
(\hat{\boldsymbol{p}}_w^{f_{j_1}} - \hat{\boldsymbol{p}}_w^{f_{j_2}})
\times \frac{\partial{(\boldsymbol{p}_w^{f_{j_3}} - \boldsymbol{p}_w^{f_{j_2}})}}{\partial{\boldsymbol{x}'}}
- (\hat{\boldsymbol{p}}_w^{f_{j_3}} - \hat{\boldsymbol{p}}_w^{f_{j_2}}) \times \frac{\partial{(\boldsymbol{p}_w^{f_{j_1}} - \boldsymbol{p}_w^{f_{j_2}})}}{\partial{\boldsymbol{x}'}}
\end{matrix}
\label{eq:rv-jac-N-x'-3}\\
&=
\begin{aligned}
& \begin{matrix}
(\hat{\boldsymbol{p}}_w^{f_{j_2}} - \hat{\boldsymbol{p}}_w^{f_{j_3}}) \times \frac{\partial{\boldsymbol{p}_w^{f_{j_1}}}}{\partial{\boldsymbol{x}'}}
+ (\hat{\boldsymbol{p}}_w^{f_{j_3}} - \hat{\boldsymbol{p}}_w^{f_{j_1}}) \times \frac{\partial{\boldsymbol{p}_w^{f_{j_2}}}}{\partial{\boldsymbol{x}'}} \end{matrix}\\
&\qquad\qquad\qquad\qquad\qquad\qquad\qquad\qquad
\begin{matrix}
+ (\hat{\boldsymbol{p}}_w^{f_{j_1}} - \hat{\boldsymbol{p}}_w^{f_{j_2}})
\times \frac{\partial{\boldsymbol{p}_w^{f_{j_3}}}}{\partial{\boldsymbol{x}'}}
\end{matrix}
\end{aligned}
\label{eq:rv-jac-N-x'-4}
\\
&=
\begin{aligned}
&
\Bigg[\begin{matrix}\begin{array}{c|c|c}
\boldsymbol{0}_{3\times(15+6M+3(j_1-1))} &
\left\lfloor(\hat{\boldsymbol{p}}_w^{f_{j_2}} - \hat{\boldsymbol{p}}_w^{f_{j_3}})\times\right\rfloor &
\boldsymbol{0}_{3\times3(j_2-j_1-1)}
\end{array}\end{matrix}\\
&
\begin{matrix}\begin{array}{c|c|c|c}
\left\lfloor(\hat{\boldsymbol{p}}_w^{f_{j_3}} - \hat{\boldsymbol{p}}_w^{f_{j_1}})\times\right\rfloor &
\boldsymbol{0}_{3\times3(j_3-j_2-1)} &
\left\lfloor(\hat{\boldsymbol{p}}_w^{f_{j_1}} - \hat{\boldsymbol{p}}_w^{f_{j_2}})\times\right\rfloor &
\boldsymbol{0}_{3\times3(N-j_3)}
\end{array}\end{matrix}\Bigg]
\end{aligned}
\label{eq:rv-jac-N-x'-5}
\end{align}

\begin{align}
\frac{\partial{\boldsymbol{a}_0}}{\partial{\boldsymbol{x}'}} &=
\frac{\partial{}}{\partial{\boldsymbol{x}'}}
\left( \boldsymbol{p}_w^{f_{j_2}} - \boldsymbol{p}_w^{c_i} \right)
\label{eq:rv-jac-a0-x'-1}\\
&=
\begin{aligned}
&
\begin{bmatrix}\begin{array}{c|c|c|c|c}
\boldsymbol{0}_{3\times(15+3(i-1))} &
- \boldsymbol{I}_3 &
\boldsymbol{0}_{3\times(15+6M+3(j_2-i-1))} &
\boldsymbol{I}_3 &
\boldsymbol{0}_{3\times3(N-j_2)}
\end{array}
\end{bmatrix}
\end{aligned}
\label{eq:rv-jac-a0-x'-2}
\end{align}

\begin{align}
\frac{\partial{\boldsymbol{b}_0}}{\partial{\boldsymbol{x}'}} &=
\frac{\partial{^w\boldsymbol{u}_{r_i}}}{\partial{\boldsymbol{x}'}}
\label{eq:rv-jac-b0-x'-1}\\
&=
\frac{\partial{}}{\partial{\boldsymbol{x}'}}
\left(
\boldsymbol{C}(\boldsymbol{q}_w^{c_i})^T {^c\boldsymbol{u}_r}
\right)
\label{eq:rv-jac-b0-x'-2}\\
&\simeq
\frac{\partial{}}{\partial{\boldsymbol{x}'}}
\left(
\left( (\boldsymbol{I}_3 - \lfloor\delta\boldsymbol{\theta}_w^{c_i}\times\rfloor) 
\boldsymbol{C}\left(\hat{\boldsymbol{q}}_w^{c_i} \right) \right)^T {^c\boldsymbol{u}_r}
\right)
\label{eq:rv-jac-b0-x'-3}\\
&\simeq
\frac{\partial{}}{\partial{\boldsymbol{x}'}}
\left(
\boldsymbol{C}\left(\hat{\boldsymbol{q}}_w^{c_i} \right)^T
(\boldsymbol{I}_3 - \lfloor\delta\boldsymbol{\theta}_w^{c_i}\times\rfloor)^T
{^c\boldsymbol{u}_r}
\right)
\label{eq:rv-jac-b0-x'-4}\\
&\simeq
\frac{\partial{}}{\partial{\boldsymbol{x}'}}
\left(
\boldsymbol{C}\left(\hat{\boldsymbol{q}}_w^{c_i} \right)^T
(\boldsymbol{I}_3 + \lfloor\delta\boldsymbol{\theta}_w^{c_i}\times\rfloor)
{^c\boldsymbol{u}_r}
\right)
\label{eq:rv-jac-b0-x'-5}\\
&\simeq
\frac{\partial{}}{\partial{\boldsymbol{x}'}}
\left(
\boldsymbol{C}\left(\hat{\boldsymbol{q}}_w^{c_i} \right)^T
(\boldsymbol{I}_3 - \lfloor {^c\boldsymbol{u}_r} \times\rfloor)
\delta\boldsymbol{\theta}_w^{c_i}
\right)
\label{eq:rv-jac-b0-x'-6}\\
&\simeq
- \boldsymbol{C}\left(\hat{\boldsymbol{q}}_w^{c_i} \right)^T
\lfloor {^c\boldsymbol{u}_r} \times\rfloor
\begin{bmatrix}\begin{array}{c|c|c}
\boldsymbol{0}_{3\times(15+3M+3(i-1))} &
\boldsymbol{I}_3 &
\boldsymbol{0}_{3\times(3N+3(M-i))}
\end{array}\end{bmatrix}
\label{eq:rv-jac-b0-x'-7}
\end{align}

\subsection{Measurement Jacobian with respect to $\boldsymbol{x}'$}

We can use Equation~(\ref{eq:rv-jac-a0b0-2}) to formulate the measurement
Jacobian with respect to $\boldsymbol{x}'$ as
\begin{align}
\boldsymbol{H}' &= \frac{\partial{^i{z}_r}}{\partial{\boldsymbol{x}'}}
\label{eq:rv-jac'-alphabeta}\\
&= \frac{1}{\hat{b}} \hat{\boldsymbol{\alpha}} - \frac{\hat{a}}{\hat{b}^2}\hat{\boldsymbol{\beta}}
\label{eq:rv-jac'-alphabeta-1}
\end{align}
with%
\begin{align}
\boldsymbol{\alpha} &=
{^w\boldsymbol{n}}^T \frac{\partial{\boldsymbol{a}_0}}{\partial{\boldsymbol{x}'}}
			+ \boldsymbol{a}^T_0 \frac{\partial{^w\boldsymbol{n}}}{\partial{\boldsymbol{x}'}}
\label{eq:rv-jac'-alpha}\\
\boldsymbol{\beta} &=
{^w\boldsymbol{n}}^T \frac{\partial{\boldsymbol{b}_0}}{\partial{\boldsymbol{x}'}}
			 + \boldsymbol{b}^T_0 \frac{\partial{^w\boldsymbol{n}}}{\partial{\boldsymbol{x}'}}
\label{eq:rv-jac'-beta}
\end{align}

Using the expressions derived in the previous section, we now can move forward
and derive $\boldsymbol{\alpha}$, $\boldsymbol{\beta}$, and then $\boldsymbol{H}'$.

\begin{equation}
\hat{\boldsymbol{\alpha}} =
\begin{aligned}
&
\Bigg[\begin{matrix}\begin{array}{c|c|c|}
\boldsymbol{0}_{3\times(15+3(i-1))} &
- ^w\hat{\boldsymbol{n}}^T &
\boldsymbol{0}_{3\times(6M+3(j_1-i-1))}
\end{array}\end{matrix}\\
&
\begin{matrix}\begin{array}{c|c|}
\left(\hat{\boldsymbol{p}}_w^{f_{j_2}} - \hat{\boldsymbol{p}}_w^{c_i}\right)^T \left\lfloor(\hat{\boldsymbol{p}}_w^{f_{j_2}} - \hat{\boldsymbol{p}}_w^{f_{j_3}})\times\right\rfloor &
\boldsymbol{0}_{3\times3(j_2-j_1-1)}
\end{array}\end{matrix}\\
&
\begin{matrix}\begin{array}{c|c|}
^w\hat{\boldsymbol{n}}^T + 
\left(\hat{\boldsymbol{p}}_w^{f_{j_2}} - \hat{\boldsymbol{p}}_w^{c_i}\right)^T \left\lfloor(\hat{\boldsymbol{p}}_w^{f_{j_3}} - \hat{\boldsymbol{p}}_w^{f_{j_1}})\times\right\rfloor &
\boldsymbol{0}_{3\times3(j_3-j_2-1)}
\end{array}\end{matrix}\\
&
\begin{matrix}\begin{array}{c|c}
\left(\hat{\boldsymbol{p}}_w^{f_{j_2}} - \hat{\boldsymbol{p}}_w^{c_i}\right)^T \left\lfloor(\hat{\boldsymbol{p}}_w^{f_{j_1}} - \hat{\boldsymbol{p}}_w^{f_{j_2}})\times\right\rfloor &
\boldsymbol{0}_{3\times3(N-j_3)}
\end{array}\end{matrix}\Bigg]
\end{aligned}
\label{eq:rv-jac'-alpha-1}
\end{equation}

\begin{equation}
\hat{\boldsymbol{\beta}} =
\begin{aligned}
&
\Bigg[\begin{matrix}\begin{array}{c|c|c|}
\boldsymbol{0}_{3\times(15+3M+3(i-1))} &
- ^w\hat{\boldsymbol{n}}^T \boldsymbol{C}\left(\hat{\boldsymbol{q}}_w^{c_i} \right)^T \lfloor {^c\boldsymbol{u}_r} \times\rfloor &
\boldsymbol{0}_{3\times(3M+3(j_1-i-1))}
\end{array}\end{matrix}\\
&
\begin{matrix}\begin{array}{c|c|}
{^c\boldsymbol{u}_r}^T \boldsymbol{C}\left(\hat{\boldsymbol{q}}_w^{c_i} \right) \left\lfloor(\hat{\boldsymbol{p}}_w^{f_{j_2}} - \hat{\boldsymbol{p}}_w^{f_{j_3}})\times\right\rfloor &
\boldsymbol{0}_{3\times3(j_2-j_1-1)}
\end{array}\end{matrix}\\
&
\begin{matrix}\begin{array}{c|c|}
{^c\boldsymbol{u}_r}^T \boldsymbol{C}\left(\hat{\boldsymbol{q}}_w^{c_i} \right) \left\lfloor(\hat{\boldsymbol{p}}_w^{f_{j_3}} - \hat{\boldsymbol{p}}_w^{f_{j_1}})\times\right\rfloor &
\boldsymbol{0}_{3\times3(j_3-j_2-1)}
\end{array}\end{matrix}\\
&
\begin{matrix}\begin{array}{c|c}
{^c\boldsymbol{u}_r}^T \boldsymbol{C}\left(\hat{\boldsymbol{q}}_w^{c_i} \right) \left\lfloor(\hat{\boldsymbol{p}}_w^{f_{j_1}} - \hat{\boldsymbol{p}}_w^{f_{j_2}})\times\right\rfloor &
\boldsymbol{0}_{3\times3(N-j_3)}
\end{array}\end{matrix}\Bigg]
\end{aligned}
\label{eq:rv-jac'-beta-1}
\end{equation}

\begin{equation}
\boldsymbol{H}' =
\begin{aligned}
&
\Bigg[\begin{matrix}\begin{array}{c|c|c|}
\boldsymbol{0}_{3\times(15+3(i-1))} &
- \frac{1}{\hat{b}} ^w\hat{\boldsymbol{n}}^T &
\boldsymbol{0}_{3\times(3(M-1))}
\end{array}\end{matrix}\\
&
\begin{matrix}\begin{array}{c|c|}
\frac{\hat{a}}{\hat{b}^2} ^w\hat{\boldsymbol{n}}^T \boldsymbol{C}\left(\hat{\boldsymbol{q}}_w^{c_i} \right)^T \lfloor {^c\boldsymbol{u}_r} \times\rfloor &
\boldsymbol{0}_{3\times(3M+3(j_1-i-1))}
\end{array}\end{matrix}\\
&
\begin{matrix}\begin{array}{c|c|}
\frac{1}{\hat{b}} \left(\hat{\boldsymbol{p}}_w^{f_{j_2}} - \hat{\boldsymbol{p}}_w^{c_i} - \frac{\hat{a}}{\hat{b}} \left({^c\boldsymbol{u}_r}^T \boldsymbol{C}\left(\hat{\boldsymbol{q}}_w^{c_i} \right)\right)^T \right)^T \left\lfloor(\hat{\boldsymbol{p}}_w^{f_{j_2}} - \hat{\boldsymbol{p}}_w^{f_{j_3}})\times\right\rfloor
\end{array}\end{matrix}\\
&
\begin{matrix}\begin{array}{c|}
\boldsymbol{0}_{3\times3(j_2-j_1-1)}
\end{array}\end{matrix}\\
&
\begin{matrix}\begin{array}{c|c|}
\frac{1}{\hat{b}} ^w\hat{\boldsymbol{n}}^T + 
\frac{1}{\hat{b}} \left(\hat{\boldsymbol{p}}_w^{f_{j_2}} - \hat{\boldsymbol{p}}_w^{c_i} - \frac{\hat{a}}{\hat{b}} \left({^c\boldsymbol{u}_r}^T \boldsymbol{C}\left(\hat{\boldsymbol{q}}_w^{c_i} \right)\right)^T \right)^T \left\lfloor(\hat{\boldsymbol{p}}_w^{f_{j_3}} - \hat{\boldsymbol{p}}_w^{f_{j_1}})\times\right\rfloor
\end{array}\end{matrix}\\
&
\begin{matrix}\begin{array}{c|}
\boldsymbol{0}_{3\times3(j_3-j_2-1)}
\end{array}\end{matrix}\\
&
\begin{matrix}\begin{array}{c|c}
\frac{1}{\hat{b}} \left(\hat{\boldsymbol{p}}_w^{f_{j_2}} - \hat{\boldsymbol{p}}_w^{c_i} - \frac{\hat{a}}{\hat{b}} \left({^c\boldsymbol{u}_r}^T \boldsymbol{C}\left(\hat{\boldsymbol{q}}_w^{c_i} \right)\right)^T \right)^T \left\lfloor(\hat{\boldsymbol{p}}_w^{f_{j_1}} - \hat{\boldsymbol{p}}_w^{f_{j_2}})\times\right\rfloor
\end{array}\end{matrix}\\
&
\begin{matrix}\begin{array}{c}
\boldsymbol{0}_{3\times3(N-j_3)}
\end{array}\end{matrix}\Bigg]
\end{aligned}
\label{eq:rv-jac'-1}
\end{equation}

We can rewrite
\begin{align}
\left(\frac{\hat{a}}{\hat{b}} {^c\boldsymbol{u}_r}^T \boldsymbol{C}\left(\hat{\boldsymbol{q}}_w^{c_i} \right) \right)^T
&= \frac{\hat{a}}{\hat{b}}\boldsymbol{C}\left(\hat{\boldsymbol{q}}_w^{c_i} \right)^T {^c\boldsymbol{u}_r}
\label{eq:rv-jac'-tf1}\\
&= \hat{\boldsymbol{p}}_w^{c_i} + ^i\hat{z}_r{^w\hat{\boldsymbol{u}}_{r_i}}
- \hat{\boldsymbol{p}}_w^{c_i}
\label{eq:rv-jac'-tf2}\\
&= \hat{\boldsymbol{p}}_w^{I_i} - \hat{\boldsymbol{p}}_w^{c_i}
\label{eq:rv-jac'-tf3}
\end{align}
where $\hat{\boldsymbol{p}}_w^{I}$ are the 3D cartesian coordinates of the impact point
of the LRF on the terrain at time instant $i$, so that we can simplify Equation~(\ref{eq:rv-jac'-1}) as
\begin{align}
\boldsymbol{H}' &=
\begin{aligned}
&
\Bigg[\begin{matrix}\begin{array}{c|c|c|}
\boldsymbol{0}_{3\times(15+3(i-1))} &
- \frac{1}{\hat{b}} ^w\hat{\boldsymbol{n}}^T &
\boldsymbol{0}_{3\times(3(M-1))}
\end{array}\end{matrix}\\
&
\begin{matrix}\begin{array}{c|c|}
\frac{\hat{a}}{\hat{b}^2} ^w\hat{\boldsymbol{n}}^T \boldsymbol{C}\left(\hat{\boldsymbol{q}}_w^{c_i} \right)^T \lfloor {^c\boldsymbol{u}_r} \times\rfloor &
\boldsymbol{0}_{3\times(3M+3(j_1-i-1))}
\end{array}\end{matrix}\\
&
\begin{matrix}\begin{array}{c|c|}
\frac{1}{\hat{b}} \left(\hat{\boldsymbol{p}}_w^{f_{j_2}}
- \hat{\boldsymbol{p}}_w^{I_i} \right)^T \left\lfloor(\hat{\boldsymbol{p}}_w^{f_{j_2}} - \hat{\boldsymbol{p}}_w^{f_{j_3}})\times\right\rfloor
\end{array}\end{matrix}\\
&
\begin{matrix}\begin{array}{c|}
\boldsymbol{0}_{3\times3(j_2-j_1-1)}
\end{array}\end{matrix}\\
&
\begin{matrix}\begin{array}{c|c|}
\frac{1}{\hat{b}} ^w\hat{\boldsymbol{n}}^T + 
\frac{1}{\hat{b}} \left(\hat{\boldsymbol{p}}_w^{f_{j_2}} -\hat{\boldsymbol{p}}_w^{I_i} \right)^T \left\lfloor(\hat{\boldsymbol{p}}_w^{f_{j_3}} - \hat{\boldsymbol{p}}_w^{f_{j_1}})\times\right\rfloor
\end{array}\end{matrix}\\
&
\begin{matrix}\begin{array}{c|}
\boldsymbol{0}_{3\times3(j_3-j_2-1)}
\end{array}\end{matrix}\\
&
\begin{matrix}\begin{array}{c|c}
\frac{1}{\hat{b}} \left(\hat{\boldsymbol{p}}_w^{f_{j_2}}
- \hat{\boldsymbol{p}}_w^{I_i} \right)^T \left\lfloor(\hat{\boldsymbol{p}}_w^{f_{j_1}} - \hat{\boldsymbol{p}}_w^{f_{j_2}})\times\right\rfloor
\end{array}\end{matrix}\\
&
\begin{matrix}\begin{array}{c}
\boldsymbol{0}_{3\times3(N-j_3)}
\end{array}\end{matrix}\Bigg]
\end{aligned}
\label{eq:rv-jac'-2}\\
&=
\begin{aligned}
&
\Bigg[\begin{matrix}\begin{array}{c|c|c|}
\boldsymbol{0}_{3\times(15+3(i-1))} &
- \frac{1}{\hat{b}} ^w\hat{\boldsymbol{n}}^T &
\boldsymbol{0}_{3\times(3(M-1))}
\end{array}\end{matrix}\\
&
\begin{matrix}\begin{array}{c|c|}
- \frac{\hat{a}}{\hat{b}^2} \left( \lfloor {^c\boldsymbol{u}_r} \times\rfloor \boldsymbol{C}\left(\hat{\boldsymbol{q}}_w^{c_i} \right) {^w\hat{\boldsymbol{n}}} \right)^T &
\boldsymbol{0}_{3\times(3M+3(j_1-i-1))}
\end{array}\end{matrix}\\
&
\begin{matrix}\begin{array}{c|c|}
\frac{1}{\hat{b}}
\left( \left\lfloor(\hat{\boldsymbol{p}}_w^{f_{j_3}} - \hat{\boldsymbol{p}}_w^{f_{j_2}})\times\right\rfloor
\left(\hat{\boldsymbol{p}}_w^{f_{j_2}} - \hat{\boldsymbol{p}}_w^{I_i} \right) \right)^T
\end{array}\end{matrix}\\
&
\begin{matrix}\begin{array}{c|}
\boldsymbol{0}_{3\times3(j_2-j_1-1)}
\end{array}\end{matrix}\\
&
\begin{matrix}\begin{array}{c|c|}
\frac{1}{\hat{b}} \left( ^w\hat{\boldsymbol{n}} + 
\left\lfloor(\hat{\boldsymbol{p}}_w^{f_{j_1}} - \hat{\boldsymbol{p}}_w^{f_{j_3}})\times\right\rfloor
\left(\hat{\boldsymbol{p}}_w^{f_{j_2}} -\hat{\boldsymbol{p}}_w^{I_i} \right) \right)^T 
\end{array}\end{matrix}\\
&
\begin{matrix}\begin{array}{c|}
\boldsymbol{0}_{3\times3(j_3-j_2-1)}
\end{array}\end{matrix}\\
&
\begin{matrix}\begin{array}{c|c}
\frac{1}{\hat{b}} \left( \left\lfloor(\hat{\boldsymbol{p}}_w^{f_{j_2}} - \hat{\boldsymbol{p}}_w^{f_{j_1}})\times\right\rfloor \left(\hat{\boldsymbol{p}}_w^{f_{j_2}} - \hat{\boldsymbol{p}}_w^{I_i} \right) \right)^T
\end{array}\end{matrix}\\
&
\begin{matrix}\begin{array}{c}
\boldsymbol{0}_{3\times3(N-j_3)}
\end{array}\end{matrix}\Bigg]
\end{aligned}
\label{eq:rv-jac'-3}
\end{align}

Or equivalently,
\begin{equation}
^i\delta z_r \simeq
\begin{aligned}
&
\boldsymbol{H}_{\boldsymbol{p}_i} \delta\boldsymbol{p}_w^{c_i}
+ \boldsymbol{H}_{\boldsymbol{\theta}_i} \delta\boldsymbol{\theta}_w^{c_i}
\\
&
+ \boldsymbol{H}_{\boldsymbol{p}_{j_1}} \delta\boldsymbol{p}_w^{f_{j_1}}
+ \boldsymbol{H}_{\boldsymbol{p}_{j_2}} \delta\boldsymbol{p}_w^{f_{j_2}}
+ \boldsymbol{H}_{\boldsymbol{p}_{j_3}} \delta\boldsymbol{p}_w^{f_{j_3}}
\\
&
+ ^in_r
\end{aligned}
\label{eq:rv-linearization-1}
\end{equation}
where
\begin{align}
\boldsymbol{H}_{\boldsymbol{p}_i} &= - \frac{1}{\hat{b}} ^w\hat{\boldsymbol{n}}^T
\label{eq:rv-hpi}\\
\boldsymbol{H}_{\boldsymbol{\theta}_i} &= - \frac{\hat{a}}{\hat{b}^2} \left( \lfloor {^c\boldsymbol{u}_r} \times\rfloor \boldsymbol{C}\left(\hat{\boldsymbol{q}}_w^{c_i} \right) {^w\hat{\boldsymbol{n}}} \right)^T
\label{eq:rv-thetai}\\
\boldsymbol{H}_{\boldsymbol{p}_{j_1}} &= \frac{1}{\hat{b}}
\left( \left\lfloor(\hat{\boldsymbol{p}}_w^{f_{j_3}} - \hat{\boldsymbol{p}}_w^{f_{j_2}})\times\right\rfloor
\left(\hat{\boldsymbol{p}}_w^{f_{j_2}} - \hat{\boldsymbol{p}}_w^{I_i} \right) \right)^T
\label{eq:rv-hpj1}\\
\boldsymbol{H}_{\boldsymbol{p}_{j_2}} &= \frac{1}{\hat{b}} \left( ^w\hat{\boldsymbol{n}} + 
\left\lfloor(\hat{\boldsymbol{p}}_w^{f_{j_1}} - \hat{\boldsymbol{p}}_w^{f_{j_3}})\times\right\rfloor
\left(\hat{\boldsymbol{p}}_w^{f_{j_2}} -\hat{\boldsymbol{p}}_w^{I_i} \right) \right)^T 
\label{eq:rv-hpj2}\\
\boldsymbol{H}_{\boldsymbol{p}_{j_3}} &= \frac{1}{\hat{b}} \left( \left\lfloor(\hat{\boldsymbol{p}}_w^{f_{j_2}} - \hat{\boldsymbol{p}}_w^{f_{j_1}})\times\right\rfloor \left(\hat{\boldsymbol{p}}_w^{f_{j_2}} - \hat{\boldsymbol{p}}_w^{I_i} \right) \right)^T
\label{eq:rv-hpj3}
\end{align}

\subsection{Measurement Jacobian with respect to $\boldsymbol{x}$}

In this last subsection, we recover the measurement Jacobian with respect to the
actual state $\boldsymbol{x}$. We remind the reader that $\boldsymbol{x}'$
includes the cartesian coordinates of feature $j$,
$^w\boldsymbol{p}_j = \begin{bmatrix}
^wx_j & ^wy_j & ^wz_j \end{bmatrix}^T$, 
whereas $\boldsymbol{x}$ includes its inverse-depth coordinates
$\boldsymbol{f}_j = \begin{bmatrix}
\alpha_j &
\beta_j &
\rho_j
\end{bmatrix}^{T}$
and an anchor pose corresponding to the camera frame at index $i_j$ in the
sliding window. The other are related according to
\begin{equation}
^w\boldsymbol{p}_j = \boldsymbol{p}_w^{c_{i_j}} +
\frac{1}{\rho_j} \boldsymbol{C}(\boldsymbol{q}_w^{c_{i_j}})^T
\begin{bmatrix} \alpha_j \\ \beta_j \\ 1\end{bmatrix}
\label{eq:inverse-depth-param2-bis}
\end{equation}

This means that to form $\boldsymbol{H}$, we need an expression of
$\frac{\partial{^i{z}_r}}{\partial{\boldsymbol{p}_w^{c_{i_j}}}}$,
$\frac{\partial{^i{z}_r}}{\partial{\delta\boldsymbol{\theta}_w^{c_{i_j}}}}$,
$\frac{\partial{^i{z}_r}}{\partial{\alpha_j}}$,
$\frac{\partial{^i{z}_r}}{\partial{\beta_j}}$,
and $\frac{\partial{^i{z}_r}}{\partial{\rho_j}}$ for each feature $j$ in
the LRF triangle.

\begin{align}
\frac{\partial{^i{z}_r}}{\partial{\boldsymbol{p}_w^{c_{i_j}}}}
&= \frac{\partial{^i{z}_r}}{\partial{^w\boldsymbol{p}_j}} \frac{\partial{^w\boldsymbol{p}_j}}{\partial{\boldsymbol{p}_w^{c_{i_j}}}}
\label{eq:rv-jac-p-anchor-1}\\
&= \frac{\partial{^i{z}_r}}{\partial{^w\boldsymbol{p}_j}}
\label{eq:rv-jac-p-anchor-2}\\
&= \boldsymbol{H}_{\boldsymbol{p}_j}
\label{eq:rv-jac-p-anchor-3}
\end{align}

\begin{align}
\frac{\partial{^i{z}_r}}{\partial{\delta\boldsymbol{\theta}_w^{c_{i_j}}}}
&= \frac{\partial{^i{z}_r}}{\partial{^w\boldsymbol{p}_j}} \frac{\partial{^w\boldsymbol{p}_j}}{\partial{\delta\boldsymbol{\theta}_w^{c_{i_j}}}}
\label{eq:rv-jac-theta-anchor-1}\\
&\simeq \frac{\partial{^i{z}_r}}{\partial{^w\boldsymbol{p}_j}} \frac{\partial{}}{\partial{\delta\boldsymbol{\theta}_w^{c_{i_j}}}}
\left( \frac{1}{\rho_j}
\left( (\boldsymbol{I}_3 - \lfloor\delta\boldsymbol{\theta}_w^{c_{i_j}}\times\rfloor) \boldsymbol{C}(\hat{\boldsymbol{q}}_w^{c_{i_j}}) \right)^T
\begin{bmatrix} \alpha_j \\ \beta_j \\ 1\end{bmatrix} \right)
\label{eq:rv-jac-theta-anchor-2}\\
&\simeq \frac{\partial{^i{z}_r}}{\partial{^w\boldsymbol{p}_j}} \frac{\partial{}}{\partial{\delta\boldsymbol{\theta}_w^{c_{i_j}}}}
\left( \frac{1}{\rho_j}
\boldsymbol{C}(\hat{\boldsymbol{q}}_w^{c_{i_j}})^T
(\boldsymbol{I}_3 - \lfloor\delta\boldsymbol{\theta}_w^{c_{i_j}}\times\rfloor)^T
\begin{bmatrix} \alpha_j \\ \beta_j \\ 1\end{bmatrix} \right)
\label{eq:rv-jac-theta-anchor-3}\\
&\simeq \frac{\partial{^i{z}_r}}{\partial{^w\boldsymbol{p}_j}} \frac{\partial{}}{\partial{\delta\boldsymbol{\theta}_w^{c_{i_j}}}}
\left( \frac{1}{\rho_j}
\boldsymbol{C}(\hat{\boldsymbol{q}}_w^{c_{i_j}})^T
(\boldsymbol{I}_3 + \lfloor\delta\boldsymbol{\theta}_w^{c_{i_j}}\times\rfloor)
\begin{bmatrix} \alpha_j \\ \beta_j \\ 1\end{bmatrix} \right)
\label{eq:rv-jac-theta-anchor-4}\\
&\simeq \frac{\partial{^i{z}_r}}{\partial{^w\boldsymbol{p}_j}} \frac{\partial{}}{\partial{\delta\boldsymbol{\theta}_w^{c_{i_j}}}}
\left( \frac{1}{\rho_j}
\boldsymbol{C}(\hat{\boldsymbol{q}}_w^{c_{i_j}})^T
\lfloor\delta\boldsymbol{\theta}_w^{c_{i_j}}\times\rfloor
\begin{bmatrix} \alpha_j \\ \beta_j \\ 1\end{bmatrix} \right)
\label{eq:rv-jac-theta-anchor-5}\\
&\simeq \frac{\partial{^i{z}_r}}{\partial{^w\boldsymbol{p}_j}} \frac{\partial{}}{\partial{\delta\boldsymbol{\theta}_w^{c_{i_j}}}}
\left( - \frac{1}{\rho_j}
\boldsymbol{C}(\hat{\boldsymbol{q}}_w^{c_{i_j}})^T
\left\lfloor \begin{bmatrix} \alpha_j \\ \beta_j \\ 1\end{bmatrix} \times \right\rfloor
\delta\boldsymbol{\theta}_w^{c_{i_j}} \right)
\label{eq:rv-jac-theta-anchor-6}\\
&\simeq - \frac{1}{\hat{\rho_j}} \frac{\partial{^i{z}_r}}{\partial{^w\boldsymbol{p}_j}} 
\boldsymbol{C}(\hat{\boldsymbol{q}}_w^{c_{i_j}})^T
\left\lfloor \begin{bmatrix} \hat{\alpha_j} \\ \hat{\beta_j} \\ 1\end{bmatrix} \times \right\rfloor
\label{eq:rv-jac-theta-anchor-7}
\\
&\simeq - \frac{1}{\hat{\rho_j}} \boldsymbol{H}_{\boldsymbol{p}_j} 
\boldsymbol{C}(\hat{\boldsymbol{q}}_w^{c_{i_j}})^T
\left\lfloor \begin{bmatrix} \hat{\alpha_j} \\ \hat{\beta_j} \\ 1\end{bmatrix} \times \right\rfloor
\label{eq:rv-jac-theta-anchor-8}
\end{align}

\begin{align}
\frac{\partial{^i{z}_r}}{\partial{\alpha_j}}
&= \frac{\partial{^i{z}_r}}{\partial{^w\boldsymbol{p}_j}} \frac{\partial{^w\boldsymbol{p}_j}}{\partial{\alpha_j}}
\label{eq:rv-jac-p-alpha-1}\\
&= \frac{1}{\hat{\rho_j}} \frac{\partial{^i{z}_r}}{\partial{^w\boldsymbol{p}_j}} \boldsymbol{C}(\hat{\boldsymbol{q}}_w^{c_{i_j}})^T
\begin{bmatrix} 1 \\ 0 \\ 0\end{bmatrix}
\label{eq:rv-jac-p-alpha-2}
\\
&= \frac{1}{\hat{\rho_j}} \boldsymbol{H}_{\boldsymbol{p}_j}  \boldsymbol{C}(\hat{\boldsymbol{q}}_w^{c_{i_j}})^T
\begin{bmatrix} 1 \\ 0 \\ 0\end{bmatrix}
\label{eq:rv-jac-p-alpha-3}
\end{align}

\begin{align}
\frac{\partial{^i{z}_r}}{\partial{\beta_j}}
&= \frac{\partial{^i{z}_r}}{\partial{^w\boldsymbol{p}_j}} \frac{\partial{^w\boldsymbol{p}_j}}{\partial{\beta_j}}
\label{eq:rv-jac-p-beta-1}\\
&= \frac{1}{\hat{\rho_j}} \frac{\partial{^i{z}_r}}{\partial{^w\boldsymbol{p}_j}} \boldsymbol{C}(\hat{\boldsymbol{q}}_w^{c_{i_j}})^T
\begin{bmatrix} 0 \\ 1 \\ 0\end{bmatrix}
\label{eq:rv-jac-p-beta-2}
\\
&= \frac{1}{\hat{\rho_j}} \boldsymbol{H}_{\boldsymbol{p}_j} \boldsymbol{C}(\hat{\boldsymbol{q}}_w^{c_{i_j}})^T
\begin{bmatrix} 0 \\ 1 \\ 0\end{bmatrix}
\label{eq:rv-jac-p-beta-3}
\end{align}

\begin{align}
\frac{\partial{^i{z}_r}}{\partial{\rho_j}}
&= \frac{\partial{^i{z}_r}}{\partial{^w\boldsymbol{p}_j}} \frac{\partial{^w\boldsymbol{p}_j}}{\partial{\rho_j}}
\label{eq:rv-jac-p-rho-1}\\
&= - \frac{1}{\hat{\rho_j}^2} \frac{\partial{^i{z}_r}}{\partial{^w\boldsymbol{p}_j}} \boldsymbol{C}(\hat{\boldsymbol{q}}_w^{c_{i_j}})^T
\begin{bmatrix} \hat{\alpha_j} \\ \hat{\beta_j} \\ 1\end{bmatrix}
\label{eq:rv-jac-p-rho-2}
\\
&= - \frac{1}{\hat{\rho_j}^2} \boldsymbol{H}_{\boldsymbol{p}_j} \boldsymbol{C}(\hat{\boldsymbol{q}}_w^{c_{i_j}})^T
\begin{bmatrix} \hat{\alpha_j} \\ \hat{\beta_j} \\ 1\end{bmatrix}
\label{eq:rv-jac-p-rho-3}
\end{align}

Thus we can extend Equation~(\ref{eq:rv-linearization-1}) to
\begin{equation}
^i\delta z_r \simeq
\begin{aligned}
&
\boldsymbol{H}_{\boldsymbol{p}_{i_1}} \delta\boldsymbol{p}_w^{c_{i_1}}
+ \boldsymbol{H}_{\boldsymbol{p}_{i_2}} \delta\boldsymbol{p}_w^{c_{i_2}}
+ \boldsymbol{H}_{\boldsymbol{p}_{i_3}} \delta\boldsymbol{p}_w^{c_{i_3}}
+ \boldsymbol{H}_{\boldsymbol{p}_i} \delta\boldsymbol{p}_w^{c_i}
\\
&
+ \boldsymbol{H}_{\boldsymbol{\theta}_{i_1}} \delta\boldsymbol{\theta}_w^{c_{i_1}}
+ \boldsymbol{H}_{\boldsymbol{\theta}_{i_2}} \delta\boldsymbol{\theta}_w^{c_{i_2}}
+ \boldsymbol{H}_{\boldsymbol{\theta}_{i_3}} \delta\boldsymbol{\theta}_w^{c_{i_3}}
+ \boldsymbol{H}_{\boldsymbol{\theta}_i} \delta\boldsymbol{\theta}_w^{c_i}
\\
&
+ \boldsymbol{H}_{\boldsymbol{f}_{j_1}} \delta\boldsymbol{f}_{j_1}
+ \boldsymbol{H}_{\boldsymbol{f}_{j_2}} \delta\boldsymbol{f}_{j_2}
+ \boldsymbol{H}_{\boldsymbol{f}_{j_3}} \delta\boldsymbol{f}_{j_3}
\\
&
+ ^in_r
\end{aligned}
\label{eq:rv-linearization-2}
\end{equation}
where
\begin{align}
\boldsymbol{H}_{\boldsymbol{p}_{i_1}} &= \boldsymbol{H}_{\boldsymbol{p}_{j_1}}
\label{eq:rv-hpi1}\\
\boldsymbol{H}_{\boldsymbol{p}_{i_2}} &= \boldsymbol{H}_{\boldsymbol{p}_{j_2}}
\label{eq:rv-hpi2}\\
\boldsymbol{H}_{\boldsymbol{p}_{i_3}} &= \boldsymbol{H}_{\boldsymbol{p}_{j_3}}
\label{eq:rv-hpi3}\\
\boldsymbol{H}_{\boldsymbol{p}_i} &= - \frac{1}{\hat{b}} ^w\hat{\boldsymbol{n}}^T
\label{eq:rv-hpi-2}\\
\boldsymbol{H}_{\boldsymbol{\theta}_{i_1}} &= - \frac{1}{\hat{\rho_{j_1}}} \boldsymbol{H}_{\boldsymbol{p}_{j_1}} 
\boldsymbol{C}(\hat{\boldsymbol{q}}_w^{c_{i_{j_1}}})^T
\left\lfloor \begin{bmatrix} \hat{\alpha_{j_1}} \\ \hat{\beta_{j_1}} \\ 1\end{bmatrix} \times \right\rfloor
\label{eq:rv-hthetai1}\\
\boldsymbol{H}_{\boldsymbol{\theta}_{i_2}} &= - \frac{1}{\hat{\rho_{j_2}}} \boldsymbol{H}_{\boldsymbol{p}_{j_2}} 
\boldsymbol{C}(\hat{\boldsymbol{q}}_w^{c_{i_{j_2}}})^T
\left\lfloor \begin{bmatrix} \hat{\alpha_{j_2}} \\ \hat{\beta_{j_2}} \\ 1\end{bmatrix} \times \right\rfloor
\label{eq:rv-hthetai2}\\
\boldsymbol{H}_{\boldsymbol{\theta}_{i_3}} &= - \frac{1}{\hat{\rho_{j_3}}} \boldsymbol{H}_{\boldsymbol{p}_{j_3}} 
\boldsymbol{C}(\hat{\boldsymbol{q}}_w^{c_{i_{j_3}}})^T
\left\lfloor \begin{bmatrix} \hat{\alpha_{j_3}} \\ \hat{\beta_{j_3}} \\ 1\end{bmatrix} \times \right\rfloor
\label{eq:rv-hthetai3}\\
\boldsymbol{H}_{\boldsymbol{\theta}_i} &= - \frac{\hat{a}}{\hat{b}^2} \left( \lfloor {^c\boldsymbol{u}_r} \times\rfloor \boldsymbol{C}\left(\hat{\boldsymbol{q}}_w^{c_i} \right) {^w\hat{\boldsymbol{n}}} \right)^T
\label{eq:rv-thetai-2}\\
\boldsymbol{H}_{\boldsymbol{f}_{j_1}} &= \frac{1}{\hat{\rho_{j_1}}} \boldsymbol{H}_{\boldsymbol{p}_{j_1}} \boldsymbol{C}(\hat{\boldsymbol{q}}_w^{c_{i_{j_1}}})^T
\begin{bmatrix} 1 & 0 & - \frac{\hat{\alpha_{j_1}}}{\hat{\rho_{j_1}}} \\ 0 & 1 & - \frac{\hat{\beta_{j_1}}}{\hat{\rho_{j_1}}} \\ 0 & 0& - \frac{1}{\hat{\rho_{j_1}}}\end{bmatrix}
\label{eq:rv-hfj1}\\
\boldsymbol{H}_{\boldsymbol{f}_{j_2}} &= \frac{1}{\hat{\rho_{j_2}}} \boldsymbol{H}_{\boldsymbol{p}_{j_2}} \boldsymbol{C}(\hat{\boldsymbol{q}}_w^{c_{i_{j_2}}})^T
\begin{bmatrix} 1 & 0 & - \frac{\hat{\alpha_{j_2}}}{\hat{\rho_{j_2}}} \\ 0 & 1 & - \frac{\hat{\beta_{j_2}}}{\hat{\rho_{j_2}}} \\ 0 & 0& - \frac{1}{\hat{\rho_{j_2}}}\end{bmatrix}
\label{eq:rv-hfj2}\\
\boldsymbol{H}_{\boldsymbol{f}_{j_3}} &= \frac{1}{\hat{\rho_{j_3}}} \boldsymbol{H}_{\boldsymbol{p}_{j_3}} \boldsymbol{C}(\hat{\boldsymbol{q}}_w^{c_{i_{j_3}}})^T
\begin{bmatrix} 1 & 0 & - \frac{\hat{\alpha_{j_3}}}{\hat{\rho_{j_3}}} \\ 0 & 1 & - \frac{\hat{\beta_{j_3}}}{\hat{\rho_{j_3}}} \\ 0 & 0& - \frac{1}{\hat{\rho_{j_3}}}\end{bmatrix}
\label{eq:rv-hfj3}
\end{align}
and
\begin{align}
\boldsymbol{H}_{\boldsymbol{p}_{j_1}} &= \frac{1}{\hat{b}}
\left( \left\lfloor(\hat{\boldsymbol{p}}_w^{f_{j_3}} - \hat{\boldsymbol{p}}_w^{f_{j_2}})\times\right\rfloor
\left(\hat{\boldsymbol{p}}_w^{f_{j_2}} - \hat{\boldsymbol{p}}_w^{I_i} \right) \right)^T
\label{eq:rv-hpj1-bis}\\
\boldsymbol{H}_{\boldsymbol{p}_{j_2}} &= \frac{1}{\hat{b}} \left( ^w\hat{\boldsymbol{n}} + 
\left\lfloor(\hat{\boldsymbol{p}}_w^{f_{j_1}} - \hat{\boldsymbol{p}}_w^{f_{j_3}})\times\right\rfloor
\left(\hat{\boldsymbol{p}}_w^{f_{j_2}} -\hat{\boldsymbol{p}}_w^{I_i} \right) \right)^T 
\label{eq:rv-hpj2-bis}\\
\boldsymbol{H}_{\boldsymbol{p}_{j_3}} &= \frac{1}{\hat{b}} \left( \left\lfloor(\hat{\boldsymbol{p}}_w^{f_{j_2}} - \hat{\boldsymbol{p}}_w^{f_{j_1}})\times\right\rfloor \left(\hat{\boldsymbol{p}}_w^{f_{j_2}} - \hat{\boldsymbol{p}}_w^{I_i} \right) \right)^T
\label{eq:rv-hpj3-bis}
\end{align}

\section{Feature State Initialization with MSCKF}
\label{sec:feature-state-init-msckf}

The equations presented in Subsection~\ref{sub:feat-state-init} to initialize a
feature state from MSCKF measurements are already demonstrated in 
\citet{Li2012b}. Some parts of this demonstration are not trivial and this
intends to fill these local gaps. Since the reader is intended to follow up the
demonstration in \citet{Li2012b}, we follow their notation and the numbering of
the equations in their paper, which differs from that in
Subsection~\ref{sub:feat-state-init}.

Even though \citet{Li2012b} and this report only provide the demonstration for
the initialization, one can verify that matrices $\boldsymbol{H}_o$,
$\boldsymbol{H}_1$ and $\boldsymbol{H}_2$ can be of any size. So the
demonstration still holds for the batch initialization of multiple features,
by stacking up these matrices.

\subsection{Covariance of the Innovation}

Starting from Equation~(16) in \citet{Li2012b},
\begin{align}
\boldsymbol{S} &=
\begin{bmatrix}
\boldsymbol{H}_o & \boldsymbol{0} \\
\boldsymbol{H}_1 & \boldsymbol{H}_2
\end{bmatrix}
\begin{bmatrix}
\boldsymbol{P} & \boldsymbol{0} \\
\boldsymbol{0} & \mu\boldsymbol{I}
\end{bmatrix}
\begin{bmatrix}
\boldsymbol{H}_o & \boldsymbol{0} \\
\boldsymbol{H}_1 & \boldsymbol{H}_2
\end{bmatrix}^T
+ \sigma^2\boldsymbol{I}
\label{eq:feature-state-init-msckf-1}\\
&=
\begin{bmatrix}
\boldsymbol{H}_o & \boldsymbol{0} \\
\boldsymbol{H}_1 & \boldsymbol{H}_2
\end{bmatrix}
\begin{bmatrix}
\boldsymbol{P} & \boldsymbol{0} \\
\boldsymbol{0} & \mu\boldsymbol{I}
\end{bmatrix}
\begin{bmatrix}
\boldsymbol{H}_o^T & \boldsymbol{H}_1^T \\
\boldsymbol{0} & \boldsymbol{H}_2^T
\end{bmatrix}^T
+ \sigma^2\boldsymbol{I}
\label{eq:feature-state-init-msckf-2}\\
&=
\begin{bmatrix}
\boldsymbol{H}_o & \boldsymbol{0} \\
\boldsymbol{H}_1 & \boldsymbol{H}_2
\end{bmatrix}
\begin{bmatrix}
\boldsymbol{P}\boldsymbol{H}_o^T & \boldsymbol{P}\boldsymbol{H}_1^T \\
\boldsymbol{0} & \mu\boldsymbol{H}_2^T
\end{bmatrix}
+ \sigma^2\boldsymbol{I}
\label{eq:feature-state-init-msckf-3}\\
&=
\begin{bmatrix}
\boldsymbol{H}_o\boldsymbol{P}\boldsymbol{H}_o^T &
\boldsymbol{H}_o\boldsymbol{P}\boldsymbol{H}_1^T \\
\boldsymbol{H}_1\boldsymbol{P}\boldsymbol{H}_o^T &
\boldsymbol{H}_1\boldsymbol{P}\boldsymbol{H}_1^T
+ \mu\boldsymbol{H}_2\boldsymbol{H}_2^T
\end{bmatrix}
+ \sigma^2\boldsymbol{I}
\label{eq:feature-state-init-msckf-4}\\
&=
\begin{bmatrix}
\boldsymbol{H}_o\boldsymbol{P}\boldsymbol{H}_o^T + \sigma^2\boldsymbol{I} &
\boldsymbol{H}_o\boldsymbol{P}\boldsymbol{H}_1^T \\
\boldsymbol{H}_1\boldsymbol{P}\boldsymbol{H}_o^T &
\boldsymbol{H}_1\boldsymbol{P}\boldsymbol{H}_1^T
+ \mu\boldsymbol{H}_2\boldsymbol{H}_2^T
+ \sigma^2\boldsymbol{I}
\end{bmatrix}
\label{eq:feature-state-init-msckf-5}
\end{align}
which is Equation~(17) in \citet{Li2012b}.

If we pose
\begin{equation}
\boldsymbol{S} =
\begin{bmatrix}
\boldsymbol{A} & \boldsymbol{B} \\
\boldsymbol{C} & \boldsymbol{D}
\end{bmatrix}
\label{eq:feature-state-init-msckf-6}
\end{equation}
with
\begin{align}
\boldsymbol{A} &= \boldsymbol{H}_o\boldsymbol{P}\boldsymbol{H}_o^T
+ \sigma^2\boldsymbol{I}
\label{eq:feature-state-init-msckf-7}\\
\boldsymbol{B} &= \boldsymbol{H}_o\boldsymbol{P}\boldsymbol{H}_1^T
\label{eq:feature-state-init-msckf-8}\\
\boldsymbol{C} &= \boldsymbol{H}_1\boldsymbol{P}\boldsymbol{H}_o^T
\label{eq:feature-state-init-msckf-9}\\
\boldsymbol{D} &= \boldsymbol{H}_1\boldsymbol{P}\boldsymbol{H}_1^T
+ \mu\boldsymbol{H}_2\boldsymbol{H}_2^T
+ \sigma^2\boldsymbol{I}
\label{eq:feature-state-init-msckf-10}
\end{align}
then if $\boldsymbol{D}$ and $\boldsymbol{A}-\boldsymbol{B}\boldsymbol{D}^{-1}
\boldsymbol{C}$ are non-singular, using block-wise inversion we get
\begin{align}
\boldsymbol{S}^{-1} &=
\begin{bmatrix}
\boldsymbol{A} & \boldsymbol{B} \\
\boldsymbol{C} & \boldsymbol{D}
\end{bmatrix}^{-1}
\label{eq:feature-state-init-msckf-11}
\\
&=
\begin{aligned}
&
\left[\begin{matrix}
\left(
\boldsymbol{A}-\boldsymbol{B}\boldsymbol{D}^{-1}\boldsymbol{C}
\right)^{-1}
\\
-\boldsymbol{D}^{-1}\boldsymbol{C}\left(
\boldsymbol{A}-\boldsymbol{B}\boldsymbol{D}^{-1}\boldsymbol{C}
\right)^{-1}
\end{matrix}\right.\\
&\qquad\qquad\qquad\qquad\left.
\begin{matrix}
-\left(
\boldsymbol{A}-\boldsymbol{B}\boldsymbol{D}^{-1}\boldsymbol{C}
\right)^{-1}\boldsymbol{B}\boldsymbol{D}^{-1}
\\
\boldsymbol{D}^{-1} +
\boldsymbol{D}^{-1}\boldsymbol{C}\left(
\boldsymbol{A}-\boldsymbol{B}\boldsymbol{D}^{-1}\boldsymbol{C}
\right)^{-1}\boldsymbol{B}\boldsymbol{D}^{-1}
\end{matrix}\right]
\end{aligned}
\label{eq:feature-state-init-msckf-12}
\\
&=
\begin{bmatrix}
\boldsymbol{F}_{11} & \boldsymbol{F}_{12} \\
\boldsymbol{F}_{21} & \boldsymbol{F}_{22}
\end{bmatrix}
\label{eq:feature-state-init-msckf-13}
\end{align}
with
\begin{align}
\boldsymbol{F}_{11} &=
\begin{aligned}
&
\bigg(
\boldsymbol{H}_o\boldsymbol{P}\boldsymbol{H}_o^T
+ \sigma^2\boldsymbol{I}\\
&\left.
- \boldsymbol{H}_o\boldsymbol{P}\boldsymbol{H}_1^T
\left(
\boldsymbol{H}_1\boldsymbol{P}\boldsymbol{H}_1^T
+ \mu\boldsymbol{H}_2\boldsymbol{H}_2^T
+ \sigma^2\boldsymbol{I}
\right)^{-1}
\boldsymbol{H}_1\boldsymbol{P}\boldsymbol{H}_o^T
\right)^{-1}
\end{aligned}
\label{eq:feature-state-init-msckf-14}
\\
\boldsymbol{F}_{12} &= -\boldsymbol{F}_{11}
\boldsymbol{H}_o\boldsymbol{P}\boldsymbol{H}_1^T
\left(
\boldsymbol{H}_1\boldsymbol{P}\boldsymbol{H}_1^T
+ \mu\boldsymbol{H}_2\boldsymbol{H}_2^T
+ \sigma^2\boldsymbol{I}
\right)^{-1}
\label{eq:feature-state-init-msckf-15}
\\
\boldsymbol{F}_{21} &= -\left(
\boldsymbol{H}_1\boldsymbol{P}\boldsymbol{H}_1^T
+ \mu\boldsymbol{H}_2\boldsymbol{H}_2^T
+ \sigma^2\boldsymbol{I}
\right)^{-1} \boldsymbol{H}_1\boldsymbol{P}\boldsymbol{H}_o^T\boldsymbol{F}_{11}
\label{eq:feature-state-init-msckf-16}
\end{align}
Since for a non-singular matrix $\boldsymbol{M}$, $(\boldsymbol{M}^T)^{-1} = 
(\boldsymbol{M}^{-1})^T$, we also get $\boldsymbol{F}_{12} =
\boldsymbol{F}_{21}^T$.

$\boldsymbol{F}_{22}$ is easier to derive if we consider the equivalent
block-wise inversion
\begin{equation}
\boldsymbol{S}^{-1} =
\begin{aligned}
&
\left[\begin{matrix}
\boldsymbol{A}^{-1} +
\boldsymbol{A}^{-1}\boldsymbol{B}\left(
\boldsymbol{D}-\boldsymbol{C}\boldsymbol{A}^{-1}\boldsymbol{B}
\right)^{-1}\boldsymbol{C}\boldsymbol{A}^{-1}
\\
-\left(
\boldsymbol{D}-\boldsymbol{C}\boldsymbol{A}^{-1}\boldsymbol{B}
\right)^{-1}\boldsymbol{C}\boldsymbol{A}^{-1}
\end{matrix}\right.\\
&\qquad\qquad\qquad\qquad\qquad\qquad\qquad\left.\begin{matrix}
-\boldsymbol{A}^{-1}\boldsymbol{B}\left(
\boldsymbol{D}-\boldsymbol{C}\boldsymbol{A}^{-1}\boldsymbol{B}
\right)^{-1}
\\
\left(
\boldsymbol{D}-\boldsymbol{C}\boldsymbol{A}^{-1}\boldsymbol{B}
\right)^{-1}
\end{matrix}\right]
\end{aligned}
\label{eq:feature-state-init-msckf-17}
\end{equation}
so that
\begin{equation}
\boldsymbol{F}_{22} = 
\begin{aligned}
&
\bigg(
\boldsymbol{H}_1\boldsymbol{P}\boldsymbol{H}_1^T
+ \mu\boldsymbol{H}_2\boldsymbol{H}_2^T
+ \sigma^2\boldsymbol{I} \\
&\left.
- \boldsymbol{H}_1\boldsymbol{P}\boldsymbol{H}_o^T
\left(\boldsymbol{H}_o\boldsymbol{P}\boldsymbol{H}_o^T
+ \sigma^2\boldsymbol{I}\right)^{-1}
\boldsymbol{H}_o\boldsymbol{P}\boldsymbol{H}_1^T
\right)^{-1}
\end{aligned}
\label{eq:feature-state-init-msckf-18}
\end{equation}

\subsection{Updated Error Covariance}

Let us define $\boldsymbol{N}$ from Equation~(26) in \citet{Li2012b}
\begin{align}
\boldsymbol{N} &=
\begin{aligned}
&
\begin{bmatrix}
\boldsymbol{P} & \boldsymbol{0} \\
\boldsymbol{0} & \mu\boldsymbol{I}
\end{bmatrix}
\begin{bmatrix}
\boldsymbol{H}_o & \boldsymbol{0} \\
\boldsymbol{H}_1 & \boldsymbol{H}_2
\end{bmatrix}^T
\begin{bmatrix}
\boldsymbol{F}_{11} & \boldsymbol{F}_{21}^T \\
\boldsymbol{F}_{21} & \boldsymbol{F}_{22}
\end{bmatrix}
\begin{bmatrix}
\boldsymbol{H}_o & \boldsymbol{0} \\
\boldsymbol{H}_1 & \boldsymbol{H}_2
\end{bmatrix}\\
&\qquad\qquad\qquad\qquad\qquad\qquad\qquad\qquad\qquad\qquad\qquad
\begin{bmatrix}
\boldsymbol{P} & \boldsymbol{0} \\
\boldsymbol{0} & \mu\boldsymbol{I}
\end{bmatrix}
\end{aligned}
\label{eq:feature-state-init-msckf-19}
\\
&=
\begin{bmatrix}
\boldsymbol{P} & \boldsymbol{0} \\
\boldsymbol{0} & \mu\boldsymbol{I}
\end{bmatrix}
\begin{bmatrix}
\boldsymbol{H}_o^T & \boldsymbol{H}_1^T \\
\boldsymbol{0} & \boldsymbol{H}_2^T
\end{bmatrix}
\begin{bmatrix}
\boldsymbol{F}_{11} & \boldsymbol{F}_{21}^T \\
\boldsymbol{F}_{21} & \boldsymbol{F}_{22}
\end{bmatrix}
\begin{bmatrix}
\boldsymbol{H}_o\boldsymbol{P} & \boldsymbol{0} \\
\boldsymbol{H}_1\boldsymbol{P} & \mu\boldsymbol{H}_2
\end{bmatrix}
\label{eq:feature-state-init-msckf-20}
\\
&=
\begin{bmatrix}
\boldsymbol{P}\boldsymbol{H}_o^T &
\boldsymbol{P}\boldsymbol{H}_1^T \\
\boldsymbol{0} &
\mu\boldsymbol{H}_2^T
\end{bmatrix}
\begin{bmatrix}
\boldsymbol{F}_{11}\boldsymbol{H}_o\boldsymbol{P} +
\boldsymbol{F}_{21}^T\boldsymbol{H}_1\boldsymbol{P} &
\mu\boldsymbol{F}_{21}^T\boldsymbol{H}_2 \\
\boldsymbol{F}_{21}\boldsymbol{H}_o\boldsymbol{P} +
\boldsymbol{F}_{22}\boldsymbol{H}_1\boldsymbol{P} &
\mu\boldsymbol{F}_{22}\boldsymbol{H}_2
\end{bmatrix}
\label{eq:feature-state-init-msckf-21}
\\
&=
\begin{bmatrix}
\boldsymbol{N}_{11} & \boldsymbol{N}_{12}
\\
\boldsymbol{N}_{21} &
\boldsymbol{N}_{22}
\end{bmatrix}
\label{eq:feature-state-init-msckf-22}
\end{align}
with
\begin{align}
\boldsymbol{N}_{11} &= 
\begin{aligned}
&
\boldsymbol{P}\boldsymbol{H}_o^T\boldsymbol{F}_{11}\boldsymbol{H}_o\boldsymbol{P}
+ \boldsymbol{P}\boldsymbol{H}_o^T\boldsymbol{F}_{21}^T\boldsymbol{H}_1\boldsymbol{P}
+ \boldsymbol{P}\boldsymbol{H}_1^T\boldsymbol{F}_{21}\boldsymbol{H}_o\boldsymbol{P} \\
&
+ \boldsymbol{P}\boldsymbol{H}_1^T\boldsymbol{F}_{22}\boldsymbol{H}_1\boldsymbol{P}
\end{aligned}
\label{eq:feature-state-init-msckf-23}
\\
\boldsymbol{N}_{12} &=
\mu\boldsymbol{P}\boldsymbol{H}_o^T\boldsymbol{F}_{21}^T\boldsymbol{H}_2 +
\mu\boldsymbol{P}\boldsymbol{H}_1^T\boldsymbol{F}_{22}\boldsymbol{H}_2
\label{eq:feature-state-init-msckf-24}
\\
\boldsymbol{N}_{21} &=
\mu\boldsymbol{H}_2^T\boldsymbol{F}_{21}\boldsymbol{H}_o\boldsymbol{P} +
\mu\boldsymbol{H}_2^T\boldsymbol{F}_{22}\boldsymbol{H}_1\boldsymbol{P}
\label{eq:feature-state-init-msckf-25}
\\
&= \boldsymbol{N}_{12}^T
\label{eq:feature-state-init-msckf-26}
\\
\boldsymbol{N}_{22} &=
\mu^2\boldsymbol{H}_2^T\boldsymbol{F}_{22}\boldsymbol{H}_2
\label{eq:feature-state-init-msckf-27}
\end{align}
which demonstrate Equation~(27).

The updated covariance matrix $P^+_{aug}$ is written as
\begin{align}
P^+_{aug} &=
\lim_{\mu\to\infty}
\begin{bmatrix}
\boldsymbol{P} - \boldsymbol{N}_{11} & -\boldsymbol{N}_{21}^T \\
-\boldsymbol{N}_{21} & \mu\boldsymbol{I} - \boldsymbol{N}_{22}
\end{bmatrix}
\label{eq:feature-state-init-msckf-28}
\\
&=
\begin{bmatrix}
\boldsymbol{P}^+ & {\boldsymbol{P}^+_{21}}^T \\
\boldsymbol{P}^+_{21} & \boldsymbol{P}^+_{22}
\end{bmatrix}\;.
\label{eq:feature-state-init-msckf-29}
\end{align}

We will now demonstrate the expressions of the various matrix blocks.

\subsubsection{Expression of $\boldsymbol{P}^+$}

\begin{align}
\boldsymbol{P}^+ &=
\begin{aligned}
&
\lim_{\mu\to\infty}
\left(
\boldsymbol{P} -
\left(
\boldsymbol{P}\boldsymbol{H}_o^T\boldsymbol{F}_{11}\boldsymbol{H}_o\boldsymbol{P}
+ \boldsymbol{P}\boldsymbol{H}_o^T\boldsymbol{F}_{21}^T\boldsymbol{H}_1\boldsymbol{P}
\right.\right.\\
&
\left.\left.
+ \boldsymbol{P}\boldsymbol{H}_1^T\boldsymbol{F}_{21}\boldsymbol{H}_o\boldsymbol{P}
+ \boldsymbol{P}\boldsymbol{H}_1^T\boldsymbol{F}_{22}\boldsymbol{H}_1\boldsymbol{P}
\right)
\right)
\end{aligned}
\label{eq:feature-state-init-msckf-30}
\\
&= \boldsymbol{P} -
\boldsymbol{P}\boldsymbol{H}_o^T
\left(\boldsymbol{H}_o\boldsymbol{P}\boldsymbol{H}_o^T
+ \sigma^2\boldsymbol{I}\right)^{-1}
\boldsymbol{H}_o\boldsymbol{P}
\label{eq:feature-state-init-msckf-31}
\end{align}
which comes trivial from Equations~(22), (23) and (24).

\subsubsection{Expression of $\boldsymbol{P}^+_{21}$}

Since $\lim_{\mu\to\infty}\boldsymbol{F}_{21}=
\lim_{\mu\to\infty}\boldsymbol{F}_{22}=\boldsymbol{0}$, the terms
$\mu\boldsymbol{F}_{21}$ and $\mu\boldsymbol{F}_{22}$ in $\boldsymbol{N}_{21}$
raise the limit indeterminate form $0 \times \infty$. To solve that, let us
note that for a non-singular matrix $\boldsymbol{M}$ and a scalar $\mu$,
$\mu\boldsymbol{M}^{-1} = (\frac{1}{\mu}\boldsymbol{M})^{-1}$. Then
\begin{align}
\lim_{\mu\to\infty}\mu\boldsymbol{F}_{21} &=
\begin{aligned}
&
\lim_{\mu\to\infty}
\left(
-\mu\left(
\boldsymbol{H}_1\boldsymbol{P}\boldsymbol{H}_1^T
+ \mu\boldsymbol{H}_2\boldsymbol{H}_2^T
+ \sigma^2\boldsymbol{I}
\right)^{-1} \right.\\
&
\boldsymbol{H}_1\boldsymbol{P}\boldsymbol{H}_o^T\boldsymbol{F}_{11}
\bigg)
\end{aligned}
\label{eq:feature-state-init-msckf-32}
\\
&=
\begin{aligned}
&
\lim_{\mu\to\infty}
\left(
-\left(
\frac{1}{\mu}\boldsymbol{H}_1\boldsymbol{P}\boldsymbol{H}_1^T
+ \boldsymbol{H}_2\boldsymbol{H}_2^T
+ \frac{1}{\mu}\sigma^2\boldsymbol{I}
\right)^{-1} \right.\\
&
\boldsymbol{H}_1\boldsymbol{P}\boldsymbol{H}_o^T\boldsymbol{F}_{11}
\Bigg)
\end{aligned}
\label{eq:feature-state-init-msckf-33}
\\
&=
-\left(
\boldsymbol{H}_2\boldsymbol{H}_2^T
\right)^{-1} \boldsymbol{H}_1\boldsymbol{P}\boldsymbol{H}_o^T
\left(\boldsymbol{H}_o\boldsymbol{P}\boldsymbol{H}_o^T
+ \sigma^2\boldsymbol{I}\right)^{-1}
\label{eq:feature-state-init-msckf-34}
\end{align}
and
\begin{align}
\lim_{\mu\to\infty}\mu\boldsymbol{F}_{22} &=
\begin{aligned}
&
\lim_{\mu\to\infty}
\Bigg(
\mu\bigg(
\boldsymbol{H}_1\boldsymbol{P}\boldsymbol{H}_1^T
+ \mu\boldsymbol{H}_2\boldsymbol{H}_2^T
+ \sigma^2\boldsymbol{I} \\
&\left.\left.
- \boldsymbol{H}_1\boldsymbol{P}\boldsymbol{H}_o^T
\left(\boldsymbol{H}_o\boldsymbol{P}\boldsymbol{H}_o^T
+ \sigma^2\boldsymbol{I}\right)^{-1}
\boldsymbol{H}_o\boldsymbol{P}\boldsymbol{H}_1^T
\right)^{-1}
\right)
\end{aligned}
\label{eq:feature-state-init-msckf-35}
\\
&=
\begin{aligned}
&
\lim_{\mu\to\infty}
\Bigg(
\left(
\frac{1}{\mu}\boldsymbol{H}_1\boldsymbol{P}\boldsymbol{H}_1^T
+ \boldsymbol{H}_2\boldsymbol{H}_2^T
+ \frac{1}{\mu}\sigma^2\boldsymbol{I} \right.\\
&\left.\left.
- \frac{1}{\mu}\boldsymbol{H}_1\boldsymbol{P}\boldsymbol{H}_o^T
\left(\boldsymbol{H}_o\boldsymbol{P}\boldsymbol{H}_o^T
+ \sigma^2\boldsymbol{I}\right)^{-1}
\boldsymbol{H}_o\boldsymbol{P}\boldsymbol{H}_1^T
\right)^{-1}
\right)
\end{aligned}
\label{eq:feature-state-init-msckf-36}
\\
&=
\left(
\boldsymbol{H}_2\boldsymbol{H}_2^T
\right)^{-1}
\label{eq:feature-state-init-msckf-37}
\end{align}

Hence
\begin{align}
\boldsymbol{P}^+_{21} &= \lim_{\mu\to\infty}
-\boldsymbol{N}_{21}
\label{eq:feature-state-init-msckf-38}
\\
&=
\begin{aligned}
&
\boldsymbol{H}_2^T
\left(
\boldsymbol{H}_2\boldsymbol{H}_2^T
\right)^{-1} \boldsymbol{H}_1\boldsymbol{P}\boldsymbol{H}_o^T
\left(\boldsymbol{H}_o\boldsymbol{P}\boldsymbol{H}_o^T
+ \sigma^2\boldsymbol{I}\right)^{-1}
\boldsymbol{H}_o\boldsymbol{P} \\
&
-
\boldsymbol{H}_2^T
\left(
\boldsymbol{H}_2\boldsymbol{H}_2^T
\right)^{-1}
\boldsymbol{H}_1\boldsymbol{P}
\end{aligned}
\label{eq:feature-state-init-msckf-39}
\\
&=
\boldsymbol{H}_2^{-1}\boldsymbol{H}_1\boldsymbol{P}\boldsymbol{H}_o^T
\left(\boldsymbol{H}_o\boldsymbol{P}\boldsymbol{H}_o^T
+ \sigma^2\boldsymbol{I}\right)^{-1}
\boldsymbol{H}_o\boldsymbol{P} -
\boldsymbol{H}_2^{-1}
\boldsymbol{H}_1\boldsymbol{P}
\label{eq:feature-state-init-msckf-40}
\\
&=
\boldsymbol{H}_2^{-1}\boldsymbol{H}_1
\left(
\boldsymbol{P}\boldsymbol{H}_o^T
\left(\boldsymbol{H}_o\boldsymbol{P}\boldsymbol{H}_o^T
+ \sigma^2\boldsymbol{I}\right)^{-1}
\boldsymbol{H}_o\boldsymbol{P} -
\boldsymbol{P}
\right)
\label{eq:feature-state-init-msckf-42}
\\
&=
-\boldsymbol{H}_2^{-1}\boldsymbol{H}_1P^+
\label{eq:feature-state-init-msckf-43}
\end{align}

\subsubsection{Expression of $\boldsymbol{P}^+_{22}$}

\begin{equation}
\boldsymbol{P}^+_{22} = \lim_{\mu\to\infty}
\left(
\mu\boldsymbol{I} - \boldsymbol{N}_{22}
\right)
\label{eq:feature-state-init-msckf-44}
\end{equation}
where
\begin{align}
\mu\boldsymbol{I} - \boldsymbol{N}_{22} &=
\mu\boldsymbol{I} - \mu^2\boldsymbol{H}_2^T\boldsymbol{F}_{22}\boldsymbol{H}_2
\label{eq:feature-state-init-msckf-45}
\\
&=
\mu\boldsymbol{H}_2^T\boldsymbol{F}_{22}\boldsymbol{H}_2
\left(
\left(\boldsymbol{H}_2^T\boldsymbol{F}_{22}\boldsymbol{H}_2\right)^{-1}
- \mu\boldsymbol{I}
\right)
\label{eq:feature-state-init-msckf-46}
\\
&=
\mu\boldsymbol{H}_2^T\boldsymbol{F}_{22}\boldsymbol{H}_2
\left(
\boldsymbol{H}_2^{-1}\boldsymbol{F}_{22}^{-1}\boldsymbol{H}_2^{-T}
- \mu\boldsymbol{I}
\right)
\label{eq:feature-state-init-msckf-47}
\\
&=
\mu\boldsymbol{H}_2^T\boldsymbol{F}_{22}\boldsymbol{H}_2
\left(
\boldsymbol{H}_2^{-1}
\left(\boldsymbol{F}_{22}^{-1} -\mu\boldsymbol{H}_2\boldsymbol{H}_2^T\right)
\boldsymbol{H}_2^{-T}
\right)
\label{eq:feature-state-init-msckf-48}
\\
&=
\begin{aligned}
&
\mu\boldsymbol{H}_2^T\boldsymbol{F}_{22}\boldsymbol{H}_2
\bigg(
\boldsymbol{H}_2^{-1}
\bigg(
\boldsymbol{H}_1\boldsymbol{P}\boldsymbol{H}_1^T
+ \sigma^2\boldsymbol{I} \\
&\left.\left.
- \boldsymbol{H}_1\boldsymbol{P}\boldsymbol{H}_o^T
\left(\boldsymbol{H}_o\boldsymbol{P}\boldsymbol{H}_o^T
+ \sigma^2\boldsymbol{I}\right)^{-1}
\boldsymbol{H}_o\boldsymbol{P}\boldsymbol{H}_1^T
\right)
\boldsymbol{H}_2^{-T}
\right)
\end{aligned}
\label{eq:feature-state-init-msckf-49}
\end{align}

Using Equation~(\ref{eq:feature-state-init-msckf-37}) in the current
demonstration, one can then write
\begin{align}
\boldsymbol{P}^+_{22} &=
\begin{aligned}
&
\boldsymbol{H}_2^T\left(
\boldsymbol{H}_2\boldsymbol{H}_2^T
\right)^{-1}\boldsymbol{H}_2
\bigg(
\boldsymbol{H}_2^{-1}
\bigg(
\boldsymbol{H}_1\boldsymbol{P}\boldsymbol{H}_1^T
+ \sigma^2\boldsymbol{I} \\
&\left.\left.
- \boldsymbol{H}_1\boldsymbol{P}\boldsymbol{H}_o^T
\left(\boldsymbol{H}_o\boldsymbol{P}\boldsymbol{H}_o^T
+ \sigma^2\boldsymbol{I}\right)^{-1}
\boldsymbol{H}_o\boldsymbol{P}\boldsymbol{H}_1^T
\right)
\boldsymbol{H}_2^{-T}
\right)
\end{aligned}
\label{eq:feature-state-init-msckf-50}
\\
&=
\begin{aligned}
&
\boldsymbol{H}_2^T\boldsymbol{H}_2^{-T}
\boldsymbol{H}_2^{-1}\boldsymbol{H}_2
\bigg(
\boldsymbol{H}_2^{-1}
\bigg(
\boldsymbol{H}_1\boldsymbol{P}\boldsymbol{H}_1^T
+ \sigma^2\boldsymbol{I} \\
&\left.\left.
- \boldsymbol{H}_1\boldsymbol{P}\boldsymbol{H}_o^T
\left(\boldsymbol{H}_o\boldsymbol{P}\boldsymbol{H}_o^T
+ \sigma^2\boldsymbol{I}\right)^{-1}
\boldsymbol{H}_o\boldsymbol{P}\boldsymbol{H}_1^T
\right)
\boldsymbol{H}_2^{-T}
\right)
\end{aligned}
\label{eq:feature-state-init-msckf-51}
\\
&=
\begin{aligned}
&
\boldsymbol{H}_2^{-1}
\bigg(
\boldsymbol{H}_1\boldsymbol{P}\boldsymbol{H}_1^T
+ \sigma^2\boldsymbol{I} \\
&\left.
- \boldsymbol{H}_1\boldsymbol{P}\boldsymbol{H}_o^T
\left(\boldsymbol{H}_o\boldsymbol{P}\boldsymbol{H}_o^T
+ \sigma^2\boldsymbol{I}\right)^{-1}
\boldsymbol{H}_o\boldsymbol{P}\boldsymbol{H}_1^T
\right)
\boldsymbol{H}_2^{-T}
\end{aligned}
\label{eq:feature-state-init-msckf-52}
\\
&=
\begin{aligned}
&
\boldsymbol{H}_2^{-1}
\bigg(
\boldsymbol{H}_1\boldsymbol{P}\boldsymbol{H}_1^T \\
&\left.
- \boldsymbol{H}_1\boldsymbol{P}\boldsymbol{H}_o^T
\left(\boldsymbol{H}_o\boldsymbol{P}\boldsymbol{H}_o^T
+ \sigma^2\boldsymbol{I}\right)^{-1}
\boldsymbol{H}_o\boldsymbol{P}\boldsymbol{H}_1^T
\right)
\boldsymbol{H}_2^{-T} \\
&
+ \sigma^2\boldsymbol{H}_2^{-1}\boldsymbol{H}_2^{-T}
\end{aligned}
\label{eq:feature-state-init-msckf-53}
\\
&=
\begin{aligned}
&
\boldsymbol{H}_2^{-1}\boldsymbol{H}_1
\left(
\boldsymbol{P}
- \boldsymbol{P}\boldsymbol{H}_o^T
\left(\boldsymbol{H}_o\boldsymbol{P}\boldsymbol{H}_o^T
+ \sigma^2\boldsymbol{I}\right)^{-1}
\boldsymbol{H}_o\boldsymbol{P}
\right)
\boldsymbol{H}_1^T\boldsymbol{H}_2^{-T} \\
&
+ \sigma^2\boldsymbol{H}_2^{-1}\boldsymbol{H}_2^{-T}
\end{aligned}
\label{eq:feature-state-init-msckf-54}
\\
&=
\boldsymbol{H}_2^{-1}\boldsymbol{H}_1\boldsymbol{P}^+
\boldsymbol{H}_1^T\boldsymbol{H}_2^{-T}
+ \sigma^2\boldsymbol{H}_2^{-1}\boldsymbol{H}_2^{-T}
\end{align}

\subsection{State Update}

Let us define $\boldsymbol{O}$ from Equation~(32) in \citet{Li2012b}
\begin{align}
\boldsymbol{O} &=
\begin{bmatrix}
\boldsymbol{P} & \boldsymbol{0} \\
\boldsymbol{0} & \mu\boldsymbol{I}
\end{bmatrix}
\begin{bmatrix}
\boldsymbol{H}_o & \boldsymbol{0} \\
\boldsymbol{H}_1 & \boldsymbol{H}_2
\end{bmatrix}^T
\begin{bmatrix}
\boldsymbol{F}_{11} & \boldsymbol{F}_{21}^T \\
\boldsymbol{F}_{21} & \boldsymbol{F}_{22}
\end{bmatrix}
\label{eq:feature-state-init-msckf-55}
\\
&=
\begin{bmatrix}
\boldsymbol{P}\boldsymbol{H}_o^T &
\boldsymbol{P}\boldsymbol{H}_1^T \\
\boldsymbol{0} &
\mu\boldsymbol{H}_2^T
\end{bmatrix}
\begin{bmatrix}
\boldsymbol{F}_{11} & \boldsymbol{F}_{21}^T \\
\boldsymbol{F}_{21} & \boldsymbol{F}_{22}
\end{bmatrix}
\label{eq:feature-state-init-msckf-56}
\\
&=
\begin{bmatrix}
\boldsymbol{P}\boldsymbol{H}_o^T\boldsymbol{F}_{11} +
\boldsymbol{P}\boldsymbol{H}_1^T\boldsymbol{F}_{21} &
\boldsymbol{P}\boldsymbol{H}_o^T\boldsymbol{F}_{21}^T +
\boldsymbol{P}\boldsymbol{H}_1^T\boldsymbol{F}_{22} \\
\mu\boldsymbol{H}_2^T\boldsymbol{F}_{21} &
\mu\boldsymbol{H}_2^T\boldsymbol{F}_{22}
\end{bmatrix}
\label{eq:feature-state-init-msckf-57}
\end{align}

Using Equations~(22)-(24) in \citet{Li2012b}, along with
(\ref{eq:feature-state-init-msckf-34}) and
(\ref{eq:feature-state-init-msckf-37})
in this demonstration, one gets
\begin{align}
\lim_{\mu\to\infty}\boldsymbol{O} &=
\begin{aligned}
&
\lim_{\mu\to\infty}
\left[\begin{matrix}
\boldsymbol{P}\boldsymbol{H}_o^T\boldsymbol{F}_{11}
\\
-\boldsymbol{H}_2^T\left(
\boldsymbol{H}_2\boldsymbol{H}_2^T
\right)^{-1} \boldsymbol{H}_1\boldsymbol{P}\boldsymbol{H}_o^T
\left(\boldsymbol{H}_o\boldsymbol{P}\boldsymbol{H}_o^T
+ \sigma^2\boldsymbol{I}\right)^{-1}
\end{matrix}\right.
\\
&\qquad\qquad\qquad\qquad\qquad\qquad\qquad\qquad
\left.\begin{matrix}
\boldsymbol{0}
\\
\boldsymbol{H}_2^T\left(
\boldsymbol{H}_2\boldsymbol{H}_2^T
\right)^{-1}
\end{matrix}\right]
\end{aligned}
\label{eq:feature-state-init-msckf-58}
\\
&= \lim_{\mu\to\infty}
\begin{bmatrix}
\boldsymbol{P}\boldsymbol{H}_o^T\boldsymbol{F}_{11} &
\boldsymbol{0} \\
-\boldsymbol{H}_2^{-1} \boldsymbol{H}_1\boldsymbol{P}\boldsymbol{H}_o^T
\boldsymbol{F}_{11} &
\boldsymbol{H}_2^{-1}
\end{bmatrix}
\label{eq:feature-state-init-msckf-59}
\end{align}
which brings Equation~(33) in \citet{Li2012b}.

\section{Feature State Reparametrization}
\label{sec:feature-reparam}

Equations~(\ref{eq:covariance-reparam})--(\ref{eq:reparam-jacobian-feature})
provide the expression of the error covariance when SLAM feature
$\boldsymbol{p}_j$ is reparametrized from anchor pose state
$\left\{c_{i_1}\right\}$ to $\left\{c_{i_2}\right\}$ in the sliding window.

If $\boldsymbol{P}_2$ and $\delta\boldsymbol{x}_2$ respectively denote the
error covariance and error state after the reparametrization, then by definition
\begin{equation}
\boldsymbol{P}_2 =
E \left[\delta\boldsymbol{x}_2 {\delta\boldsymbol{x}_2}^T\right]
\label{eq:cov-reparam-1}
\end{equation}
where the operator $E$ is the expected value. If we further denote
$\boldsymbol{J}_j =
\frac{\partial{\boldsymbol{x}_2}}{\partial{\boldsymbol{x}_1}}$
the Jacobian of the reparametrization, then at first order
\begin{equation}
\delta\boldsymbol{x}_2 \simeq \boldsymbol{J}_j \delta\boldsymbol{x}_1
\label{eq:cov-reparam-2}
\end{equation}
and
\begin{align}
\boldsymbol{P}_2 &\simeq
E \left[\boldsymbol{J}_j \delta\boldsymbol{x}_1
{\delta\boldsymbol{x}_1}^T \boldsymbol{J}_j^T\right]
\label{eq:cov-reparam-3}\\
&\simeq
\boldsymbol{J}_j E \left[\delta\boldsymbol{x}_1
{\delta\boldsymbol{x}_1}^T \right] \boldsymbol{J}_j^T
\label{eq:cov-reparam-4}\\
&\simeq
\boldsymbol{J}_j \boldsymbol{P}_1 \boldsymbol{J}_j^T\;,
\label{eq:cov-reparam-5}
\end{align}
which demonstrates Equation~(\ref{eq:covariance-reparam}).

We define the vector $\boldsymbol{g}_j$ from
Equation~(\ref{eq:inverse-depth-param-conv})
\begin{equation}
\boldsymbol{g}_j = 
\begin{aligned}
&
\frac{1}{\rho_{j_2}}
\begin{bmatrix} \alpha_{j_2} \\ \beta_{j_2} \\ 1\end{bmatrix} =
\boldsymbol{C}(\boldsymbol{q}_w^{c_{i_2}})
\Bigg(
- \boldsymbol{p}_w^{c_{i_2}} +
\boldsymbol{p}_w^{c_{i_1}} \\
&
+ \frac{1}{\rho_{j_1}} \boldsymbol{C}(\boldsymbol{q}_w^{c_{i_1}})^T
\begin{bmatrix} \alpha_{j_1} \\ \beta_{j_1} \\ 1\end{bmatrix}
\Bigg)
\end{aligned}
\label{eq:cov-reparam-6}
\end{equation}
At first order, by differentiating on both sides, we get
\begin{align}
\delta\boldsymbol{g}_j \simeq
\begin{bmatrix}
\frac{\partial{\big(\frac{\alpha_{j_2}}{\rho_{j_2}}}\big)}{\partial{\alpha_{j_2}}} &
\frac{\partial{\big(\frac{\alpha_{j_2}}{\rho_{j_2}}}\big)}{\partial{\beta_{j_2}}} &
\frac{\partial{\big(\frac{\alpha_{j_2}}{\rho_{j_2}}}\big)}{\partial{\rho_{j_2}}} \\
\frac{\partial{\big(\frac{\beta_{j_2}}{\rho_{j_2}}}\big)}{\partial{\alpha_{j_2}}} &
\frac{\partial{\big(\frac{\beta_{j_2}}{\rho_{j_2}}}\big)}{\partial{\beta_{j_2}}} &
\frac{\partial{\big(\frac{\beta_{j_2}}{\rho_{j_2}}}\big)}{\partial{\rho_{j_2}}} \\
\frac{\partial{\big(\frac{1}{\rho_{j_2}}}\big)}{\partial{\alpha_{j_2}}} &
\frac{\partial{\big(\frac{1}{\rho_{j_2}}}\big)}{\partial{\beta_{j_2}}} &
\frac{\partial{\big(\frac{1}{\rho_{j_2}}}\big)}{\partial{\rho_{j_2}}}
\end{bmatrix}
\begin{bmatrix}
\delta\alpha_{j_2} \\ \delta\beta_{j_2} \\ \delta\rho_{j_2}
\end{bmatrix}
& \simeq
\underbrace{\frac{\partial{\boldsymbol{g}_j}}{\partial{\boldsymbol{x}_1}}}_{\boldsymbol{H}_{1_j}}
\delta\boldsymbol{x}_1
\label{eq:cov-reparam-7}\\
\begin{bmatrix}
\frac{1}{\hat{\rho}_{j_2}} &
0 &
- \frac{\hat{\alpha}_{j_2}}{\hat{\rho}_{j_2}^2} \\
0 &
\frac{1}{\hat{\rho}_{j_2}} &
- \frac{\hat{\beta}_{j_2}}{\hat{\rho}_{j_2}^2} \\
0 &
0 &
- \frac{1}{\hat{\rho}_{j_2}^2}
\end{bmatrix}
\begin{bmatrix}
\delta\alpha_{j_2} \\ \delta\beta_{j_2} \\ \delta\rho_{j_2}
\end{bmatrix}
& \simeq
\boldsymbol{H}_{1_j}\delta\boldsymbol{x}_1
\label{eq:cov-reparam-8}\\
\begin{bmatrix}
\frac{1}{\hat{\rho}_{j_2}}\delta\alpha_{j_2}
- \frac{\hat{\alpha}_{j_2}}{\hat{\rho}_{j_2}^2} \delta\rho_{j_2} \\
\frac{1}{\hat{\rho}_{j_2}}\delta\beta_{j_2}
- \frac{\hat{\alpha}_{j_2}}{\hat{\rho}_{j_2}^2} \delta\rho_{j_2} \\
- \frac{1}{\hat{\rho}_{j_2}^2} \delta\rho_{j_2}
\end{bmatrix}
& \simeq
\boldsymbol{H}_{1_j}\delta\boldsymbol{x}_1
\label{eq:cov-reparam-9}
\end{align}
The third row gives
\begin{equation}
\delta\rho_{j_2} \simeq - \hat{\rho}_{j_2}^2
\begin{bmatrix} 0 & 0 & 1 \end{bmatrix}
\boldsymbol{H}_{1_j}\delta\boldsymbol{x}_1
\label{eq:cov-reparam-10}
\end{equation}
which can be injected the first two rows, so that
\begin{align}
&\begin{cases}
\frac{1}{\hat{\rho}_{j_2}}\delta\alpha_{j_2}
+ \hat{\alpha}_{j_2} \begin{bmatrix} 0 & 0 & 1 \end{bmatrix}
\boldsymbol{H}_{1_j}\delta\boldsymbol{x}_1
\simeq \begin{bmatrix} 1 & 0 & 0 \end{bmatrix}
\boldsymbol{H}_{1_j}\delta\boldsymbol{x}_1 \\
\frac{1}{\hat{\rho}_{j_2}}\delta\beta_{j_2}
+ \hat{\alpha}_{j_2} \begin{bmatrix} 0 & 1 & 0 \end{bmatrix}
\boldsymbol{H}_{1_j}\delta\boldsymbol{x}_1
\simeq \begin{bmatrix} 0 & 1 & 0 \end{bmatrix}
\boldsymbol{H}_{1_j}\delta\boldsymbol{x}_1
\end{cases}
\label{eq:cov-reparam-11}\\
&\begin{cases}
\delta\alpha_{j_2} \simeq \hat{\rho}_{j_2}
\begin{bmatrix} 1 & 0 & -\hat{\alpha}_{j_2} \end{bmatrix}
\boldsymbol{H}_{1_j}\delta\boldsymbol{x}_1 \\
\delta\beta_{j_2} \simeq \hat{\rho}_{j_2}
\begin{bmatrix} 0 & 1 & -\hat{\alpha}_{j_2} \end{bmatrix}
\boldsymbol{H}_{1_j}\delta\boldsymbol{x}_1
\end{cases}
\label{eq:cov-reparam-12}
\end{align}

By identifying Equations~(\ref{eq:cov-reparam-10}) and (\ref{eq:cov-reparam-12})
in Equation~(\ref{eq:cov-reparam-2}), we can write
\begin{equation}
\boldsymbol{J}_j = \left[\;\begin{matrix}

\boldsymbol{I}_{15+6M+3(j-1)} &
\boldsymbol{0}
\vspace{2pt}
\\
\hline
 &
\hspace{-70pt}\raisebox{-2pt}{\mbox{$\boldsymbol{J}_{1_j}\boldsymbol{H}_{1_j}$}}
\vspace{2pt}
\\
\hline
\raisebox{-2pt}{\mbox{$\boldsymbol{0}$}} &
\raisebox{-2pt}{\mbox{$\boldsymbol{I}_{3(N-j)}$}}
\end{matrix}\;\right]\;,
\label{eq:covariance-augmentation-jacobian-bis}
\end{equation}
with
\begin{equation}
\boldsymbol{J}_{1_j} = \hat{\rho}_{j_2}
\begin{bmatrix}
1 & 0 & -\hat{\alpha}_{j_2} \\
0 & 1 & -\hat{\beta}_{j_2} \\
0 & 0 & -\hat{\rho}_{j_2}
\end{bmatrix}\;,
\label{eq:reparam-jacobian-prefix-bis}
\end{equation}
which are Equations~(\ref{eq:covariance-augmentation-jacobian}) and
(\ref{eq:reparam-jacobian-prefix}), respectively.

Now we proceed to deriving the non-null components of $\boldsymbol{H}_{1_j}$
when $i_1 = 1$ and $i_2 = M$.

\begin{align}
\frac{\partial{\boldsymbol{g}_j}}
{\partial{\boldsymbol{p}_w^{c_1}}} &=
\begin{aligned}
&
\frac{\partial{}}
{\partial{\boldsymbol{p}_w^{c_1}}}\left(
\boldsymbol{C}(\boldsymbol{q}_w^{c_M})
\Bigg(
- {\boldsymbol{p}_w^{c_M}} +
{\boldsymbol{p}_w^{c_1}} \right.\\
&
+ \frac{1}{\rho_{j_1}} \boldsymbol{C}(\boldsymbol{q}_w^{c_1})^T
\begin{bmatrix} \alpha_{j_1} \\ \beta_{j_1} \\ 1\end{bmatrix}
\Bigg)
\Bigg)
\end{aligned}
\label{eq:cov-reparam-13}\\
&=
\boldsymbol{C}(\hat{\boldsymbol{q}}_w^{c_M})
\label{eq:cov-reparam-14}\\
&=
{^j\boldsymbol{H}}_{\boldsymbol{p}_1}
\label{eq:cov-reparam-15}
\\[20pt]
\frac{\partial{\boldsymbol{g}_j}}
{\partial{\boldsymbol{p}_w^{c_M}}} &=
- \boldsymbol{C}(\hat{\boldsymbol{q}}_w^{c_M})
\label{eq:cov-reparam-16}\\
&=
{^j\boldsymbol{H}}_{\boldsymbol{p}_M}
\label{eq:cov-reparam-17}
\end{align}

Using the small angle assumption like in
Equation~(\ref{eq:feature-frame-tf2-lin-2}), we can write
\begin{align}
\frac{\partial{\boldsymbol{g}_j}}
{\partial{\delta\boldsymbol{\theta}_w^{c_1}}} &\simeq
\begin{aligned}
&
\frac{\partial{}}
{\partial{\delta\boldsymbol{\theta}_w^{c_1}}}
\Bigg(
(\boldsymbol{I}_3 - \lfloor\delta\boldsymbol{\theta}_w^{c_M}\times\rfloor)
\boldsymbol{C}(\hat{\boldsymbol{q}}_w^{c_M})
\Bigg(- \boldsymbol{p}_w^{c_M}
+ \boldsymbol{p}_w^{c_1} \\
&
+ \frac{1}{\rho_{j_1}}
\boldsymbol{C}(\hat{\boldsymbol{q}}_w^{c_1})^T
(\boldsymbol{I}_3 + \lfloor\delta\boldsymbol{\theta}_w^{c_1}\times\rfloor)
\begin{bmatrix} \alpha_{j_1} \\ \beta_{j_1} \\ 1\end{bmatrix}
\Bigg)\Bigg)
\end{aligned}
\label{eq:cov-reparam-18}\\
&\simeq
\begin{aligned}
&
\frac{\partial{}}
{\partial{\delta\boldsymbol{\theta}_w^{c_1}}}
\Bigg(
(\boldsymbol{I}_3 - \lfloor\delta\boldsymbol{\theta}_w^{c_M}\times\rfloor)
\boldsymbol{C}(\hat{\boldsymbol{q}}_w^{c_M}) \\
&\qquad\qquad\qquad\qquad\qquad\qquad
\frac{1}{\rho_{j_1}}
\boldsymbol{C}(\hat{\boldsymbol{q}}_w^{c_1})^T
\lfloor\delta\boldsymbol{\theta}_w^{c_1}\times\rfloor
\begin{bmatrix} \alpha_{j_1} \\ \beta_{j_1} \\ 1\end{bmatrix}
\Bigg)
\end{aligned}
\label{eq:cov-reparam-19}\\
&\simeq
\begin{aligned}
&
\frac{\partial{}}
{\partial{\delta\boldsymbol{\theta}_w^{c_1}}}
\Bigg( - \frac{1}{\rho_{j_1}}
(\boldsymbol{I}_3 - \lfloor\delta\boldsymbol{\theta}_w^{c_M}\times\rfloor)
\boldsymbol{C}(\hat{\boldsymbol{q}}_w^{c_M}) \\
&\qquad\qquad\qquad\qquad\qquad\qquad
\boldsymbol{C}(\hat{\boldsymbol{q}}_w^{c_1})^T
\left\lfloor
\begin{bmatrix} \alpha_{j_1} \\ \beta_{j_1} \\ 1\end{bmatrix}
\times\right\rfloor
\delta\boldsymbol{\theta}_w^{c_1}
\Bigg)
\end{aligned}
\label{eq:cov-reparam-20}\\
&\simeq
 - \frac{1}{\hat{\rho}_{j_1}}
\boldsymbol{C}(\hat{\boldsymbol{q}}_w^{c_M})
\boldsymbol{C}(\hat{\boldsymbol{q}}_w^{c_1})^T
\left\lfloor
\begin{bmatrix} \hat{\alpha}_{j_1} \\ \hat{\beta}_{j_1} \\ 1
\end{bmatrix}
\times\right\rfloor
\label{eq:cov-reparam-21}\\
&\simeq {^j\boldsymbol{H}}_{\boldsymbol{\theta}_1}
\label{eq:cov-reparam-22}
\end{align}

\begin{align}
\frac{\partial{\boldsymbol{g}_j}}
{\partial{\delta\boldsymbol{\theta}_w^{c_M}}} &\simeq
\begin{aligned}
&
\frac{\partial{}}
{\partial{\delta\boldsymbol{\theta}_w^{c_M}}}
\Bigg(
- \lfloor\delta\boldsymbol{\theta}_w^{c_M}\times\rfloor
\boldsymbol{C}(\hat{\boldsymbol{q}}_w^{c_M})
\Bigg( - \boldsymbol{p}_w^{c_M}
+ \boldsymbol{p}_w^{c_1} \\
&
+ \frac{1}{\rho_{j_1}}
\boldsymbol{C}(\hat{\boldsymbol{q}}_w^{c_1})^T
(\boldsymbol{I}_3 + \lfloor\delta\boldsymbol{\theta}_w^{c_1}\times\rfloor)
\begin{bmatrix} \alpha_{j_1} \\ \beta_{j_1} \\ 1\end{bmatrix}
\Bigg)\Bigg)
\end{aligned}
\label{eq:cov-reparam-23}\\
&\simeq
\begin{aligned}
&
\frac{\partial{}}
{\partial{\delta\boldsymbol{\theta}_w^{c_M}}}
\Bigg(
\Biggl\lfloor
\boldsymbol{C}(\hat{\boldsymbol{q}}_w^{c_M})
\Bigg( - \boldsymbol{p}_w^{c_M}
+ \boldsymbol{p}_w^{c_1} \\
&
+ \frac{1}{\rho_{j_1}}
\boldsymbol{C}(\hat{\boldsymbol{q}}_w^{c_1})^T
(\boldsymbol{I}_3 + \lfloor\delta\boldsymbol{\theta}_w^{c_1}\times\rfloor)
\begin{bmatrix} \alpha_{j_1} \\ \beta_{j_1} \\ 1\end{bmatrix}
\Bigg)
\times\Biggl\rfloor
\delta\boldsymbol{\theta}_w^{c_M}
\Bigg)
\end{aligned}
\label{eq:cov-reparam-24}\\
&\simeq \left\lfloor
\boldsymbol{C}(\hat{\boldsymbol{q}}_w^{c_M})
\left( - \hat{\boldsymbol{p}}_w^{c_M}
+ \hat{\boldsymbol{p}}_w^{c_1} + \frac{1}{\hat{\rho}_{j_1}}
\boldsymbol{C}(\hat{\boldsymbol{q}}_w^{c_1})^T
\begin{bmatrix} \hat{\alpha}_{j_1} \\ \hat{\beta}_{j_1} \\ 1\end{bmatrix}
\right)
\times\right\rfloor
\label{eq:cov-reparam-25}\\
&\simeq {^j\boldsymbol{H}}_{\boldsymbol{\theta}_M}
\label{eq:cov-reparam-26}
\end{align}

\begin{align}
\frac{\partial{\boldsymbol{g}_j}}
{\partial{\alpha_{j_1}}} &=
\frac{\partial{}}
{\partial{\alpha_{j_1}}}\left(
\frac{1}{\rho_{j_1}}
\boldsymbol{C}(\boldsymbol{q}_w^{c_M})
\boldsymbol{C}(\boldsymbol{q}_w^{c_1})^T
\begin{bmatrix} \alpha_{j_1} \\ \beta_{j_1} \\ 1\end{bmatrix}
\right)
\label{eq:cov-reparam-27}
\\
&=
\frac{1}{\hat{\rho}_{j_1}}
\boldsymbol{C}(\hat{\boldsymbol{q}}_w^{c_M})
\boldsymbol{C}(\hat{\boldsymbol{q}}_w^{c_1})^T
\begin{bmatrix} 1 \\ 0 \\ 0\end{bmatrix}
\label{eq:cov-reparam-28}
\\[20pt]
\frac{\partial{\boldsymbol{g}_j}}
{\partial{\beta_{j_1}}} &=
\frac{1}{\hat{\rho}_{j_1}}
\boldsymbol{C}(\hat{\boldsymbol{q}}_w^{c_M})
\boldsymbol{C}(\hat{\boldsymbol{q}}_w^{c_1})^T
\begin{bmatrix} 0 \\ 1 \\ 0\end{bmatrix}
\label{eq:cov-reparam-29}
\\[20pt]
\frac{\partial{\boldsymbol{g}_j}}
{\partial{\rho_{j_1}}} &=
- \frac{1}{\hat{\rho}_{j_1}^2}
\boldsymbol{C}(\hat{\boldsymbol{q}}_w^{c_M})
\boldsymbol{C}(\hat{\boldsymbol{q}}_w^{c_1})^T
\begin{bmatrix} \hat{\alpha}_{j_1} \\ \hat{\beta}_{j_1} \\ 1\end{bmatrix}
\label{eq:cov-reparam-30}
\end{align}
Hence
\begin{align}
\frac{\partial{\boldsymbol{g}_j}}{\partial{\boldsymbol{f}_{j_1}}} &=
\begin{bmatrix}
\frac{\partial{\boldsymbol{g}_j}}{\partial{\alpha_{j_1}}} &
\frac{\partial{\boldsymbol{g}_j}}{\partial{\beta_{j_1}}} &
\frac{\partial{\boldsymbol{g}_j}}{\partial{\rho_{j_1}}}
\end{bmatrix}
\label{eq:cov-reparam-31}
\\
&=
\frac{1}{\hat{\rho}_{j_1}}
\boldsymbol{C}(\hat{\boldsymbol{q}}_w^{c_M})
\boldsymbol{C}(\hat{\boldsymbol{q}}_w^{c_1})^T
\begin{bmatrix}
1 & 0 & -\frac{\hat{\alpha}_{j_1}}{\hat{\rho}_{j_1}} \\
0 & 1 & -\frac{\hat{\beta}_{j_1}}{\hat{\rho}_{j_1}} \\
0 & 0 & -\frac{1}{\hat{\rho}_{j_1}}
\end{bmatrix}
\label{eq:cov-reparam-32}
\\
&=
{\boldsymbol{H}}_{\boldsymbol{f}_j}
\label{eq:cov-reparam-33}
\end{align}


\chapter{Range-Visual-Inertial Odometry Observability}
\label{ch:rvio-obs}

\section{Derivation of the Observability Matrix}
\label{sec:obsmat-der}

In this subsection, we demonstrate the expressions of the k-th block row of observability matrix
$\boldsymbol{M}$ in Equations~(\ref{eq:obs-matrix}-\ref{eq:obs-matrix-f3}).

To simplify the equations, our analysis assumes the state vector
\begin{equation}
    \boldsymbol{x}^0=\begin{bmatrix}{\boldsymbol{x}'_{I}}^T & {\boldsymbol{x}_P}^{T}\end{bmatrix}^{T}
    \label{eq:x_obs}
\end{equation}
where
\begin{equation}
\boldsymbol{x}'_{I} =
\begin{bmatrix}
{\boldsymbol{q}_{w}^{i}}^T &
{\boldsymbol{b}_{g}}^{T} &
{\boldsymbol{v}_w^i}^T &
{\boldsymbol{b}_{a}}^{T} &
{\boldsymbol{p}_w^i}^T 
\end{bmatrix}^{T}
\label{eq:state-vec-i-obs}
\end{equation}
and $\boldsymbol{x}_P$ was defined in Equation~(\ref{eq:state-vec-f}) with the cartesian coordinates of the $N$ SLAM
features, $N \geq 3$. $\boldsymbol{x}'_{I}$ includes the same states as $\boldsymbol{x}_{I}$ defined in Equation~(1)
of the paper, but in a different order so we can refer the reader to \citet{hesch2012minn} for the expression and
derivation of the state transition matrix $\boldsymbol{\Phi}_{k,1}$ from time $1$ to time $k$.

The block row associated to the ranged facet update is defined as
\begin{equation}
    \boldsymbol{M}_k = \boldsymbol{H}_k \boldsymbol{\Phi}_{k,1}
    \label{eq:obs-row-def}
\end{equation}
where $\boldsymbol{H}_k$ is the Jacobian of the ranged facet measurement at time $k$ with respect to $\boldsymbol{x}^0$,
from Equations~(\ref{eq:rv-linearization-1}-\ref{eq:rv-hpj3}).

Without loss of generality, we can assume the ranged facet is constructed from the first three SLAM features in $\boldsymbol{x}_P$, then
\begin{align}
    &\mkern-35mu
    \begin{aligned}
        &
        \boldsymbol{M}_k = \big[
        \begin{matrix}
            \begin{array}{c|c|c|c|c|c|c}
                \boldsymbol{H}_{\boldsymbol{\theta}_i} &
                \boldsymbol{0}_{1\times9} &
                \boldsymbol{H}_{\boldsymbol{p}_i} &
                \boldsymbol{H}_{\boldsymbol{p}_1} &
                \boldsymbol{H}_{\boldsymbol{p}_2} &
                \boldsymbol{H}_{\boldsymbol{p}_3} &
                \boldsymbol{0}_{1\times3(N-3)}
            \end{array}
        \end{matrix}\big] \boldsymbol{\Phi}_{k,1}
    \end{aligned}
    \label{eq:obs-row-1}
    \\
    &\begin{aligned}
        &= \frac{1}{b}\Big[
        \begin{matrix}
            \begin{array}{c|c|c|}
                - \frac{a}{b} {^w}\boldsymbol{n}^T \boldsymbol{C}\left(\boldsymbol{q}_w^{c_k} \right)^T \lfloor {^c\boldsymbol{u}_r} \times\rfloor ^T &
                \boldsymbol{0}_{1\times9} &
                - {^w}\boldsymbol{n}^T
            \end{array}
        \end{matrix}
        \\
        &\quad\quad\;\;\,
        \begin{matrix}
            \begin{array}{c|}
            		\left( \left\lfloor(\boldsymbol{p}_w^{F_{3}} - \boldsymbol{p}_w^{F_{2}})\times\right\rfloor \left(\boldsymbol{p}_w^{F_{2}} - \boldsymbol{p}_w^{I_k} \right) \right)^T
            \end{array}
        \end{matrix}
        \\
        &\quad\quad\;\;\,
        \begin{matrix}
            \begin{array}{c|}
            		\left( ^w\boldsymbol{n} + \left\lfloor(\boldsymbol{p}_w^{F_{1}} - \boldsymbol{p}_w^{F_{3}})\times\right\rfloor \left(\boldsymbol{p}_w^{F_{2}} - \boldsymbol{p}_w^{I_k} \right) \right)^T
            \end{array}
        \end{matrix}
        \\
        &\quad\quad\;\;\,
        \begin{matrix}
            \begin{array}{c|c}
                \left( \left\lfloor(\boldsymbol{p}_w^{F_{2}} - \boldsymbol{p}_w^{F_{1}})\times\right\rfloor \left(\boldsymbol{p}_w^{F_{2}} - \boldsymbol{p}_w^{I_k} \right) \right)^T &
                \boldsymbol{0}_{1\times3(N-3)}
            \end{array}
        \end{matrix}\Big] \boldsymbol{\Phi}_{k,1}
    \end{aligned}
    \label{eq:obs-row-2}
    \\
    &\begin{aligned}
        &= \frac{1}{b}\Big[
        \begin{matrix}
            \begin{array}{c|}
                - \frac{a}{b} {^w}\boldsymbol{n}^T \boldsymbol{C}\left(\boldsymbol{q}_w^{c_k} \right)^T \lfloor {^c\boldsymbol{u}_r} \times\rfloor ^T  \boldsymbol{\phi}_{11}
                - {^w}\boldsymbol{n}^T \boldsymbol{\phi}_{51}
            \end{array}
        \end{matrix}
        \\
        &\quad\quad\;\;\,
        \begin{matrix}
            \begin{array}{c|}
                - \frac{a}{b} {^w}\boldsymbol{n}^T \boldsymbol{C}\left(\boldsymbol{q}_w^{c_k} \right)^T \lfloor {^c\boldsymbol{u}_r} \times\rfloor ^T  \boldsymbol{\phi}_{12}
                - {^w}\boldsymbol{n}^T \boldsymbol{\phi}_{52}
            \end{array}
        \end{matrix}
        \\
        &\quad\quad\;\;\,
        \begin{matrix}
            \begin{array}{c|c|c|}
                - {^w}\boldsymbol{n}^T \boldsymbol{\phi}_{53} &
                - {^w}\boldsymbol{n}^T \boldsymbol{\phi}_{54} &
                - {^w}\boldsymbol{n}^T
            \end{array}
        \end{matrix}
        \\
        &\quad\quad\;\;\,
        \begin{matrix}
            \begin{array}{c|}
                \left( \left\lfloor(\boldsymbol{p}_w^{F_{3}} - \boldsymbol{p}_w^{F_{2}})\times\right\rfloor \left(\boldsymbol{p}_w^{F_{2}} - \boldsymbol{p}_w^{I_k} \right) \right)^T
            \end{array}
        \end{matrix}
        \\
        &\quad\quad\;\;\,
        \begin{matrix}
            \begin{array}{c|}
                \left( ^w\boldsymbol{n} + \left\lfloor(\boldsymbol{p}_w^{F_{1}} - \boldsymbol{p}_w^{F_{3}})\times\right\rfloor \left(\boldsymbol{p}_w^{F_{2}} - \boldsymbol{p}_w^{I_k} \right) \right)^T
            \end{array}
        \end{matrix}
        \\
        &\quad\quad\;\;\,
        \begin{matrix}
            \begin{array}{c|c}
                \left( \left\lfloor(\boldsymbol{p}_w^{F_{2}} - \boldsymbol{p}_w^{F_{1}})\times\right\rfloor \left(\boldsymbol{p}_w^{F_{2}} - \boldsymbol{p}_w^{I_k} \right) \right)^T &
                \boldsymbol{0}_{1\times3(N-3)}
            \end{array}
        \end{matrix}\Big]
    \end{aligned}
    \label{eq:obs-row-3}
    \\
    &\begin{aligned}
        &= \frac{1}{b}\big[
        \begin{matrix}
            \begin{array}{c|c|c|c|c|c|c|c|}
                \boldsymbol{M}_{k,q} &
                \boldsymbol{M}_{k,b_g} &
                \boldsymbol{M}_{k,v} &
                \boldsymbol{M}_{k,b_a} &
                \boldsymbol{M}_{k,p} &
                \boldsymbol{M}_{k,p_1} &
                \boldsymbol{M}_{k,p_2} &
                \boldsymbol{M}_{k,p_3}
            \end{array}
        \end{matrix}
        \\
        &\quad\quad\;\;\,
        \begin{matrix}
            \begin{array}{c}
                \boldsymbol{0}_{1\times3(N-3)}
            \end{array}
        \end{matrix}
        \big]
        \label{eq:obs-row-4}
    \end{aligned}
\end{align}
with
\begin{align}
    &\mkern-50mu\begin{aligned}
        \boldsymbol{M}_{k,\theta} &= - \frac{a}{b} {^w}\boldsymbol{n}^T \boldsymbol{C}\left(\boldsymbol{q}_w^{c_k} \right)^T \lfloor {^c\boldsymbol{u}_r} \times\rfloor ^T  \boldsymbol{\phi}_{11} - {^w}\boldsymbol{n}^T \boldsymbol{\phi}_{51}
    \end{aligned}
    \\
    \boldsymbol{M}_{k,b_g} &= - \frac{a}{b}  {^w}\boldsymbol{n}^T \boldsymbol{C}\left(\boldsymbol{q}_w^{c_k}\right)^T \lfloor {^c\boldsymbol{u}_r} \times\rfloor \boldsymbol{\phi}_{12}  - {^w}\boldsymbol{n} \boldsymbol{\phi}_{52}
    \\
    \boldsymbol{M}_{k,v} &= - {^w}\boldsymbol{n}^T \boldsymbol{\phi}_{53}
    \\
    \boldsymbol{M}_{k,b_a} &= - {^w}\boldsymbol{n}^T \boldsymbol{\phi}_{54}
    \\
    \boldsymbol{M}_{k,p} &= - {^w}\boldsymbol{n}^T
    \\
    \boldsymbol{M}_{k,p_1} &= \left( \left\lfloor(\boldsymbol{p}_w^{F_{3}} - \boldsymbol{p}_w^{F_{2}})\times\right\rfloor
\left(\boldsymbol{p}_w^{F_{2}} - \boldsymbol{p}_w^{I_k} \right) \right)^T
    \\
    \boldsymbol{M}_{k,p_2} &= \left( {^w}\boldsymbol{n} + 
    \left\lfloor(\boldsymbol{p}_w^{F_{1}} - \boldsymbol{p}_w^{F_{3}})\times\right\rfloor
    \left(\boldsymbol{p}_w^{F_{2}} -\boldsymbol{p}_w^{I_k} \right) \right)^T 
    \\
    \boldsymbol{M}_{k,p_3} &= \left( \left\lfloor(\boldsymbol{p}_w^{F_{2}} - \boldsymbol{p}_w^{F_{1}})\times\right\rfloor \left(\boldsymbol{p}_w^{F_{2}} - \boldsymbol{p}_w^{I_k} \right) \right)^T
\end{align}
where $\boldsymbol{\phi}_*$ are integral terms defined in \citet{hesch2012minn}.

Since \citet{hesch2012minn} showed that $\boldsymbol{\phi}_{53} = (k-1) \delta t$, we can further expand
\begin{equation}
    \boldsymbol{M}_{k,v} = - (k-1) \delta t {^w}\boldsymbol{n}^T
    \label{eq:obs-matrix-v-final}
\end{equation}

Similarly, since
\begin{equation}
\boldsymbol{\phi}_{11} = \boldsymbol{C}\left(\boldsymbol{q}_{i_1}^{i_k}\right)
\end{equation}

and
\begin{equation}
\boldsymbol{\phi}_{51} = \Bigl\lfloor \boldsymbol{p}_w^{i_1} - \boldsymbol{v}_w^{i_1} (k-1) \delta t - \frac{1}{2}{^w}\boldsymbol{g}(k-1)^2 \delta t^2 - \boldsymbol{p}_w^{i_k}\times \Bigr\rfloor \boldsymbol{C}\left(\boldsymbol{q}_{i_1}^w\right)
\end{equation}

We can write
\begin{align}
	&\mkern-50mu
    \begin{aligned}
    		\boldsymbol{M}_{k,\theta} &= - \frac{a}{b} {^w}\boldsymbol{n}^T \boldsymbol{C}\left(\boldsymbol{q}_w^{c_k} \right)^T \lfloor {^c\boldsymbol{u}_r} \times\rfloor ^T  \boldsymbol{C}\left(\boldsymbol{q}_{i_1}^{i_k}\right)
    		\\
    		&\quad\,
    		- {^w}\boldsymbol{n}^T \Bigl\lfloor \boldsymbol{p}_w^{i_1} - \boldsymbol{v}_w^{i_1} (k-1) \delta t - \frac{1}{2}{^w}\boldsymbol{g}(k-1)^2 \delta t^2 - \boldsymbol{p}_w^{i_k}\times \Bigr\rfloor \boldsymbol{C}\left(\boldsymbol{q}_{i_1}^w\right)
    	\end{aligned}
    \label{eq:obs-matrix-theta1}
    \\
    &\mkern-10mu
    \begin{aligned}
    		&= {^w}\boldsymbol{n} \Bigg( - \frac{a}{b} \boldsymbol{C}\left(\boldsymbol{q}_w^{c_k}\right)^T \left\lfloor {^c\boldsymbol{u}_r} \times \right\rfloor \boldsymbol{C}\left(\boldsymbol{q}_w^{i_k}\right)
    		\\
    		&\quad\,
    		- \Bigl\lfloor \boldsymbol{p}_w^{i_1} - \boldsymbol{v}_w^{i_1} (k-1) \delta t - \frac{1}{2}{^w}\boldsymbol{g}(k-1)^2 \delta t^2 - \boldsymbol{p}_w^{i_k}\times \Bigr\rfloor \Bigg) \boldsymbol{C}\left(\boldsymbol{q}_{i_1}^w\right)
    	\end{aligned}
    \label{eq:obs-matrix-theta3}
\end{align}
End of the proof.

\section{Observability Under Constant Acceleration}
\label{sec:obs-aconst}

In this subsection, we want to prove that under a constant acceleration ${\boldsymbol{a}_w^{i}}$, the vector
\begin{equation}
\boldsymbol{N}_s =
\begin{bmatrix}
{\boldsymbol{p}_w^{i_1}}^T &
{\boldsymbol{v}_w^{i_1}}^T &
{\boldsymbol{0}_{6\times1}}^T &
- {^{i}\boldsymbol{a}_w^{i}}^T &
{\boldsymbol{p}_w^{F_1}}^T &
... &
{\boldsymbol{p}_w^{F_N}}^T
\end{bmatrix}^{T}
\label{eq:scale-null-vector}
\end{equation}
is not in the right nullspace of $\boldsymbol{M}_k$. Hence, we need to demonstrate that $\boldsymbol{M}_k \boldsymbol{N}_s \neq \boldsymbol{0}$.
\begin{align}
    &\begin{aligned}
    \boldsymbol{M}_k \boldsymbol{N}_s =&
        - {^w}\boldsymbol{n}^T {\boldsymbol{p}_w^{i_1}}
        - (k-1) \delta t {^w}\boldsymbol{n}^T {\boldsymbol{v}_w^{i_1}}
        + {^w}\boldsymbol{n}^T \boldsymbol{\phi}_{54} {^{i}\boldsymbol{a}_w^{i}}
        \\&
        + \left( \left\lfloor(\boldsymbol{p}_w^{F_{3}} - \boldsymbol{p}_w^{F_{2}})\times\right\rfloor \left(\boldsymbol{p}_w^{F_{2}} - \boldsymbol{p}_w^{I_k} \right) \right)^T {\boldsymbol{p}_w^{F_1}}
        \\&
        + \left( {^w}\boldsymbol{n} + \left\lfloor(\boldsymbol{p}_w^{F_{1}} - \boldsymbol{p}_w^{F_{3}})\times\right\rfloor \left(\boldsymbol{p}_w^{F_{2}} -\boldsymbol{p}_w^{I_k} \right) \right)^T {\boldsymbol{p}_w^{F_2}}
        \\&
        + \left( \left\lfloor(\boldsymbol{p}_w^{F_{2}} - \boldsymbol{p}_w^{F_{1}})\times\right\rfloor \left(\boldsymbol{p}_w^{F_{2}} - \boldsymbol{p}_w^{I_k} \right) \right)^T {\boldsymbol{p}_w^{F_3}}
    \end{aligned}
    \label{eq:mk-ns-1}
\end{align}

\citet{wu2016minn} show that, under constant acceleration,
\begin{equation}
\boldsymbol{\phi}_{54} {^{i}\boldsymbol{a}_w^{i}} = - \left( {\boldsymbol{p}_w^{i_k}} - {\boldsymbol{p}_w^{i_1}} - (k-1)\delta t {\boldsymbol{v}_w^{i_1}} \right)
\end{equation}

so
\begin{align}
    &\mkern-60mu\begin{aligned}
    \boldsymbol{M}_k \boldsymbol{N}_s =&
        - {^w}\boldsymbol{n}^T {\boldsymbol{p}_w^{i_1}}
        - (k-1) \delta t {^w}\boldsymbol{n}^T {\boldsymbol{v}_w^{i_1}}
        - {^w}\boldsymbol{n}^T \left( {\boldsymbol{p}_w^{i_k}} - {\boldsymbol{p}_w^{i_1}} - (k-1)\delta t {\boldsymbol{v}_w^{i_1}} \right)
        \\&
        + \left( \left\lfloor(\boldsymbol{p}_w^{F_3} - \boldsymbol{p}_w^{F_2})\times\right\rfloor \left(\boldsymbol{p}_w^{F_2} - \boldsymbol{p}_w^{I_k} \right) \right)^T {\boldsymbol{p}_w^{F_1}}
        \\&
        + \left( {^w}\boldsymbol{n} + \left\lfloor(\boldsymbol{p}_w^{F_1} - \boldsymbol{p}_w^{F_3})\times\right\rfloor \left(\boldsymbol{p}_w^{F_2} -\boldsymbol{p}_w^{I_k} \right) \right)^T {\boldsymbol{p}_w^{F_2}}
        \\&
        + \left( \left\lfloor(\boldsymbol{p}_w^{F_2} - \boldsymbol{p}_w^{F_1})\times\right\rfloor \left(\boldsymbol{p}_w^{F_2} - \boldsymbol{p}_w^{I_k} \right) \right)^T {\boldsymbol{p}_w^{F_3}}
    \end{aligned}
    \\
    &\begin{aligned}
        =&
        {^w}\boldsymbol{n}^T \left( \boldsymbol{p}_w^{F_2} - {\boldsymbol{p}_w^{i_k}}\right)
        + \left(\boldsymbol{p}_w^{F_2} - \boldsymbol{p}_w^{I_k} \right)^T \Big( \left\lfloor(\boldsymbol{p}_w^{F_3} - \boldsymbol{p}_w^{F_2})\times\right\rfloor^T {\boldsymbol{p}_w^{F_1}}
        \\&
        + \left\lfloor(\boldsymbol{p}_w^{F_1} - \boldsymbol{p}_w^{F_3})\times\right\rfloor^T {\boldsymbol{p}_w^{F_2}} + \left\lfloor(\boldsymbol{p}_w^{F_2} - \boldsymbol{p}_w^{F_1})\times\right\rfloor^T {\boldsymbol{p}_w^{F_3}}\Big)
    \end{aligned}
\end{align}
The cross product of the first term can be modified such that
\begin{align}
    \left(\boldsymbol{p}_w^{F_2} - \boldsymbol{p}_w^{I_k} \right)^T &\left\lfloor(\boldsymbol{p}_w^{F_3} - \boldsymbol{p}_w^{F_2})\times\right\rfloor^T {\boldsymbol{p}_w^{F_1}} 
    \\
    &= \left(\boldsymbol{p}_w^{F_2} - \boldsymbol{p}_w^{I_k} \right)^T \left\lfloor(\boldsymbol{p}_w^{F_3} - \boldsymbol{p}_w^{F_2})\times\right\rfloor^T \left( {\boldsymbol{p}_w^{F_1} - \boldsymbol{p}_w^{I_k} + \boldsymbol{p}_w^{I_k}} \right)
    \\
    &= \left(\boldsymbol{p}_w^{F_2} - \boldsymbol{p}_w^{I_k} \right)^T \left\lfloor(\boldsymbol{p}_w^{F_3} - \boldsymbol{p}_w^{F_2})\times\right\rfloor^T \boldsymbol{p}_w^{I_k}
\end{align}
By definition of the cross product,
\begin{equation}
\exists\,\lambda \in \mathbb{R}, \left\lfloor(\boldsymbol{p}_w^{F_3} - \boldsymbol{p}_w^{F_2})\times\right\rfloor^T \left( \boldsymbol{p}_w^{F_1}-\boldsymbol{p}_w^{I_k} \right) = \lambda {^w}\boldsymbol{n}
\end{equation}
and 
\begin{equation}
\lambda \left(\boldsymbol{p}_w^{F_2} - \boldsymbol{p}_w^{I_k} \right)^T {^w}\boldsymbol{n} = 0
\end{equation}
Thus, by applying this to all cross-product terms,
\begin{align}
    &\mkern-40mu
    \begin{aligned}
    		\boldsymbol{M}_k \boldsymbol{N}_s &= {^w}\boldsymbol{n}^T \left( \boldsymbol{p}_w^{F_2} - {\boldsymbol{p}_w^{i_k}}\right)
    		\\&\quad
    		+ \left( \boldsymbol{p}_w^{F_2} - \boldsymbol{p}_w^{I_k} \right)^T \bigl\lfloor( \underbrace{\boldsymbol{p}_w^{F_3} - 						\boldsymbol{p}_w^{F_2} + \boldsymbol{p}_w^{F_1} - \boldsymbol{p}_w^{F_3} + \boldsymbol{p}_w^{F_2} - \boldsymbol{p}_w^{F_1} 				}_{\boldsymbol{0}})\times\bigr\rfloor^T \boldsymbol{p}_w^{I_k}
    \end{aligned}
    \\
    &\quad
    = {^w}\boldsymbol{n}^T \left( \boldsymbol{p}_w^{F_2} - {\boldsymbol{p}_w^{i_k}}\right)
\end{align}
By definition, if the three features of the facet are not aligned in the image, ${^w}\boldsymbol{n}^T \left( \boldsymbol{p}_w^{F_2} - {\boldsymbol{p}_w^{i_k}}\right) \neq \boldsymbol{0}$.

End of proof.

As a note in the hover case, looking back at Equation~(\ref{eq:mk-ns-1}), we can note that when ${\boldsymbol{v}_w^{i_1}} = \boldsymbol{0}$, then ${\boldsymbol{p}_w^{i_1}} = {\boldsymbol{p}_w^{i_k}}$ and
\begin{equation}
\boldsymbol{N}_h =
\begin{bmatrix}
{\boldsymbol{0}_{24\times1}}^T &
{\boldsymbol{p}_w^{F_4}}^T &
... &
{\boldsymbol{p}_w^{F_N}}^T
\end{bmatrix}^{T}
\label{eq:partial-scale-hover-null-vector}
\end{equation}
becomes a vector of the right nullspace, which spans the depth of the SLAM features not included in the facet,
as discussed in Section~\ref{sec:unobs-dir}.


\end{appendices}

\cleardoublepage
\phantomsection
\bibliographystyle{apalike}
\addcontentsline{toc}{chapter}{References}
\bibliography{bibliography}

\end{document}